\documentclass[lettersize,journal,compsoc]{IEEEtran}
\usepackage{caption}
\usepackage{amsmath,amsfonts}
\usepackage{algorithmic}
\usepackage{algorithm}
\usepackage{array}
\usepackage[caption=false,font=normalsize,labelfont=sf,textfont=sf]{subfig}
\usepackage{textcomp}
\usepackage{stfloats}
\usepackage{url}
\usepackage{verbatim}
\usepackage{graphicx}
\usepackage{cite}
\usepackage{wrapfig,lipsum}
\usepackage[utf8]{inputenc} 
\usepackage[T1]{fontenc}    
\usepackage{hyperref}       
\usepackage{url}            
\usepackage{booktabs}       
\usepackage{amsfonts}       
\usepackage{nicefrac}       
\usepackage{microtype}      
\usepackage{xcolor,colortbl}
\usepackage{multirow}

\usepackage{microtype}
\usepackage{graphicx}

\usepackage{url}
\usepackage{tikz}
\usepackage{comment}
\usepackage{amsmath,amssymb} 
\usepackage{color}
\usepackage[ruled,algo2e]{algorithm2e}

\usepackage[accsupp]{axessibility}  

\usepackage{amsmath}
\usepackage{amssymb}
\usepackage{mathtools}
\usepackage{amsthm}
\newtheorem{theorem}{Theorem}
\newtheorem{lemma}{Lemma}
\theoremstyle{definition}

\newtheorem{property}{\em Property}
\newcolumntype{P}[1]{>{\centering\arraybackslash}p{#1}}

\usepackage[capitalize,noabbrev]{cleveref}
\theoremstyle{plain}

\theoremstyle{definition}
\newtheorem{definition}{Definition}

\theoremstyle{remark}

\usepackage[textsize=tiny]{todonotes}

\def\Renyi/{\text{R\'{e}nyi}}

\begin{document}

\bstctlcite{IEEEexample:BSTcontrol}

\def\eqc{\stackrel{c}{=}}
\def\wc{{\bf v}}
\def\wf{{\bf w}}
\def\uc{{\bf u}}
\def\y{{\bf y}}
\def\t{{\bf t}}
\def\softmax{{\sigma}}
\def\sigmoid{{\varsigma}}
\def\FL{{\cal FL}}

\title{Discriminative Entropy Clustering \\
and its Relation to K-means and SVM}

\author{
    \IEEEauthorblockN{Zhongwen(Rex) Zhang\IEEEauthorrefmark{1},  Yuri Boykov\IEEEauthorrefmark{1}}\\
    \IEEEauthorblockA{\IEEEauthorrefmark{1}University of Waterloo
    \\\{z889zhan, yboykov\}@uwaterloo.ca}
}


\markboth{IEEE Trans. on Pattern Analysis and Machine Intelligence, submitted in September 2024}{Zhang and Boykov: Discriminative Entropy Clustering and its relation to K-means and SVM}


\maketitle

\begin{abstract}
Maximization of mutual information between the model's input and output is formally related to ``decisiveness'' and ``fairness'' of the softmax predictions \cite{MacKay1991}, motivating these unsupervised entropy-based criteria for clustering. First, in the context of linear softmax models, we discuss some general properties of 
entropy-based clustering. Disproving some earlier claims, we point out fundamental differences with K-means. On the other hand, we prove the margin maximizing property for decisiveness establishing a relation to SVM-based clustering. Second, we propose a new self-labeling formulation of entropy clustering for general softmax models. The pseudo-labels 
are introduced as auxiliary variables ``splitting'' the fairness and decisiveness. 
The derived self-labeling loss includes the reverse cross-entropy robust to pseudo-label errors and
allows an efficient EM solver for pseudo-labels. 
Our algorithm improves the state of the art on several standard benchmarks for deep clustering.
\end{abstract}


\section{Introduction} \label{sec:intro}

{\em Mutual information} (MI) was proposed as a criterion for clustering by Bridle et al. \cite{MacKay1991}.
It is motivated as a general information-theoretic measure of the ``correlation'' between the data $X$ and the class labels $C$. Starting from MI definition as Kullback–Leibler (KL) divergence between the joint density and the product of the marginals for $X$ and $C$, 
Bridle et al.\cite{MacKay1991} derive a simple clustering loss for softmax models. 

The MI criterion also unifies some well-known generative 
and discriminative approaches to clustering. In particular,   
consider two equivalent entropy-based MI formulations
\begin{eqnarray} 
 MI(C,X) &    =   & H(X) \;\; -\;\; H(X\,|\,C) \quad\quad\text{(gen.)} \label{eq:mi_generative} \\ 
         & \equiv & H(C)  \;\;-\;\; H(C\,|\,X) \quad\quad\text{(disc.)} \label{eq:mi_discriminative}
\end{eqnarray}
where $H(\cdot)$ and $H(\cdot\,|\,\cdot)$ are the {\em entropy} and the {\em conditional entropy} functions over
distributions corresponding to the random variables $X$ and $C$.
Two terms in \eqref{eq:mi_discriminative} can directly evaluate classes $C$ 
predicted by discriminative posterior models, e.g. softmax models. As detailed in Section \ref{sec:entropy clustering}, 
these two terms represent standard clustering criteria, {\em fairness} and {\em decisiveness} \cite{MacKay1991},
used for ``deep'' clustering with neural networks \cite{MacKay1991,Perona2010,hu2017learning,chang2017deep,ji2019invariant,YM.2020Self-labelling,ismail2021}. 
On the other hand, the equivalent formulation of MI in \eqref{eq:mi_generative} relates 
to standard generative algorithms for clustering \cite{jiang2017variational,yang2017towards,caron2018deep} and unsupervised or weakly-supervised image segmentation \cite{zhu1996region,chan2001active,rother2004grabcut}. 
Such algorithms minimize the entropy $H(X|C)$ of data $X$ in each cluster $C$, where fitting density models helps to 
estimate the entropy. Section \ref{sec:Kmeans} details the relation of criterion
\eqref{eq:mi_generative} to the most basic generative clustering algorithm, K-means. 

\begin{figure*}[t!]
    \centering
    \begin{tabular}{c @{\extracolsep{1cm}} c}
    \multicolumn{2}{c}{\small \sc two linear decision functions over 2D features $X\in{\cal R}^2$} \\[0.2ex]
    {\footnotesize $ k_\mu (X) := \arg\min_k\, \|X-\mu_k\| $} &  
    {\footnotesize $\sigma_\wc (X) := \sigma(\wc^\top X)  \equiv \text{soft-max}(\wc^\top X) $}  \\
        \includegraphics[width=0.3\linewidth]{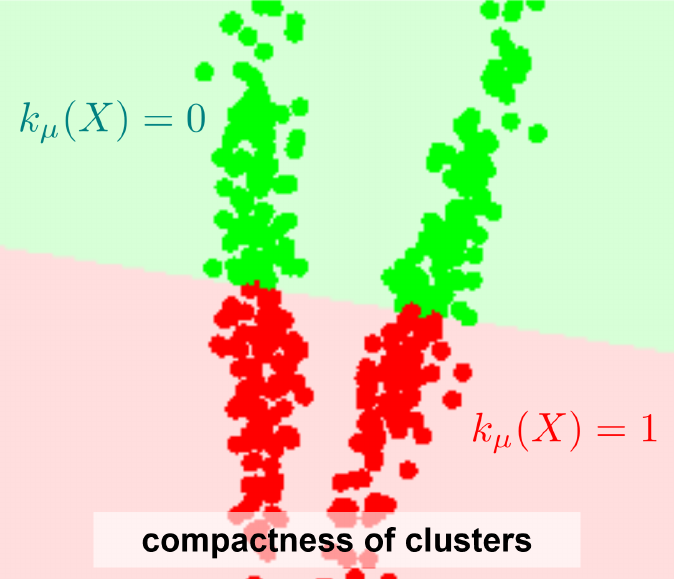} &  
        \includegraphics[width=0.3\linewidth]{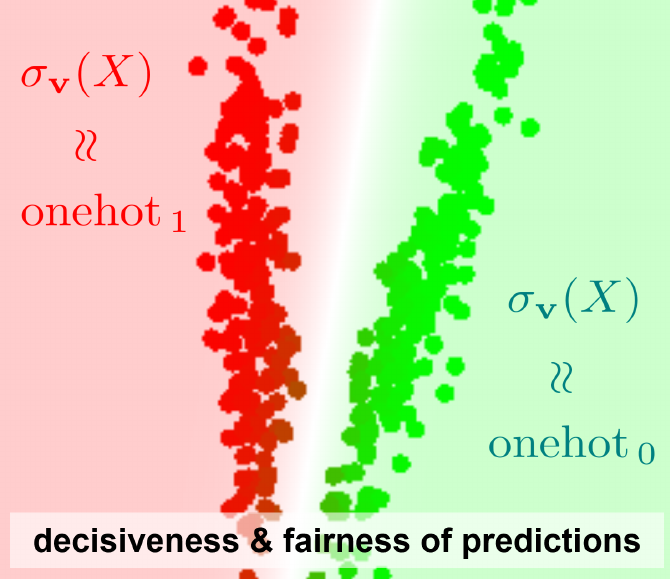} \\  
         (a) variance clustering (K-means)
         & (b) entropy clustering (\ref{eq:mi},\ref{eq:postmodel_shallow})
    \end{tabular}
    \caption{Entropy clustering vs. K-means - binary example ($K=2$) for 2D data $\{X_i\}$
    comparing two linear methods of similar parametric complexity: (a) $K$-means $\mu_k\in{\cal R}^2$ and (b) entropy clustering \eqref{eq:mi} with a linear model \eqref{eq:postmodel_shallow} defined by $K$-column matrix $\wc=[\wc_k]$ 
    with linear discriminants $\wc_k\in{\cal R}^{2+1}$ (incl. bias).
    Red and green colors in (a) and (b) illustrate decision/prediction functions, 
    $k_\mu(X) :=\arg\min_k\, \|X-\mu_k\|$ and $\sigma_\wc(X):=\sigma(\wc^\top X)$, 
    corresponding to the optimal parameters $\mu$ and $\wc$ minimizing two losses:
    (a) compactness or {\em variance} of clusters $\sum_{i} \|X_i-\mu_{k_i}\|^2 $ where $k_i=k_\mu(X_i)$, and
    (b) {\em decisiveness} and {\em fairness} $\overline{H(\sigma)} - H ( \bar{\sigma} )$, 
    see \eqref{eq:mi}, where $\sigma_i = \sigma(\wc^\top X_i)$. 
    Color transparency in (b) visualizes ``soft'' decisions $\sigma(\wc^\top X)$; the linear boundary ``blur'' 
    is proportional to $\frac{1}{\|v\|}$.  
    Unlike {\em low-variance} (a), the optimal clusters in (b) have the {\em maximum margin} among all fair/balanced solutions, assuming ``infinitesimal'' norm regularization $\|\wc\|^2$ discussed in Sec.\ref{sec:SVM}.
    \label{fig:teaser}
}
\end{figure*}

Despite equivalence, criteria (\ref{eq:mi_generative},\ref{eq:mi_discriminative}) can lead to clustering algorithms producing 
different results depending on their specific generative or discriminative model choices, i.e. {\em hypothesis spaces}. 
For example, Figure \ref{fig:teaser} shows optimal solutions for (a) K-means minimizing the {\em variance} of each cluster, i.e.
entropy $H(X|C)$ assuming isotropic Gaussian density, and (b) the linear softmax model minimizing \eqref{eq:mi_discriminative}. While both algorithms make linear decisions, K-means produces compact clusters due to its implicit bias to
isotropic Gaussian densities. In contrast, the linear softmax model finds balanced or ``fair'' clusters with ``decisive'' decision boundary corresponding to the {\em maximum margin}, as we later prove in Theorem \ref{th:Ralpha} ($\alpha=2$).
We will revisit Figure \ref{fig:teaser} again.

This paper's focus is clustering based on softmax models and unsupervised entropy loss functions \eqref{eq:mi_discriminative} 
inspired by \cite{MacKay1991}. We refer to this general group of methods as {\em discriminative entropy clustering}.
The rest of the introduction provides the background and motivation for our work. 
Sections \ref{sec:entropy clustering}-\ref{sec:related_work}
present the necessary terminology for the discriminative entropy clustering problem, 
its regularization, and its {\em self-labeling} formulations. In particular, 
Section \ref{sec:clusterting VS representation} discusses the significance of model complexity and data representation.
Finally, Section \ref{sec:contributions} summarizes our main contributions presented in the main parts of the paper.


\subsection{Discriminative entropy clustering basics} \label{sec:entropy clustering}

This Section introduces our terminology for entropy-based clustering 
with softmax models. We consider discriminative classification models that could be linear (single layer) or 
complex multi-layered, and assume probability-type output often interpreted as a 
(pseudo-) posterior. Typically, such outputs are
produced by the {\em softmax} function $\sigma:{\cal R}^K\rightarrow\Delta^K $
mapping $K$ raw classifier outputs $x=(x^1,\dots,x^K)$, e.g. real-valued ``logits'' or ``scores'', 
to $K$ class probabilities $$\sigma^k (x)\;:=\;\frac{\exp {x^k}}{\sum_c \exp {x^c}}\quad\quad\text{for}\quad 1\leq k\leq K$$
forming a categorical distribution $\sigma=(\sigma^1,\dots,\sigma^K)\in\Delta^K$ 
representing a point on the {\em probability simplex}
$$ \Delta^K\;:=\;\{(p^1,\dots,p^K)\in {\cal R}^K\,|\,p^k\geq 0,\,\sum p^k=1 \}. $$
For shortness, this paper uses the same symbol for functions and examples of their output, e.g. specific prediction values $\sigma\in\Delta^K$. In particular, $\sigma_X$ may denote the prediction for any given input $X$. If $i$ is an integer, $\sigma_i$ denotes the prediction for 
the specific example $X_i$ in the training dataset $\{X_i\}_{i=1}^N$.

The simplest {\em linear} softmax model can be defined as
\begin{equation} \label{eq:postmodel_shallow}
    \sigma(\wc^\top X)
\end{equation}
where for any input vector $X$ the matrix of parameters $\wc$ produces $K$-logit vector $x=\wc^\top X$ 
mapped to a probability distribution by the softmax function.
For shortness, we use a {\em homogeneous} representation for the linear classifier so that
$\wc^\top X$ stands for an affine transformation including the {\em bias}. 

More complex non-linear network models compose a linear softmax model \eqref{eq:postmodel_shallow} 
with some {\em representation} layers mapping input $X$ to ``deep'' features $f_\wf(X)$
\begin{equation} \label{eq:postmodel_deep}
    \sigma(\wc^\top f_\wf(X))
\end{equation}
where the trainable parameters $\wf$ of the {\em embedding function} $f_\wf$ are distinguished from the linear classifier 
parameters $\wc$. The linear model \eqref{eq:postmodel_shallow} 
is a special case of \eqref{eq:postmodel_deep} for $f(X)=X$. Typical ``deep'' representation $f_\wf(X)$ significantly 
changes the dimensions of the input $X$. It is convenient to assume that $M$ always represents the dimensions of 
the linear head's input, i.e. the (homogeneous) matrix $\wc$ has size $(M+1)\times K$.

Assuming softmax models as above, Bridle et al. \cite{MacKay1991} derive the following clustering loss for data 
$\{X_i\}_{i=1}^N$
\begin{equation} \label{eq:mi}
    L_{ec} \;\;: = \;- MI(C,X)\;\;\;\approx\;\;\;\;\overline{H(\sigma)} \;-\;  H(\overline{\sigma}) 
\end{equation}
based on the Shannon entropy $H(p):=-\sum_k p^k \ln p^k$ for categorical distributions $p\in\Delta^K$.
The average entropy of the model output $$\overline{H(\sigma)} := \frac{1}{N}\sum_i H(\sigma_i)$$ represents $H(C|X)$ in \eqref{eq:mi_discriminative}.
The entropy of the average output $$H(\overline{\sigma})\quad\quad\text{where}\quad\quad\overline{\sigma}:=\frac{1}{N}\sum_i \sigma_i$$ 
is the entropy of class predictions over the whole dataset corresponding to $H(C)$ in \eqref{eq:mi_discriminative}.

Loss \eqref{eq:mi} is minimized over model parameters $\wc$ and $\wf$ in \eqref{eq:postmodel_shallow} or 
\eqref{eq:postmodel_deep}, e.g. by {\em gradient descent} 
or {\em backpropagation} \cite{Hinton1986learning}.
Larger entropy $H(\overline{\sigma})$ encourages ``fair'' predictions with a balanced support of all categories across 
the whole dataset, while smaller $\overline{H(\sigma)}$ encourages confident or ``decisive'' prediction 
at each data point suggesting that decision boundaries are away from the training examples \cite{bengio2004semi}.

\subsection{Model complexity and representation learning} \label{sec:clusterting VS representation} 

Criterion \eqref{eq:mi} provides strong constraints for 
well-regularized or simple parametric models. For example, in the case of linear softmax models \eqref{eq:postmodel_shallow}, 
we prove the margin-maximizing property, see Figure~\ref{fig:teaser}(b), relating \eqref{eq:mi}  to SVM clustering \cite{Schuurmans2004}.

Clustering algorithms for softmax models can also be motivated by 
the powerful representation of data behind the deep network models \eqref{eq:postmodel_deep}.
Some criteria related to MI are also studied in the context of 
{\em representation learning} \cite{boudiaf2020unifying}\footnote{Boudiaf et al. \cite{boudiaf2020unifying} also discuss
{\em generative} and {\em discriminative} ``views'' on MI exactly as in (\ref{eq:mi_generative}-\ref{eq:mi_discriminative}),
but focus on { \bf supervised} representation learning 
where features $X$ are optimized assuming given class labels $C$.}, 
\cite{hjelm2018learning,oord2018representation,tschannen2019mutual,YM.2020Self-labelling}, 
but this is not our focus. We study (\ref{eq:mi_discriminative},\ref{eq:mi}) as a clustering criterion 
where decisions $C$ are optimized for fixed data $X$. Nevertheless, this approach applies to
complex networks \eqref{eq:postmodel_deep} where internal layers can be seen as 
responsible for representation $f_\wf(X)$. Our experiments with networks do not evaluate the quality of 
representation separately from clustering and view the internal layers mainly as an integral part of a complex model.
Instead, we are concerned with regularization of complex models in the context of clustering.

\subsection{Regularized entropy clustering} \label{sec:into_reg}

\begin{figure*}
    \centering
    \begin{tabular}{c @{\extracolsep{1cm}} c}
       \includegraphics[width=0.35\linewidth]{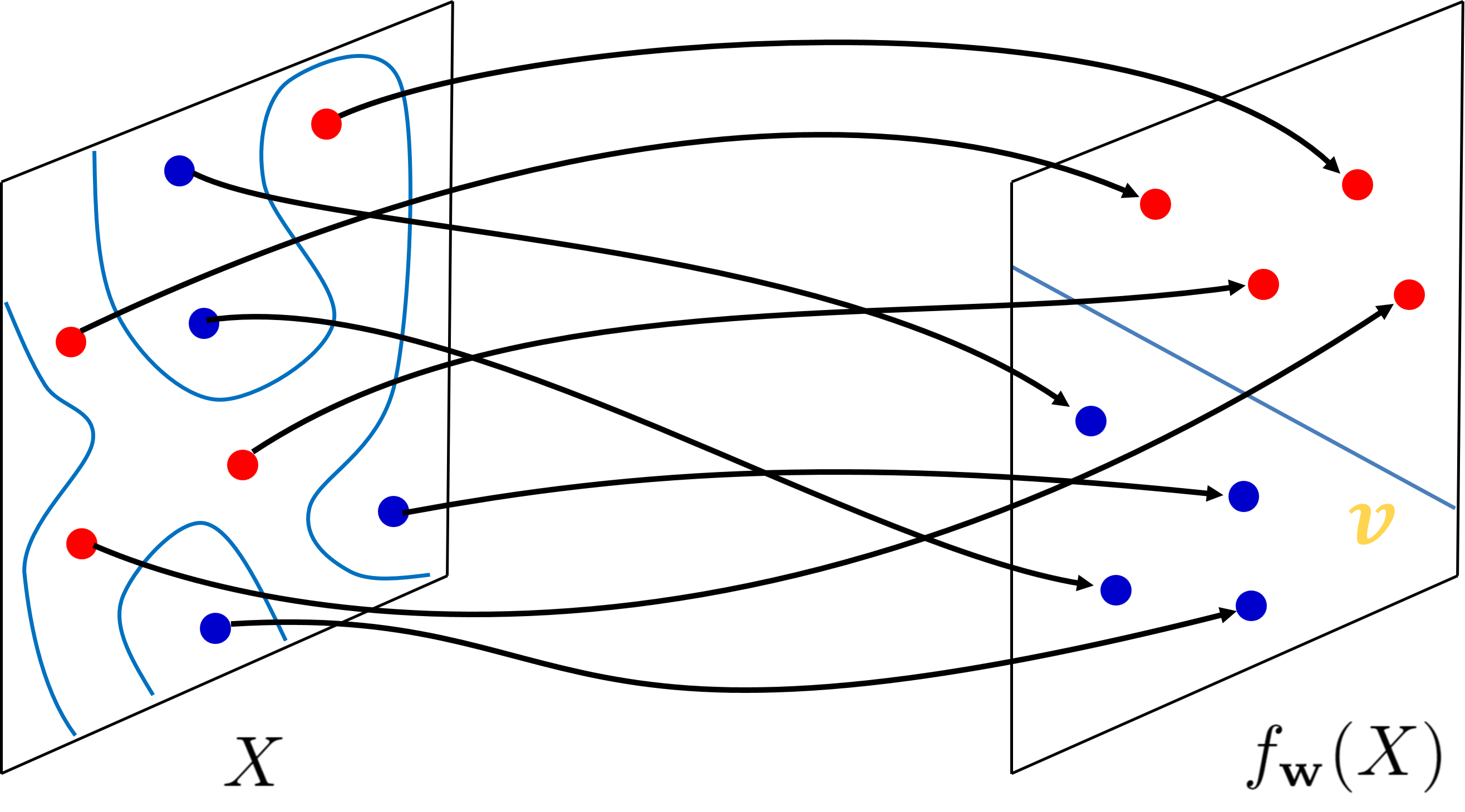}  & 
       \includegraphics[width=0.35\linewidth]{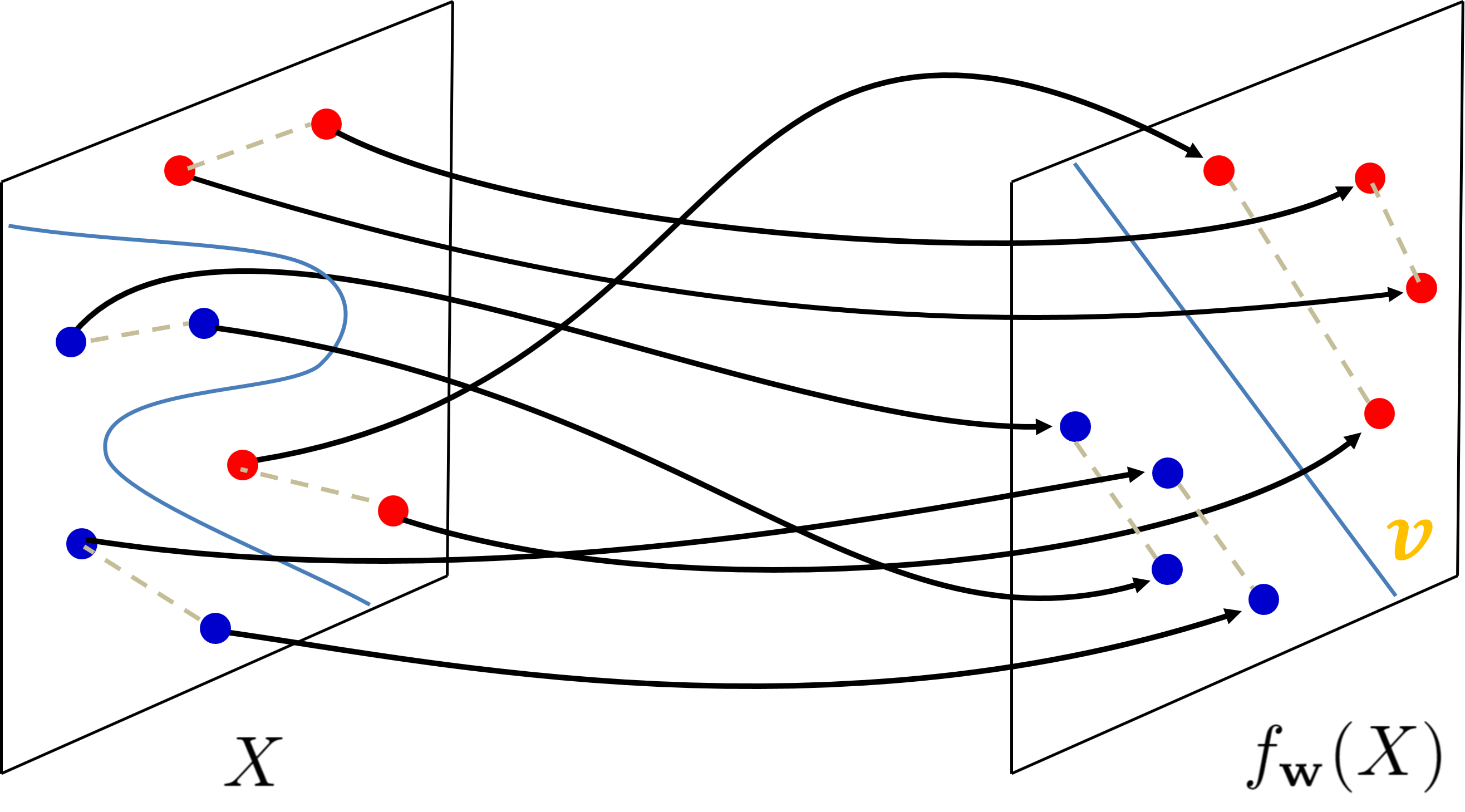} \\
        (a) under-regularized model $\Rightarrow$  random clusters & (b) self-augmentation $\Rightarrow$ regular clusters, weak isometry
    \end{tabular}
    \caption{Model regularization is required: (a) arbitrarily complex model \eqref{eq:postmodel_deep} can create any random clusters of the input $\{X_i\}$ despite the linear partitioning in the space of deep features $f_\wf(X)$. 
    Self-augmentation loss \eqref{eq:self-augment loss} regularizes the clusters (b), 
    as well as the mapping $f$. Indeed, similar inputs $\{X,X'\}$ 
    are mapped to features equidistant from the decision boundary.
    Stronger forms of isometry can be enforced by {\em contrastive losses} \cite{chopra2005learning,schroff2015facenet,sohn2016improved,chen2020simple}, particularly if negative pairs are also available. In general, model regularization with domain-specific constraints or augmentation is important for the quality of clustering.}
    \label{fig: underconstrain mapping}
\end{figure*}

Bridle \& McKay \cite{MacKay1991} argue that MI maximization may allow arbitrarily complex solutions for 
under-regularized network models, as illustrated in Figure \ref{fig: underconstrain mapping}(a) for \eqref{eq:postmodel_deep}. 
They note that
\begin{quote}
''{\em ... [MI] could be maximized by any implausible classification of the input... when we use more complex models. This criterion encourages... objective techniques for regularising classification 
networks... \cite{mackay1992practical,mackay1992bayesian}.}''
\end{quote}
For example, a Bayesian approach to network regularization \cite{mackay1992practical}   
combines training losses with the squared $L_2$ norm of all network weights interpreted 
as a {\em log-prior} or {\em weight energy}.
Following \cite{mackay1992practical} and \cite{Perona2010}, regularized version of the entropy clustering loss \eqref{eq:mi} incorporates the norm of network parameters motivated as their isotropic Gaussian prior
\begin{align} \nonumber
    L_{mi+decay}\;\; = & \;\;\;\;\; \overline{H(\sigma)} \;\;-\; \;\;\; H(\overline{\sigma})  
    \;\;\;\;\;\;+\;  \| [\wc,\wf] \|^2  \\
  \label{eq:mi+decay}
      \eqc & \;\;\;\;\;\overline{H(\sigma)} \;\;+\; KL(\overline{\sigma}\,\|\,u)   \;+\; \|[\wc,\wf]\|^2 
\end{align}
where $\eqc$ represents equality up to an additive constant and $u$ is a uniform distribution over $K$ classes.
The equivalent loss formulation \eqref{eq:mi+decay} uses KL divergence 
motivated in \cite{Perona2010} by the possibility to generalize the fairness constraint to any target balancing 
distribution different from the uniform.

Unsupervised representation learning techniques \cite{hjelm2018learning,oord2018representation,chen2020simple,he2020momentum} are also relevant as mechanisms for constraining the network. In particular, {\em self-augmentation} techniques are widely used 
for both clustering and representation learning \cite{hu2017learning,ji2019invariant,YM.2020Self-labelling}. 
For example, maximization of MI between predictions for input $X$ and its augmentation $X'$
can improve representation \cite{ji2019invariant}. 

For large network models employed in this work, we use only generic network 
regularization techniques based on squared $L_2$ norm of the network weights \eqref{eq:mi+decay}
and a standard self-augmentation loss \cite{hu2017learning,YM.2020Self-labelling,chen2021exploring} directly enforcing consistent clustering
$\sigma_X\approx\sigma_{X'}$ for input pairs $\{X,X'\}$ 
\begin{equation} \label{eq:self-augment loss}
L_{sa}\;\;=\sum_{ \{X,X'\}\in{\cal N}_{a}} KL(\sigma_X\|\sigma_{X'}) + KL(\sigma_{X'}\|\sigma_X)  
\end{equation}
where ${\cal N}_a$ is the set of all pairs of augmented examples. This loss implies that 
similar inputs are mapped to deep features equidistant from the decision boundary. 
We refer to this as {\em weak isometry} between the space of inputs $X$ and 
the embedding space of ``deep'' features $f_\wf(X)$, see Figure \ref{fig: underconstrain mapping}(b). 

\subsection{{\em Self-labeling} methods for entropy clustering} \label{sec:related_work}

Optimization of losses \eqref{eq:mi} or \eqref{eq:mi+decay} during network training could be
done with standard gradient descent or backpropagation
\cite{MacKay1991,Perona2010,hu2017learning}. However, the difference between the two entropy
terms implies non-convexity presenting challenges for the gradient descent. 
This motivates alternative approaches to optimization. It is common to approximate \eqref{eq:mi} with some {\em surrogate loss} incorporating auxiliary or hidden variables $y$ representing {\em pseudo-labels} for unlabeled data points $X$, 
which are estimated jointly with optimization of the network parameters
\cite{ghasedi2017deep,YM.2020Self-labelling,ismail2021}.
Typically, such {\em self-labeling} approaches to entropy clustering iteratively 
optimize the surrogate loss over pseudo-labels and network parameters, similarly to 
Lloyd's algorithm for $K$-means or EM algorithm for Gaussian mixtures \cite{bishop:2006}. 
While the network parameters are still optimized via gradient descent, 
the pseudo-labels can be optimized via more powerful algorithms.

For example, \cite{YM.2020Self-labelling} formulate self-labeling using the following constrained optimization problem with discrete pseudo-labels $y$ tied to predictions by {\em cross entropy} $H(y,\sigma)$
\begin{align} \label{eq:vidaldi}
    L_{ce} \;\;=\;\; \overline{H(y,\sigma)} \;\;\;\;\;\;\;\; s.t.\;\;\;  y\in\Delta^K_{0,1}  \;\;\;and\;\;\; \bar{y}=u
\end{align}
where $\Delta^K_{0,1}$ are {\em one-hot} distributions, i.e. the vertices 
of the probability simplex $\Delta^K$.
Besides proximity between $\sigma$ and $y$, the cross-entropy in \eqref{eq:vidaldi} 
encourages the decisiveness, while the fairness is represented by the hard constraint $\bar{y}=u$. Assuming fixed pseudo-labels $y$, the network training is done by minimizing 
the standard cross entropy loss $H(y,\sigma)$ convex w.r.t. $\sigma$. 
Then, model predictions are fixed and \eqref{eq:vidaldi} is minimized over $y$. 
Note that $H(y,\sigma)$ is linear with respect to $y$ and its minimum over 
simplex $\Delta^K$ is achieved by one-hot distributions corresponding to 
$\arg\max_k(\sigma)$ at each data point. 
However, the ``fairness'' constraint $\bar{y}=u$
converts minimization of the cross-entropy loss over all pseudo-labels $y$ 
into a non-trivial integer programming problem that can be approximately solved via {\em optimal transport} \cite{cuturi2013sinkhorn}.

Self-labeling methods for entropy clustering can also use ``soft'' 
pseudo-labels $y\in\Delta^K$ as targets in $H(y,\sigma)$. In general,
soft target distributions $y$ are common
in the context of noisy labels \cite{tanaka2018joint,song2022learning} and network calibration \cite{guo2017calibration,muller2019does}. They can also improve generalization by reducing over-confidence \cite{pereyra2017regularizing}.
In the context of entropy clustering, soft pseudo-labels $y$ correspond to
a {\em relaxation} of cross-entropy. Nevertheless, cross-entropy encourages decisive pseudo-labels $y$ since $H(y,\sigma)\equiv H(y)+KL(y\|\sigma)$. 
The last term also implies proximity $\sigma \approx y$ and, therefore, 
the decisiveness of predictions $\sigma$. 
Many soft self-labeling methods \cite{ghasedi2017deep,ismail2021} represent the fairness constraint using $-H(\bar{y})$ or $KL(\Bar{y}\,\|\,u)$, as in \eqref{eq:mi+decay}. 
In particular, \cite{ismail2021} formulates the following entropy-based  soft self-labeling loss
\begin{align} \label{eq:ismail}
    L_{ce+kl} \;\;= &\;\;\;\;\;\;\overline{H(y,\sigma)} \;\;\;\;\;- \;\; H(\bar{y})     
\end{align}
representing the decisiveness and the fairness constraints.
Similarly to \eqref{eq:vidaldi}, the network parameters in \eqref{eq:ismail} are trained by the standard cross-entropy loss $H(y,\sigma)$ when $y$ are fixed.  Optimization over the relaxed pseudo-labels $y\in\Delta^K$ is 
relatively easy since negative entropy is convex and cross-entropy is linear w.r.t. $y$. 
While there is no closed-form solution, the authors offer an efficient approximate solver.

\subsection{Summary of our contributions} \label{sec:contributions}

This paper provides new theories and algorithms for discriminative entropy clustering. 
First, we examine its conceptual relations to K-means and SVM. We disprove the theories 
in \cite{ismail2021} on the equivalence between the soft K-means and the linear case of entropy clustering
(\ref{eq:mi},\ref{eq:postmodel_shallow}). 
Figures \ref{fig:teaser}, \ref{fig:minima}, \ref{fig:gamma}, \ref{fig:multi-label} 
study a counterexample and Section \ref{sec:Kmeans} points out specific technical errors in their proof.
Despite the equivalence \eqref{eq:mi_generative} $\equiv$ \eqref{eq:mi_discriminative}, 
two discriminative and generative clustering algorithms can operate within different hypothesis spaces even 
if both produce linear results. Further contradicting equivalence to K-means, 
our theories in Section \ref{sec:SVM} prove that linear entropy clustering (\ref{eq:mi},\ref{eq:postmodel_shallow}) 
has a {\em margin-maximizing} property establishing a formal relation to SVM-based clustering \cite{vapnik:2001,Schuurmans2004}.
In the general context of entropy clustering \eqref{eq:mi} with deep models \eqref{eq:postmodel_deep}, 
our results imply margin maximization in the {\em embedding space}, similar to kernel SVM \cite{cristianini2000introduction}. 
Such non-linear methodologies, however, can not use arbitrary embeddings. Indeed, SVM should
restrict kernels (implicit embeddings) and networks should regularize (explicit) embedding functions $f_\wf(X)$.

Second, our Section \ref{sec:our_approach} proposes a new self-labeling loss and algorithm for 
general entropy clustering (\ref{eq:mi},\ref{eq:postmodel_deep}). 
We derive a surrogate for \eqref{eq:mi} with numerically important differences from \eqref{eq:vidaldi}, \eqref{eq:ismail}.
Assuming relaxed pseudo-labels $y$, we observe that the standard formulation 
of decisiveness $\overline{H(y,\sigma)}$ is sensitive to uncertainty/errors natural for pseudo-labels. 
We demonstrate that our {\em reverse cross-entropy} is significantly more 
robust to the label noise. We also propose a {\em zero-avoiding} form of KL-divergence as a stronger fairness term that  
does not tolerate trivial clusters, unlike the standard fairness in \eqref{eq:ismail}. 
Our new self-labeling loss is convex w.r.t $y$ and allows 
an efficient EM solver for pseudo-labels with closed-form E and M steps.
The new algorithm improves the state-of-the-art on many standard benchmarks 
for deep clustering, as shown in Section \ref{sec:experiments} empirically confirming our technical insights.

\section{Linear entropy clustering} 
\label{sec:relations}

This Section analyses the theoretical properties of entropy clustering loss \eqref{eq:mi}
in the context of linear discriminative models \eqref{eq:postmodel_shallow}. Even such simple models 
may require some regularization to avoid degenerate clusters. Section \ref{sec:SVM} shows a form of regularization implying
{\em margin maximization}. But first, Section \ref{sec:Kmeans} juxtaposes (\ref{eq:mi},\ref{eq:postmodel_shallow}) 
and K-means as representative {\em linear} cases of discriminative and generative clustering 
(\ref{eq:mi_generative},\ref{eq:mi_discriminative}).

\subsection{Relation to K-means}
\label{sec:Kmeans}

There are many similarities between entropy clustering \eqref{eq:mi} with linear model
\eqref{eq:postmodel_shallow} and K-means. Both produce linear boundaries 
and use (nearly) the same number of parameters, $K\times (M+1)$ vs. $K\times M$. 
The former corresponds to $K$ linear discriminants $\wc_k$ (w. bias) forming the columns of matrix $\wc$ in \eqref{eq:postmodel_shallow}, and the latter
corresponds to $K$ means $\mu_k$ representing density models at each cluster.
Both approaches have good approximation algorithms for their non-convex \eqref{eq:mi} and NP-hard \cite{mahajan2012planar} objectives.
Two methods also generalize to non-linear clustering using more complex representations, 
e.g. learned $f_w(X)$ or implicit (kernel K-means). 

There is a limited understanding of the differences between linear entropy clustering and 
K-means as most prior literature, including \cite{MacKay1991}, 
discusses \eqref{eq:mi} in the context of networks \eqref{eq:postmodel_deep}.
One 2D linear example in \cite{Perona2010} (Fig.1) helps, but unlike our Figure \ref{fig:teaser}, 
they employ trivial compact clusters typical for the textbook's illustrations of K-means. The two methods
are indistinguishable from the example in \cite{Perona2010}.
Moreover, there is a prior theoretical claim \cite{ismail2021} about equivalence 
between soft K-means and linear entropy clustering, assuming certain regularization. 
We disprove this claim later in this Section.

Generative and discriminative formulations of MI (\ref{eq:mi_generative},\ref{eq:mi_discriminative}) may also suggest the equivalence of linear entropy clustering and K-means. We already discussed how \eqref{eq:mi} relates to \eqref{eq:mi_discriminative}, and now we show how K-means relates to \eqref{eq:mi_generative}. 
Indeed, the Lloyd's algorithm for K-means minimizes the following self-labeling objective for hard pseudo-labels $y\in\Delta^K_{0,1}$ and parameters $\mu_k$ 
\begin{eqnarray}
  L_{k\mu} & := & \;\;\sum_k\sum_i y_i^k \|X_i - \mu_k\|^2  \label{eq:KM}  \\
            & \eqc &  -\sum_k\sum_i y_i^k\ln N_{\mu_k}(X_i) \nonumber \\
            & \approx & \;\;\; \sum_k \,|X^k|\; H(X^k,N_{\mu_k}) \nonumber 
\end{eqnarray}
where each point $X_i$ contributes a squared distance to the
mean $\mu_k$ of the assigned cluster such that $y_i^k=1$.
Isotropic Gaussian densities $N_{\mu}$ with covariances $\Sigma_k=I/2$ allow an equivalent formulation on the second line 
below \eqref{eq:KM}. 
Its Monte Carlo approximation on the third line produces an expression with cross-entropy where
$X^k$ represents the ``true'' density of data in cluster $k$ and $|X^k|$ is its cardinality. 
The standard relationship between cross-entropy and entropy functions further implies an inequality
$$\sum_k |X^k|\; H(X^k,N_{\mu_k}) \;\;\geq\;\; 
\sum_k |X^k|\; H(X^k)\;\;\propto\;\;H(X|C)$$
concerning the conditional entropy in \eqref{eq:mi_generative}. The last $\propto$ relation
only ignores a constant factor, the whole dataset cardinality. 
When recomputing cluster means, Lloyd's algorithm minimizes the cross-entropy above. It
achieves its low bound, the entropy in the second expression, only when the clusters are isotropic Gaussian blobs, which
is an implicit assumption in K-means. This is when K-means works well and, as we just showed, 
when it approximates generative MI clustering \eqref{eq:mi_generative}. 
Thus, K-means and linear entropy clustering are equivalent in the case of 
isotropic Gaussian clusters. This equivalence is consistent with the toy example in \cite{Perona2010}.

The arguments above also suggest how the equivalence may fail for more complex clusters.
Indeed, our Figure \ref{fig:teaser} demonstrates no equivalence without isotropy. We can also refute 
the general claim in \cite{ismail2021} about the equivalence between the following {\em soft} variant of 
K-means loss \eqref{eq:KM} 
\begin{equation} \label{eq:sKM} 
   L_{sk\mu} :=   \overline{\overline{y \|X - \mu\|^2}} \;-\; \gamma \overline{H(y)} \;-\; \overline{\|X\|^2}   
\end{equation}
and the linear case of their self-labeling entropy clustering formulation (\ref{eq:ismail},\ref{eq:postmodel_shallow}) 
with linear classifier's norm regularization
\begin{equation} \label{eq:rec_ismail}
    L_{ce+} :=  \overline{H(y,\sigma)} \;-\;  H(\overline{y}) \;+\; \gamma \|\wc\|^2 
\end{equation}
where the {\em single bar} operator averaging over data points $i$ was introduced in Section \ref{sec:entropy clustering}, 
and the {\em double bar} represents averaging over both $i$ and $k$ shortening the expression for 
loss $L_{k\mu}$ in \eqref{eq:KM}. The negative entropy in \eqref{eq:sKM} 
encourages soft labeling, i.e. $y_i\in\Delta^K$ could be any categorical distribution.
This term is standard for {\em soft K-means} formulations. 
The last term in \eqref{eq:sKM} is a constant needed in the equivalence claim \cite{ismail2021}. 

\begin{figure}
\centering
\resizebox{0.48\textwidth}{!}{\begin{tabular}{c|cc}
& \multicolumn{2}{c}{$\gamma \leq 0.00001$} \\[1ex] \hline
    \rotatebox{90}{\parbox[c]{4.8cm}{\centering {\small\bf \hspace{-4ex} linear entropy clustering}  (\ref{eq:rec},\ref{eq:postmodel_shallow})}} &
    \includegraphics[height=4.8cm]{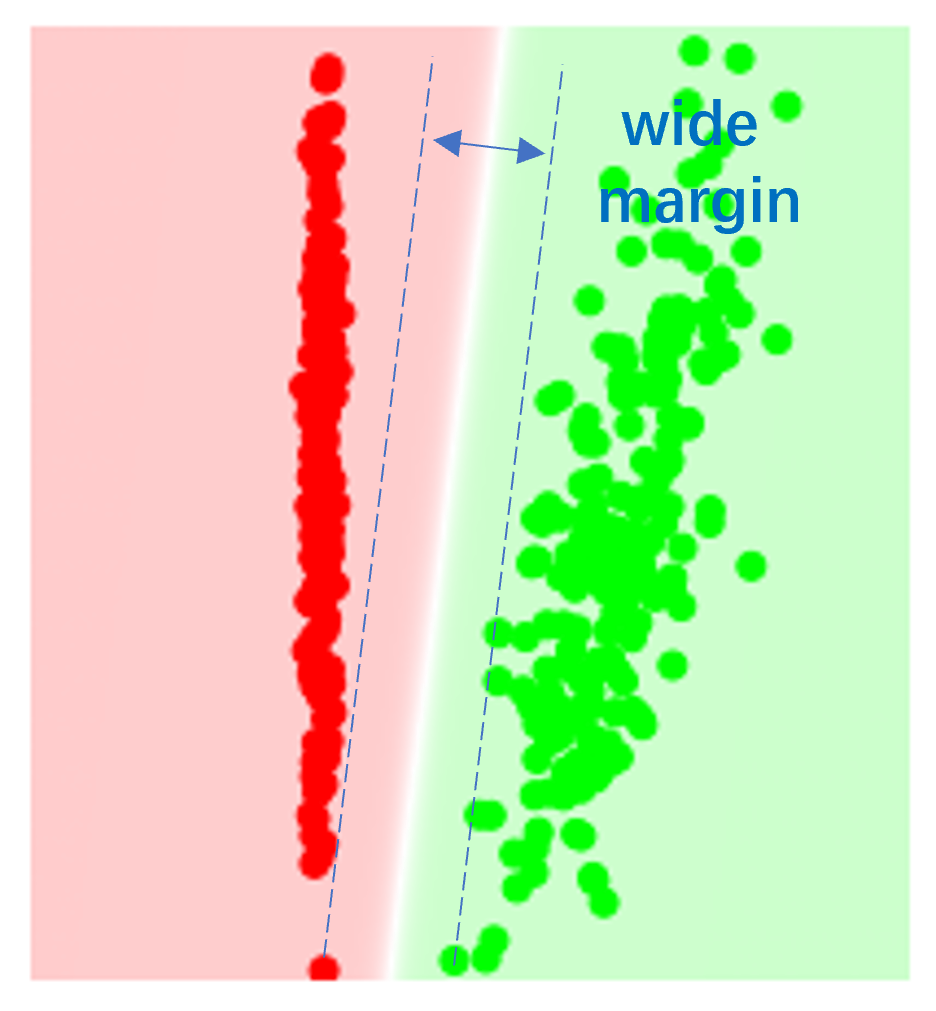} &  
    \includegraphics[height=4.8cm]{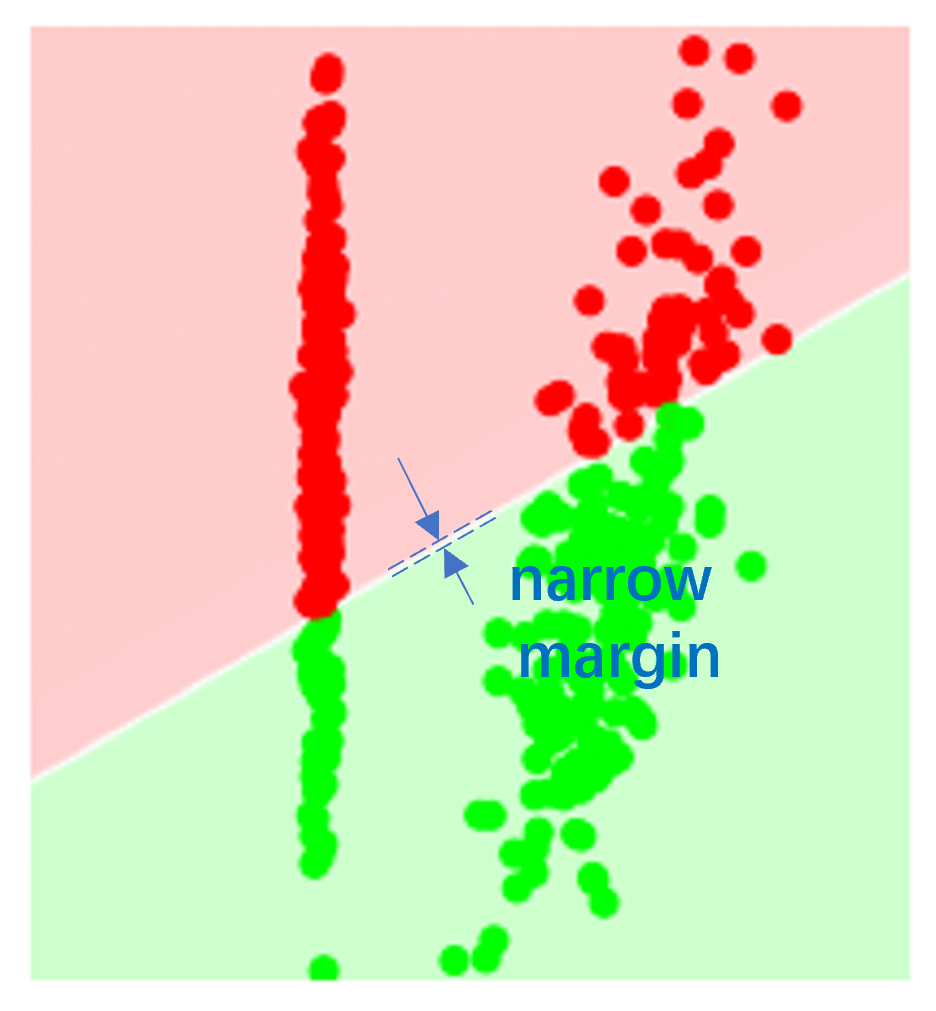} \\[-1ex]
    &
     (a) \hspace{1ex} rEC loss = {\bf -0.6910} 
     & (b) \hspace{1ex} rEC loss = -0.6610 \\[1ex] \hline
            \rotatebox{90}{\parbox[c]{4.8cm}{\centering {\small\bf soft K-means} \eqref{eq:sKM} }} &
     \includegraphics[height=4.8cm]{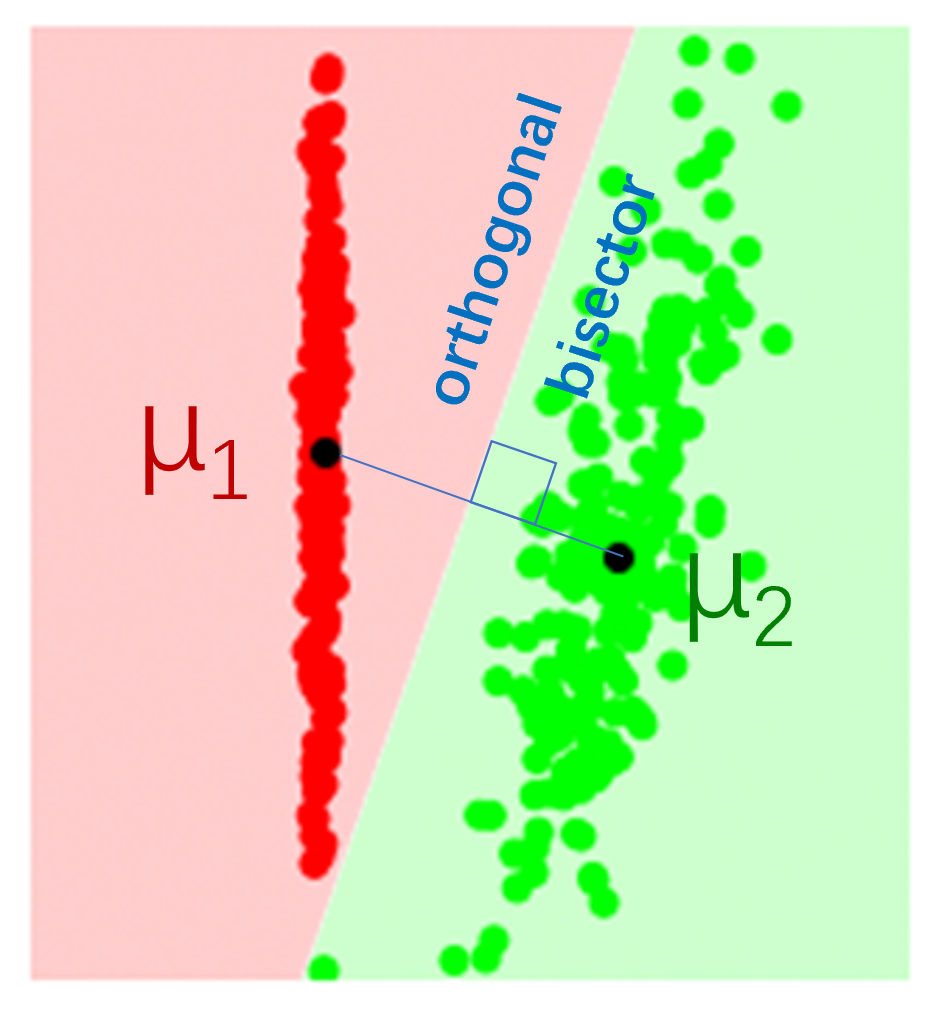} &  
     \includegraphics[height=4.8cm]{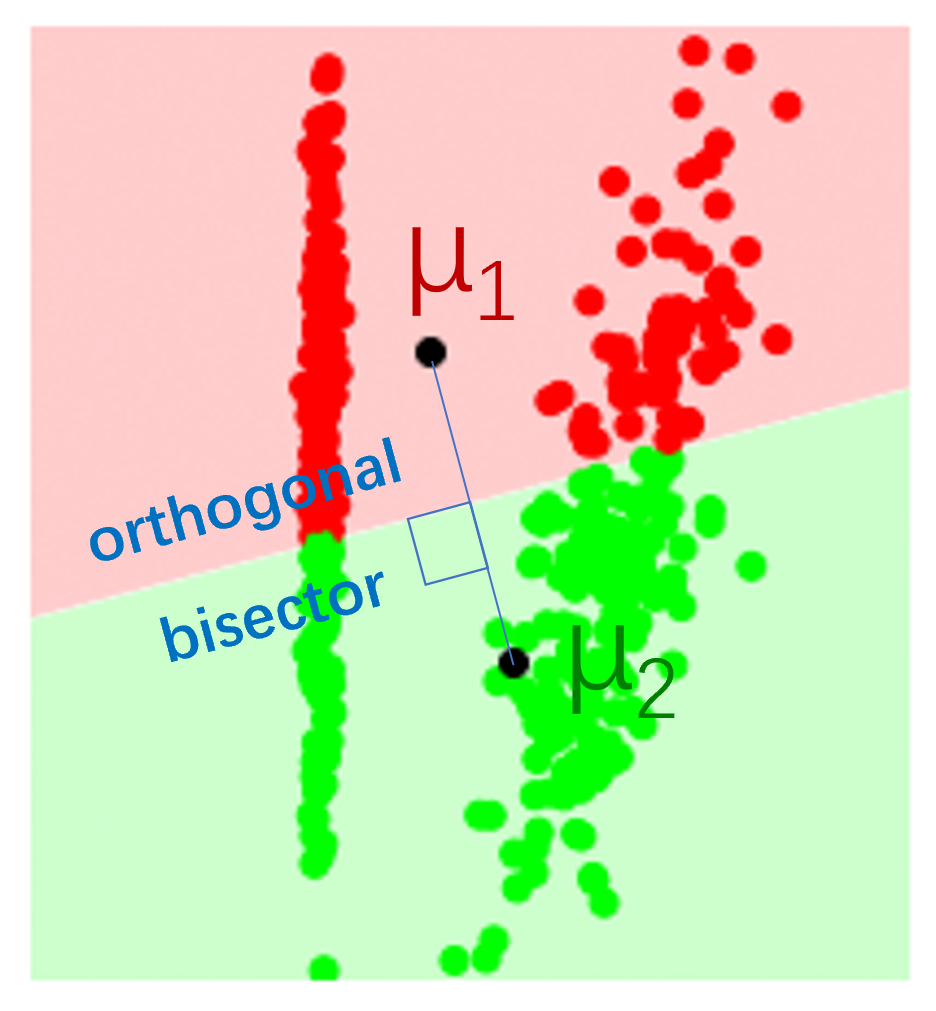} \\[-1ex]
    &
     (c) \hspace{1ex} sKM loss = 419.60 
     & (d) \hspace{1ex} sKM loss = {\bf 401.45} \\[1ex] \hline
\end{tabular}}
\caption{Global \& local minima: linear {\em regularized entropy clustering} (rEC) versus {\em soft K-means} (sKM).
For both losses, global optima (a) and (d) are consistent for all $\gamma\in(0,0.00001]$. Variations in the optimal loss values 
are negligible. sKM is nearly identical to hard K-means for such  $\gamma$; it softens only for larger $\gamma$, 
see Figure \ref{fig:gamma}. The local minimum for sKM (c) is obtained by Lloyd's algorithm initialized at (a). 
Vice-versa, gradient descent for rEC converges to (a) from (c).
The same ``cross-check'' works for (b) and (d). Local minima for rEC (a,b) are balanced clusterings with (locally) maximum margins. 
In contrast, local minima for sKM (c,d) are {\em orthogonal bisectors} for the cluster centers. K-means ignores the margins.} 
\label{fig:minima}
\end{figure}

\begin{figure*}
    \centering
    \begin{tabular}{c|c|c|c|c}
    & $\gamma_1 = 0.00001$ & $\gamma_2 = 0.005$ & $\gamma_3 = 0.01$ & $\gamma_4 = 0.15$ \\ \hline
        \rotatebox{90}{\parbox[c]{3.1cm}{\centering {\bf linear \\ entropy clustering} \\ (\ref{eq:rec},\ref{eq:postmodel_shallow}) }} &
        \includegraphics[height=3.0cm]{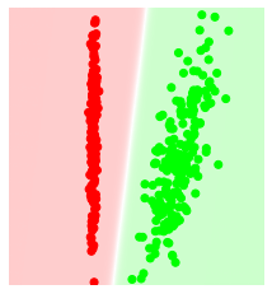} \vspace{1ex} &  
        \includegraphics[height=3.0cm]{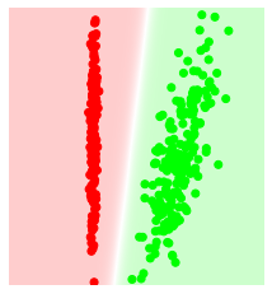} &  
        \includegraphics[height=3.0cm]{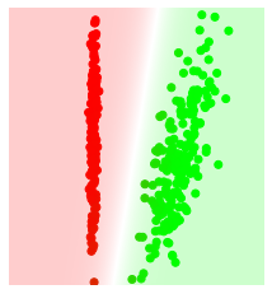} &  
        \includegraphics[height=3.0cm]{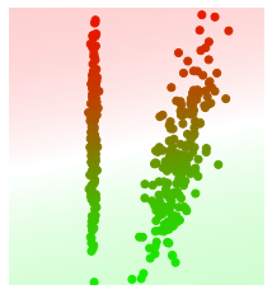} \\[-1ex] \hline
                \rotatebox{90}{\parbox[c]{3.1cm}{\centering {\bf soft \\ K-means} \\ \eqref{eq:sKM} }} &
         \includegraphics[height=3.0cm]{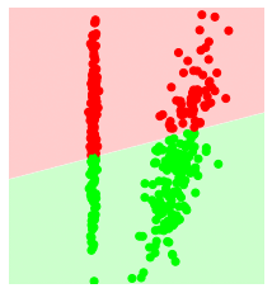} \vspace{1ex} &  
         \includegraphics[height=3.0cm]{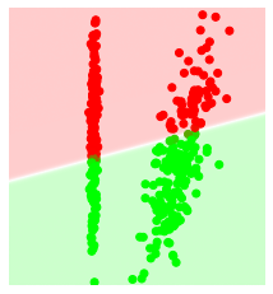} &  
         \includegraphics[height=3.0cm]{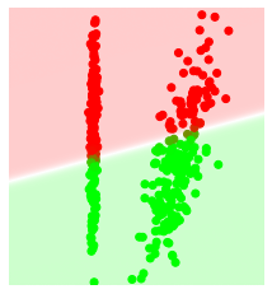} &  
         \includegraphics[height=3.0cm]{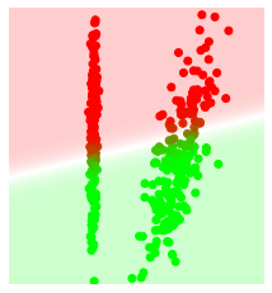} \\[-1ex] \hline
    \end{tabular}
    \caption{Global minima for various $\gamma$: linear {\em regularized entropy clustering} (rEC) versus {\em soft K-means} (sKM). 
    As $\gamma\rightarrow 0$, both approaches converge to hard, but different, clusters. 
    Optimal/low-variance sKM clusters are consistent for all $\gamma$.
    In contrast, rEC produces max-margin clusters for small $\gamma$ and changes the solution for larger $\gamma_4$. 
    The latter reduces norm $\|\wc\|$ implying a wider ``indecisiveness'' zone around the linear decision boundary.
    Due to the decisiveness term in \eqref{eq:mi}, entropy clustering finds the boundary minimizing the overlap 
    between the data and such ``softness'' zone, explaining the result for $\gamma_4$. } 
    \label{fig:gamma}
\end{figure*}

The only difference between the entropy clustering losses \eqref{eq:ismail} and \eqref{eq:rec_ismail} is the squared norm
of the linear classifier parameters $\|\wc\|$, excluding the bias. This standard regularization encourages ``softness'' 
of the linear classifier's soft-max predictions, similar to the effect of the entropy term in soft K-means \eqref{eq:sKM}. 
However, unlike entropy clustering\footnote{The entropy clustering results in the toy examples in Figs. \ref{fig:teaser}, \ref{fig:minima}, \ref{fig:gamma}, \ref{fig:multi-label} do not depend on the specific formulation.
Self-labeling algorithm \cite{ismail2021} for \eqref{eq:rec_ismail} and our entropy clustering algorithm 
in Sec.\ref{sec:our_approach} produce the same results. 
For simplicity, these Figures use basic gradient descent for the regularized version of generic 
linear entropy clustering (\ref{eq:mi},\ref{eq:postmodel_shallow}), as in \eqref{eq:rec}.}, soft K-means
fails on elongated clusters in Figures \ref{fig:minima} and \ref{fig:gamma} in the same way as 
the basic K-means in Figure \ref{fig:teaser}. Indeed, soft formulations of K-means are typically 
motivated by a different problem - overlapping clusters. The proper generative mechanism to address 
anisotropic clusters is to drop the constraint on the covariance matrices $\Sigma_k\sim I$ 
allowing a wider class of Gaussian density models, i.e. extending the {\em hypothesis space}.
For example, GMM can be viewed as an anisotropic extension of soft K-means, and it would work perfectly 
in Figures \ref{fig:teaser}, \ref{fig:minima}, \ref{fig:gamma}, \ref{fig:multi-label}, similar to entropy clustering. 

We also have a general argument for why linear entropy clustering is stronger than K-means. 
As easy to check using Bayes formula, the posterior for isotropic Gaussian densities estimated by K-means 
is consistent with model \eqref{eq:postmodel_shallow}. However, discriminative entropy clustering optimizes 
\eqref{eq:postmodel_shallow} without any assumptions on densities, implying a larger hypothesis space.

Besides our counterexample in Figure \ref{fig:minima} and the general argument above, we found a critical error in \cite{ismail2021}. They ignore the normalization/denominator in the definition of softmax. Symbol $\propto$ hides it in their equation (5), which is treated as equality in the proof of Proposition 2.
Their proof does not work with the normalization, which is important for training softmax models. 
The regularization term $\|\wc\|^2$ and constant $\|X\|^2$ play the following algebraic role in their proof of 
equivalence between \eqref{eq:rec_ismail} and \eqref{eq:sKM}. Without softmax normalization, $\ln\sigma$ inside 
cross-entropy $H(y,\sigma)$ in \eqref{eq:rec_ismail} turns into a linear term w.r.t. logits $\wc^\top X$ and combining it 
with $\|\wc\|^2$ and $\|X\|^2$ creates a quadratic form  $\|X-\wc\|^2$ resembling squared errors in K-means \eqref{eq:sKM}. 
In contrast, Section \ref{sec:SVM} shows that regularization $\|\wc\|^2$ in \eqref{eq:rec_ismail} is needed for
the {\em margin maximization} property of entropy clustering illustrated in Figures \ref{fig:minima}(a) and 
\ref{fig:multi-label}(b). Our theories show that instead of K-means, in general, 
linear entropy clustering is related to discriminative SVM clustering \cite{vapnik:2001,Schuurmans2004}.

\subsection{Margin maximization} 
\label{sec:SVM}

This section establishes a {\em margin-maximizing} property of the regularized decisiveness
for linear model \eqref{eq:postmodel_shallow}
\begin{equation} \label{eq:reg_decisiveness}
\gamma \|\wc\|^2\;+\; \overline{H(\sigma)}
\end{equation}
where $\|\wc\|$ is $L_2$ norm of the linear discriminant w/o the bias.
Without extra constraints, the decisiveness allows a trivial solution with a single cluster. 
This can be avoided by focusing on a set of {\em feasible} solutions, where
feasibility can represent balanced clusters or 
consistency with partial data labels as in {\em semi-supervised learning}.

Our theory assumes an arbitrary set of feasible solutions but is also relevant for soft feasibility. 
For example, the regularized entropy clustering loss for linear model \eqref{eq:postmodel_shallow}
\begin{equation} \label{eq:rec}
    L_{rec} \;\;:=\;\;  \gamma \|\wc\|^2 \;+\; \overline{H(\sigma)} \;-\;  H(\overline{\sigma})  
\end{equation}
combines the margin-maximizing decisiveness with the fairness term 
encouraging balanced clusters, see Fig. \ref{fig:minima}(a). 
The direct relation to MI \cite{MacKay1991} provides an information-theoretic motivation for this loss, 
as discussed in Section \ref{sec:entropy clustering}.
Our theories below extend the general conceptual understanding of the decisiveness, 
entropy clustering, and establish their relation to unsupervised SVM methods
\cite{vapnik:2001,Schuurmans2004}.

\subsubsection{Overview of max-margin clustering}

As typical in SVM, our formal theories on the max-margin property of the regularized decisiveness \eqref{eq:reg_decisiveness} 
in Sections \ref{sec:Rinfty}-\ref{sec:Ralpha} are focused on the binary linear models, but some multi-class extensions are discussed in Sec. \ref{sec:Ralpha_K}. 
Below we informally preview our max-margin property 
claims about \eqref{eq:reg_decisiveness} providing some intuition that may be helpful due 
to differences with standard margin maximization in SVM. 

For example, basic SVM assumes {\em linearly separable} fully-labeled data. {\em Soft SVM} extension allows a trade-off between margin maximization and linear-separation violations.
In clustering, however, any data can be {\em linearly partitioned} in many different ways as there are no ground-truth labels to be violated. The ``gap'' can be measured for any pair of such clusters, 
see Figs. \ref{fig:minima}(a,b).
Thus,  the largest margin solution among all feasible (e.g. fair) linear clusterings 
is well-defined. 

SVM clustering \cite{Schuurmans2004} uses the {\em hinge loss} instead of decisiveness in \eqref{eq:reg_decisiveness} and
also assumes fairness, see Sec.\ref{sec:EC_vs_SVM}. 
In both cases, however, linear classifier regularization $\gamma\|\wc\|^2$ is important for margin maximization. 
The regularized hinge loss outputs a max-margin solution for all $\gamma$ below some threshold.
In contrast, our theories show that optimal solutions for the regularized decisiveness \eqref{eq:reg_decisiveness} 
converge to a max-margin clustering only as $\gamma\rightarrow 0$, see Figure \ref{fig:gamma}, 
though this subtle difference may be hard to discern in practice.

Our results are based on the max-margin theories for supervised losses in \cite{rosset2003margin}. We generalize them
to unsupervised clustering problems. Also, instead of \eqref{eq:reg_decisiveness}, we prove the max-margin property for
a larger class of regularized decisiveness measures using {\em Renyi} entropy $R_\alpha$ 
\cite{Renyi1961} of any order $\alpha>0$ (formally defined in the following Sections)
\begin{equation} \label{eq:reg_decisiveness_R}
\gamma \|\wc\|^2\;+\; \overline{R_\alpha(\sigma)}
\end{equation}
where Shannon's entropy in \eqref{eq:reg_decisiveness} is a special case $H=R_1$.
Focusing on binary clustering, Section \ref{sec:Rinfty} establishes our results for $\alpha=\infty$.
In this base case, Renyi entropy $R_\infty$ can be seen as a ``self-labeled'' {\em logistic loss} allowing us to prove our result in Theorem \ref{th:Rinfty} by extending the standard max-margin property for logistic regression \cite{rosset2003margin}. Theorem \ref{th:Ralpha} in Section \ref{sec:Ralpha} 
extends our results for \eqref{eq:reg_decisiveness_R} to all $\alpha>0$, including \eqref{eq:reg_decisiveness} as a special case. Then, Section \ref{sec:Ralpha_K} discusses 
the multi-class case $K>2$. Finally, Section \ref{sec:EC_vs_SVM} summarizes 
margin maximization by regularized entropy \eqref{eq:reg_decisiveness_R}
and juxtaposes it with the regularized hinge loss in SVM clustering \cite{Schuurmans2004}.

\subsubsection{Terminology and Renyi decisiveness for $K=2$}

We extend the earlier notation, in part to suit the binary case. 
For example, besides {\em softmax} output $\softmax$, we also use a scalar 
{\em sigmoid} function $\sigmoid$. Below, we summarize our notation. 
In particular, there is one significant modification specific to this theoretical Section \ref{sec:SVM}. 
Here we use only hard class indicators, thus it is convenient to have integer labels $y$ 
instead of categorical distributions, as in the rest of the paper. 
This change should not cause ambiguity as labels $y$ appear mainly 
in the proofs and their type is clear from the context. 

Our notation concerning labels $y$, labelings $\y$, and linear classifiers $\wc$ is 
mostly general. 
Some parts specific to $K=2$ are designed to transition smoothly 
to multi-class problems. Note that for $K>2$ linear classifiers typically generate 
$K$ logits $\wc^\top X\in R^K$, output a $K$-categorical distribution using softmax
$\softmax(\wc^\top X)\in\Delta^K$, and use natural numbers as class labels. 
In contrast, binary classifiers typically use a single logit, 
{\em sigmoid} output, and class labels $\{\pm 1\}$ or $\{0,1\}$. 
Our notation for $K=2$ rectifies the differences using natural class indicators $\{1,2\}$
and categorical distribution output $\softmax=(\softmax^1,\softmax^2)=(\sigmoid,1-\sigmoid)$.
The latter is justified by a simple property that a single-logit sigmoid classifier 
is equivalent to a softmax  for two logits
as only their difference matters.

\begin{itemize}
\item $\wc$ - linear classifier is a vector for $K=2$. It generates scalar raw output
$\wc^\top X$. Vector $\wc$ includes the {\em bias} as we assume 
``homogeneous'' data representation $X$.
\item $\|\wc\|$ - $L_2$ norm that excludes the bias. 
\item $\softmax = (\softmax^1,\softmax^2) := (\sigmoid,1-\sigmoid)$ - soft-max for $K=2$
can be represented using a scalar {\em sigmoid} function 
$$\sigmoid(x)\;:=\;(1+e^{-x})^{-1} \;\equiv\; \frac{e^{x/2}}{e^{x/2} + e^{-x/2}}\;\in\;[0,1]$$
for $x=\wc^\top X$. 
Similarly to $\sigma$, we use symbol $\sigmoid$ both for specific values in $[0,1]$ 
and as a function name. 
\item $y \in\{1,2\}$ - binary class indicator/label
\item $y^\softmax$ - hard class label produced by soft prediction $\softmax$ 
$$y^\softmax \;:=\;  \arg\max_k  \softmax^k \quad \text{(a.k.a. ``{\em hard-max}'')}$$
\item $y_X$ - class label at arbitrary data point $X$ 
\item $y_i = y_{X_i}$ - class label at point $X_i$ 
\item $\y:=\{y_i\}_{i=1}^N$ - labeling of the whole dataset $\{X_i\}_{i=1}^N$
\item $y^\wc_X:= y^{\softmax(\wc^\top X)}$ - hard class label produced by classifier $\wc$ 
for an arbitrary data point $X$.
\item $y^\wc_i := y^\wc_{X_i}$ - hard label produced by $\wc$ for point $X_i$. 
\item $\y^\wc := \{y_i^\wc \}_{i=1}^N$ - dataset labeling by classifier $\wc$
\item ${\bf V}^\y:= \{\wc\,|\,\y^\wc=\y\}$ - a set of all linear classifiers consistent with given labeling $\y$.
By default, this paper uses ``weak consistency'' allowing data points on the decision boundary, i.e. zero margins.
Set ${\bf V}^\y$ is non-empty only if labeling $\y$ is (weakly) linearly separable.
\item $\FL$ - a set of feasible labelings $\y$ representing allowed clusterings. 
It could be arbitrary. One example is a set of all ``fair'' linearly separable clusterings. 
\item ${\bf V}^\FL := \cup_{\y\in\FL} {\bf V}^\y$ - a set of all linear classifiers $\wc$ consistent with allowed labelings $\y$ in $\FL$. 
\item $\uc$ - unit norm linear classifier such that $\|\uc\|=1$
\item ${\bf U}^\y:= \{\wc\,|\,\y^\wc = \y,\;\|\wc\|=1\}$ - a set of all unit-norm linear classifiers consistent with given labeling $\y$.
\item ${\bf U}^\FL:= \cup_{\y\in\FL} {\bf U}^\y  $ - a set of all unit-norm linear classifiers consistent with labelings in $\FL$.
\item $\uc^\y$ - maximum margin linear classifier of unit norm corresponding to given linearly separable labeling $\y$
\end{itemize}

\begin{figure}
    \centering
    \includegraphics[width=7.0cm]{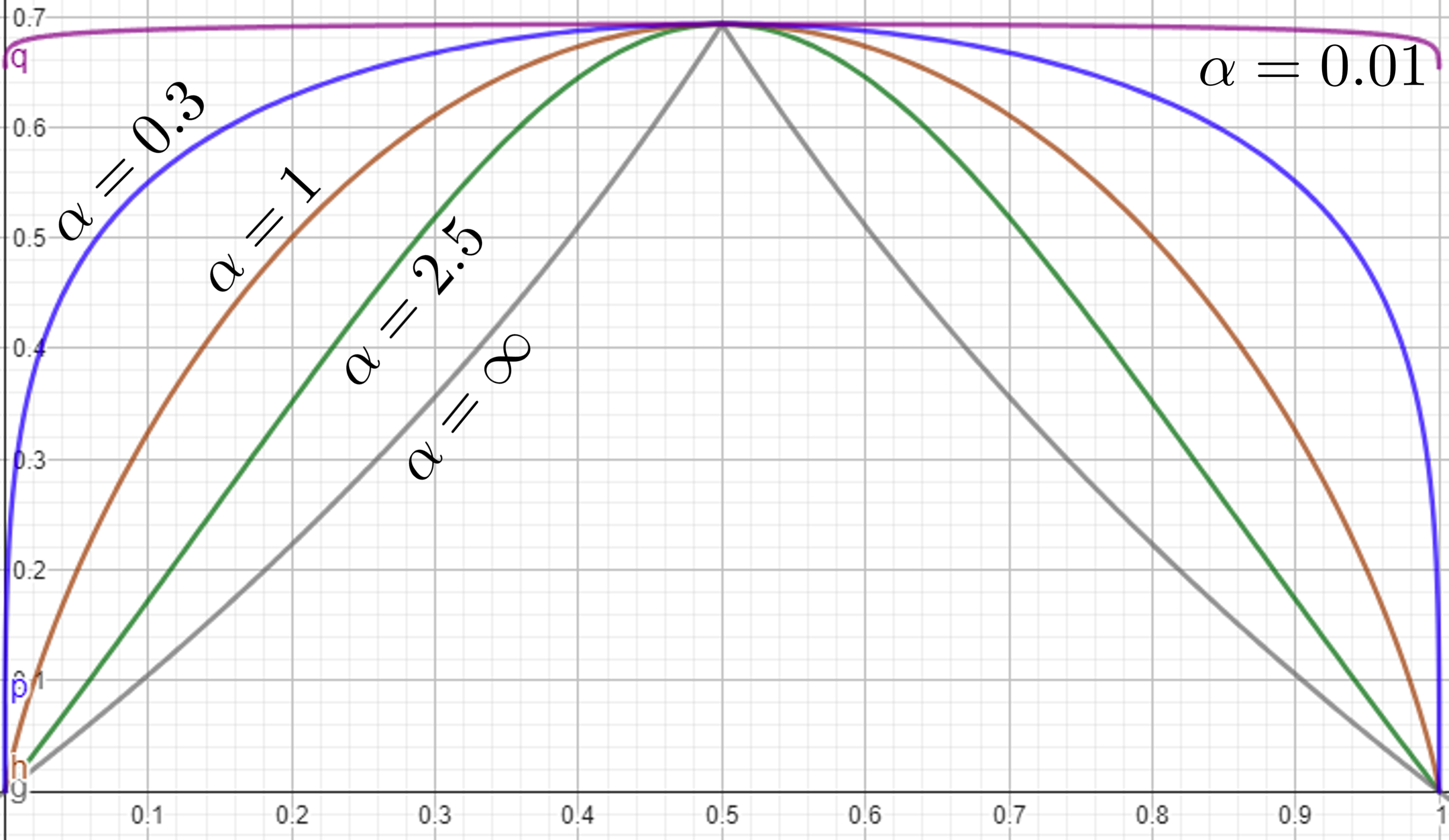}
    \caption{Renyi entropy \cite{Renyi1961} of order $\alpha$: assuming $K=2$ the plots above
    show $R_\alpha(\softmax)$ in \eqref{eq:binary_Ra_def} for binary distributions 
    $\softmax=(\sigmoid,1-\sigmoid)$ as functions of scalar $\sigmoid\in[0,1]$. 
    Shannon entropy \eqref{eq:binary_H_def} is a special case corresponding to $\alpha=1$. 
    \label{fig:renyi_a}}
\end{figure}

Finally, we define the binary version of Renyi entropy $R_\alpha$ \cite{Renyi1961} of order $\alpha\geq 0$ studied by our max-margin clustering theories for $K=2$ in Sec. \ref{sec:Rinfty}- 
\ref{sec:Ralpha}. 
One-to-one correspondence between distributions $\softmax=(\softmax^1,\softmax^2)=(\sigmoid,1-\sigmoid)$ and scalars $\sigmoid$ makes it convenient to define for $K=2$ two equivalent functions $R_\alpha(\softmax)$ and $R_\alpha(\sigmoid)$ 
\begin{equation} \label{eq:binary_Ra_def}
R_\alpha(\softmax) \;\equiv\; R_\alpha(\sigmoid)\;:=\;\frac{\ln(\sigmoid^\alpha + (1-\sigmoid)^\alpha)}{1-\alpha} . 
\end{equation}
The expression in \eqref{eq:binary_Ra_def} is numerically ill-conditioned 
when $\alpha\in\{0,1,\infty\}$. In these three cases, Renyi entropy is better defined by the asymptotically consistent formulations
\begin{eqnarray}
R_0(\softmax) = R_0(\sigmoid) & = &  \ln 2 \cdot [0<\sigmoid<1]    \label{eq:binary_R0_def} \\ 
R_1(\softmax) = R_1(\sigmoid) & = & H(\softmax) \quad \text{\footnotesize (binary Shannon entropy)} \nonumber  \\ 
R_\infty(\softmax) = R_\infty(\sigmoid) & = & -\ln \max\{\sigmoid,1-\sigmoid\}    \label{eq:binary_Rinfty_def}
\end{eqnarray}
where $[\cdot]$ is {\em Iverson bracket} indicator returning $1$ or $0$ depending
if the condition inside is true or not. For $\alpha=1$ one can use 
the standard formula for the binary Shannon entropy 
\begin{equation} \label{eq:binary_H_def} 
H(\softmax)\;\equiv\;H(\sigmoid) \;=\;-\sigmoid\ln\sigmoid - (1-\sigmoid)\ln(1-\sigmoid).
\end{equation}
that has no numerical issues.
Figure \ref{fig:renyi_a} illustrates binary Renyi entropy functions $R_\alpha$ for different orders $\alpha$.

\subsubsection{Max-margin theories for binary $R_\infty$ decisiveness} \label{sec:Rinfty}

Our base case is Renyi entropy $R_\infty$ for $K=2$. 
As evident from \eqref{eq:binary_Rinfty_def}, it resembles the standard 
{\em negative log-likelihood} (NLL) loss for supervised classification, a.k.a. {\em logistic regression}.
To emphasize this relation, we denote supervised NLL loss by $R_\infty(\softmax\,|\,y)$. 
For $\softmax=(\softmax^1,\softmax^2)=(\sigmoid,1-\sigmoid)$ it is
\begin{equation} \label{eq:binary_Rinfty_sup}
R_\infty(\softmax\,|\,y)\;:=\;-\ln \softmax^y \;\equiv\; 
\begin{cases}  -\ln \sigmoid  & \mbox{if}\;y=1 \\ -\ln (1-\sigmoid) & \mbox{if}\;y=2 \end{cases}
\end{equation}
where $y$ is a given binary label at a data point.
The obvious relation between the entropy \eqref{eq:binary_Rinfty_def} and NLL \eqref{eq:binary_Rinfty_sup}, 
see Fig. \ref{fig:renyi_supervised}(a),
\begin{equation} \label{eq:Rinfty_and_NLL}
R_\infty(\softmax) \; = \; \min\left\{\,R_\infty(\softmax\,|\,1) ,\,R_\infty(\softmax\,|\,2)\,\right\}  
\end{equation}
can be equivalently represented as ``self-labeling'' identity 
\begin{equation} \label{eq:self_labeling}
R_\infty(\softmax) \;\; \equiv \;\; R_\infty(\softmax\,|\,y^\softmax) 
\end{equation}
where $y^\softmax$ is the ``hard-max'' label for prediction $\softmax$.

We exploit the ``self-labeling'' identity \eqref{eq:self_labeling}
to prove the max-margin clustering property for entropy $R_\infty$ 
by extending the known margin-maximizing property for logistic regression \cite{rosset2003margin}
reviewed below.

{\bf Max-margin in logistic regression:}
With some exceptions \cite{vapnik:2001,Schuurmans2004}, margin maximization is 
typically discussed in a fully supervised context assuming a given linearly separable labeling.
The max-margin property is easy to prove for the {\em hinge loss} \cite{vapnik:1995} as
it vanishes above some threshold. Some monotone-decreasing losses, 
e.g. {\em logistic loss}, also have max-margin property, but the proof requires careful analysis \cite{rosset2003margin}. 
They prove a sufficient condition, ``fast-enough'' decay, for monotone non-increasing 
classification losses that depend only on distances to classification boundary, a.k.a. {\em margins}.

\begin{figure}
\centering
\begin{tabular}{c@{\extracolsep{2ex}}c}
classification & clustering \\[1ex]
    \centering
    \includegraphics[width=4.0cm]{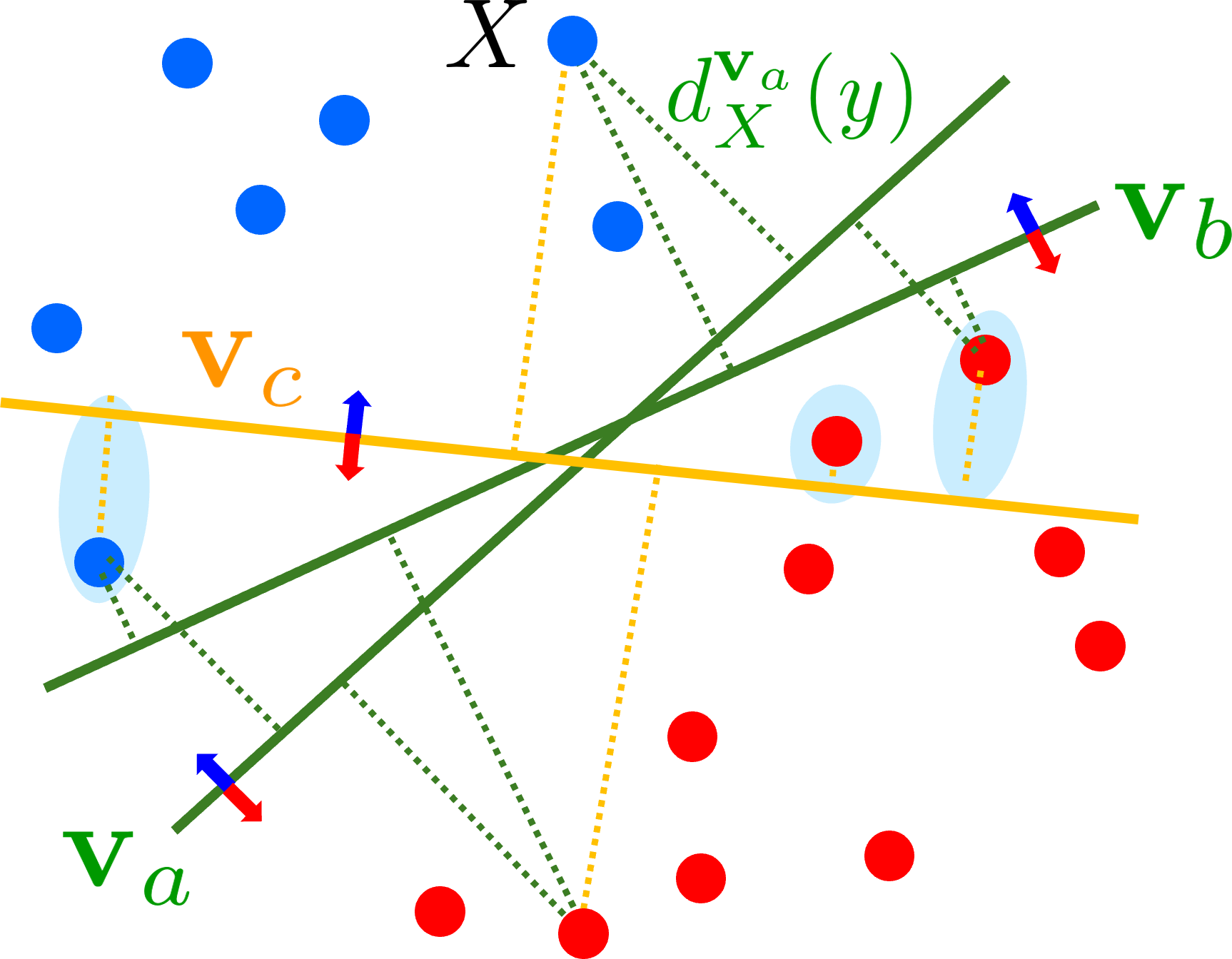}  &
     \includegraphics[width=3.8cm]{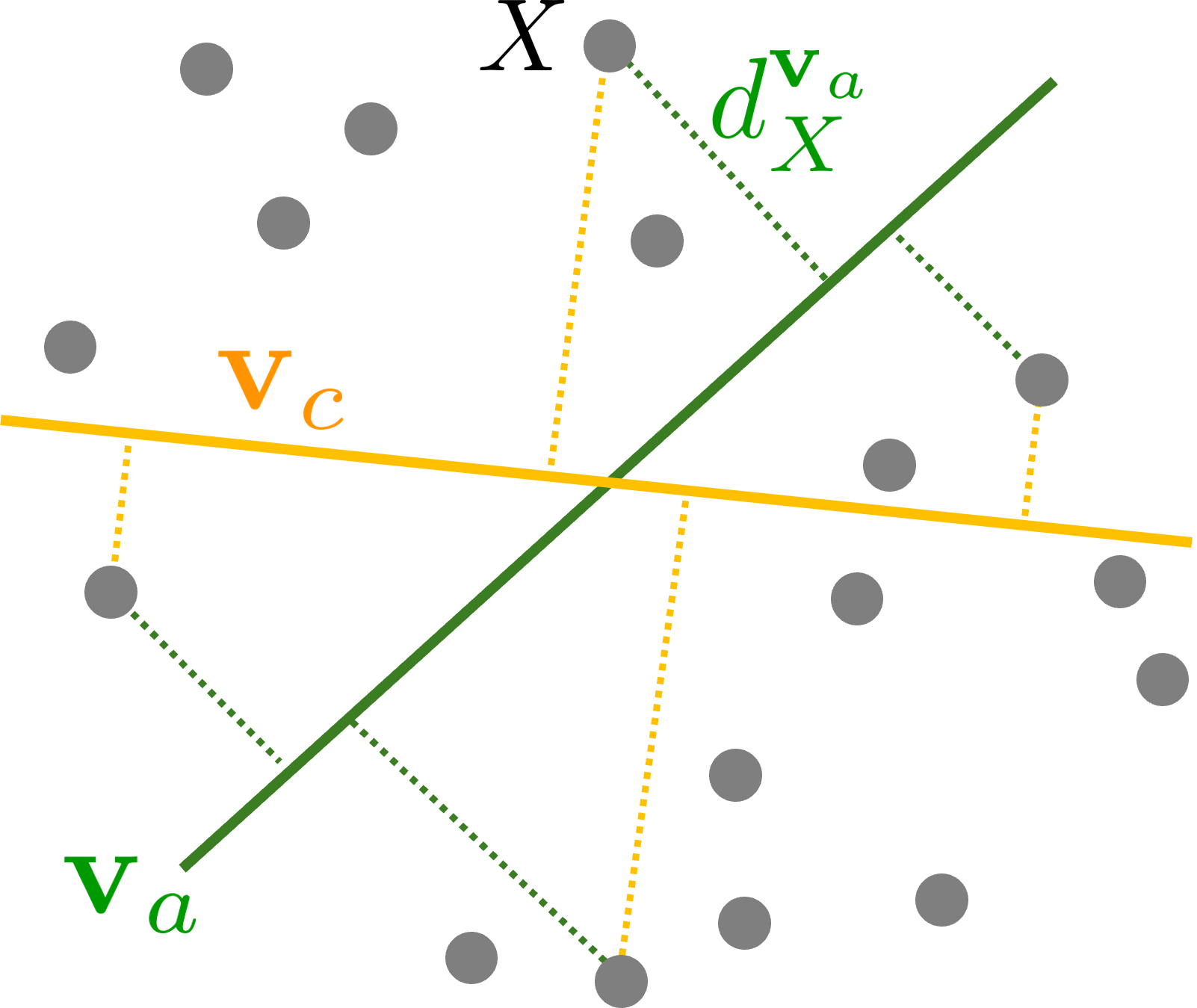}      \\   
    (a) signed margins \eqref{eq:d_signed} &
    (b) unsigned margins \eqref{eq:d_unsigned}
\end{tabular}
\caption{Margins: both examples (a) and (b) use distinct colors for classifiers partitioning the data differently. (a) Point colors show given true labeling $\y$. 
Green classifiers $\wc\in{\bf V}^\y$ are consistent with $\y$ and produce only non-negative margins $d_X^\wc (y)\geq 0$ in \eqref{eq:d_signed}. 
Yellow classifier $\wc_c\not\in{\bf V}^\y$ has inconsistent labeling $\y^{\wc_c}\neq \y$. It has negative (cyan) margins $d_X^{\wc_c}(y)<0$ at incorrectly classified points.
In clustering (b), any classifier (green or yellow) produces only non-negative margins $d_X^\wc(\y^\wc) \equiv d_X^\wc\geq 0$ in \eqref{eq:d_unsigned} w.r.t. its 
labeling/clustering $\y^\wc$. 
\label{fig:margins}}
\end{figure}

\begin{figure*}
\centering
\begin{tabular}{cc}
    \centering
    \includegraphics[width=6.5cm]{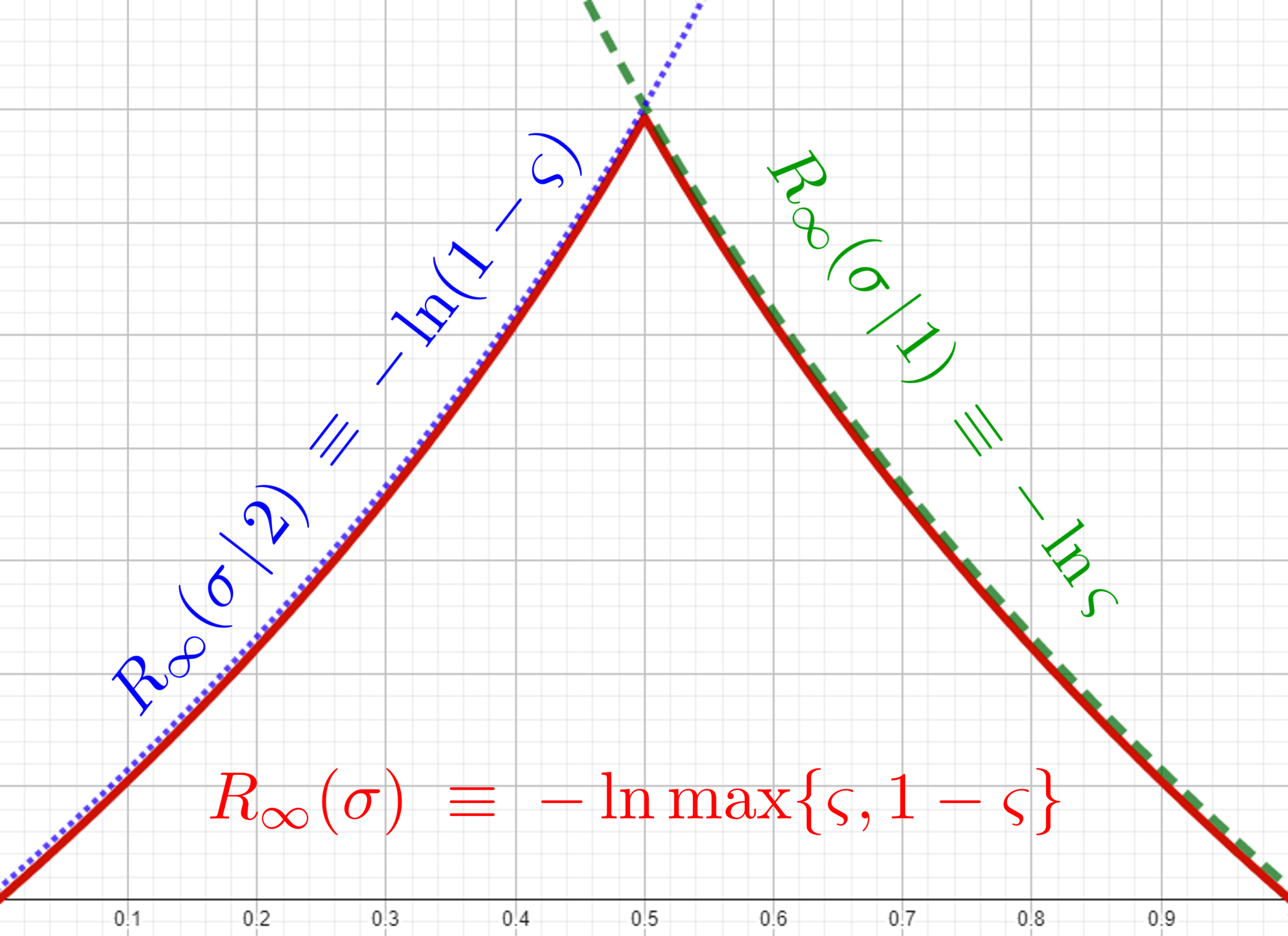}  &
     \includegraphics[width=6.5cm]{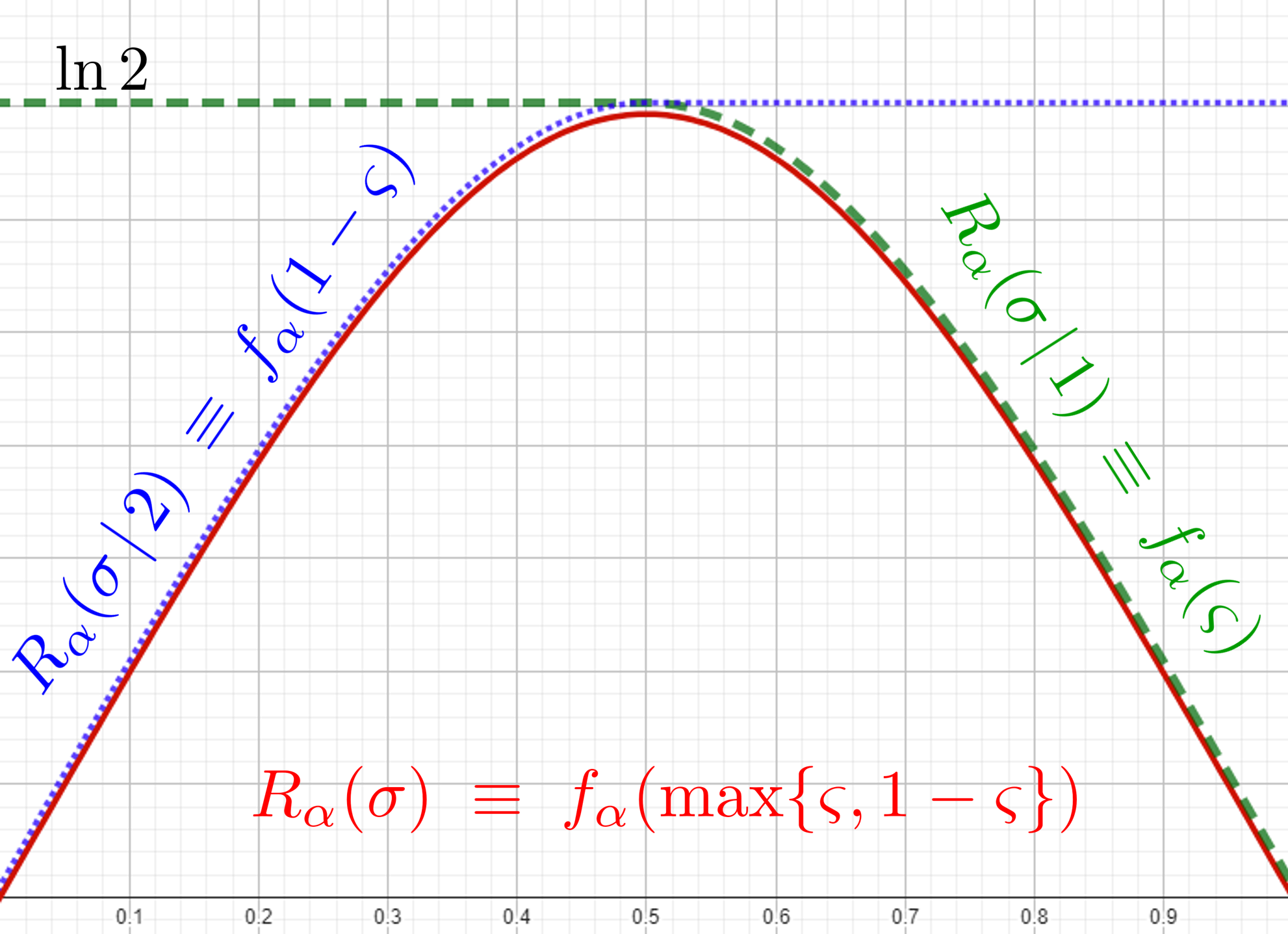}      \\   
    (a) entropy $R_\infty(\softmax)$ and logistic regression $R_\infty(\softmax\,|\,y)$ &
    (b) entropy $R_\alpha(\softmax)$ and supervised loss $R_\alpha(\softmax\,|\,y)$ for $\alpha=2$ 
\end{tabular}
\caption{Renyi entropies $R_\alpha(\softmax)$ and their supervised counterparts $R_\alpha(\softmax\,|\,y)$: 
assuming $K=2$ the plots above visualize these functions defined  in 
\eqref{eq:binary_Ra_def}-\eqref{eq:binary_Rinfty_def} and \eqref{eq:R_a_supervised} 
for binary distributions $\softmax=(\sigmoid,1-\sigmoid)\in\Delta^2$ plotted over interval 
$\sigmoid\in[0,1]$
\label{fig:renyi_supervised}}
\end{figure*}

The standard logistic loss is defined w.r.t. raw classifier output $x=\wc^\top X$ as
\begin{equation} \label{eq:logloss}
r(x)\;\;:=\;\; \ln(1+e^{-x}) \;\;\equiv\;\;   -\ln(\sigmoid(x)) 
\end{equation}
which is a composition of NLL \eqref{eq:binary_Rinfty_sup} and sigmoid function. 
Given true label $y$, the corresponding supervision loss is
\begin{equation} \label{eq:margins_signed}
R_\infty(\softmax(\wc^\top X)\,|\,y)\;\;\equiv\;\;r(d_X^\wc(\y)\|\wc\|) 
\end{equation}
where 
\begin{equation} \label{eq:d_signed}
d_X^\wc(y)\;\;:=\;\;\frac{ \wc^\top X}{\|\wc\| } \cdot [y=1]^{\pm} 
\end{equation}
is a signed {\em margin}, i.e. distance from point $X$ to the classification boundary. 
The operator $[\cdot]^{\pm}$ returns $+1$ if the argument is true and $-1$ 
otherwise\footnote{Theories in \cite{rosset2003margin} use binary labels $y\in\{\pm 1\}$ simplifying 
the signed margin expression in \eqref{eq:d_signed} to $d_X^\wc(y)=\frac{ y \, \wc^\top X}{\|\wc\| }$. 
Instead, our natural labels $y\in\{1,\cdots,K\}$ simplify notation for prediction $\softmax^y$ at each class $y$.}\@. 
As easy to check, the sign of distance $d_X^\wc(y)$ in \eqref{eq:d_signed} is positive for 
correctly classified points $X$, and it is negative otherwise,
see cyan color in Figure~\ref{fig:margins}(a).

Logistic regression is known to satisfy conditions for 
margin maximizing classification \cite{rosset2003margin}.  
\begin{property}[\bf exponential decay for $r$] \label{prop:decay_infty}
Logistic loss $r(x)$ in \eqref{eq:logloss} satisfies the following ``fast-enough'' decay 
condition
\begin{equation} \nonumber
 \lim_{x\rightarrow\infty}\frac{r(x\cdot(1-\epsilon))} {r(x)}=\infty,\;\;\;\;\;\forall\epsilon 
\end{equation}
which is a sufficient condition for margin-maximizing classification. This is case "$T=\infty$"
in Theorem 2.1 from \cite{rosset2003margin}.
\end{property}

For completeness, we also state a special case of Theorem 2.1 in \cite{rosset2003margin} 
for logistic regression. 
We  use the following total loss formulation based on NLL \eqref{eq:binary_Rinfty_sup} 
\begin{equation} \label{eq:log_reg}
R_\infty(\wc|\y) := -\overline{\ln\softmax^{y_X}(\wc^\top X)}
\end{equation}
where the bar indicates averaging over all data points $\{X_i\}$.
\begin{theorem}[\bf  max-margin for logistic regression \cite{rosset2003margin}] \label{th:logreg}
Consider any given linearly separable labeling 
$\y:=\{y_i\}_{i=1}^N$ for dataset $\{X_i\}_{i=1}^N$. 
Assuming linear classifier $\wc(\gamma)$ minimizes
regularized logistic loss over classifiers $\wc\in{\bf V}^\y$ 
consistent with some given (e.g. ground truth) labeling $\y$
$$\wc(\gamma) \;\;:=\;\; \arg\min_{\wc\in {\bf V}^\y}  \gamma\|\wc\|^2 + R_\infty(\wc|\,\y)$$ 
then $$\frac{\wc(\gamma)}{\| \wc(\gamma)\|} \;\;\xrightarrow{\gamma\rightarrow 0} \;\; \uc^\y$$
where  $\uc^\y$ is a unit-norm max-margin linear classifier for $\y$. 
\end{theorem}

{\bf Max-margins for $R_\infty$-decisiveness:}
To establish the margin-maximizing property for clustering, we should drop full supervision where
true labeling $\y$ is given. Instead, we formulate max-margin clustering for a given set $\FL$ of 
allowed or {\em feasible} linearly separable binary labelings $\y$ for dataset $\{X_i\}_{i=1}^N$. 
In this unsupervised context, labelings $\y\in\FL$ represent possible data clusterings. 
For example, $\FL$ could represent ``fair'' clusterings\footnote{The exact definition 
of ``fair'' clustering is irrelevant here.}. 
As a minimum, we normally assume that linearly separable labelings in $\FL$ 
exclude trivial solutions with empty clusters.

Naturally, labelings $\y\in\FL$ divide the data into clusters differently. 
Thus, the corresponding inter-cluster gaps vary. 
\begin{definition} \label{def:labeling_margin}
  Assume $\y=\{y_i\}_{i=1}^N$ is a binary linearly separable labeling of 
  the dataset $\{X_i\}_{i=1}^N$. Then, the {\em gap} or {\em margin size} 
  for labeling (clustering) $\y$ is defined as 
  $$ |\y|\;\;:=\;\;\max_{\uc\in {\bf U}^\y} \min_i \; |\uc^\top X_i | 
  \;\;\equiv\;\; \min_i \; | (\uc^\y)^\top X_i | $$
  where $\uc^y$ is the max-margin unit-norm linear classifier for $\y$.
\end{definition}
This definition associates the term {\em margin size} with labelings. 
Alternatively, {\em margin size} is often associated with classifiers. 
Given a separable labeling $\y$, any consistent linear classifier $\uc\in{\bf U}^\y$ has 
margin size $min_i |\uc^\top X_i|$. 
The {\em margin size} for labeling $\y$ in Definition~\ref{def:labeling_margin} is the 
margin size of the optimal max-margin classifier $\uc^\y$.

Now we define {\em max-margin clustering} in an unsupervised setting restricted to $\FL$.
It extends the standard concept of {\em max-margin classification} restricted to one (true) labeling $\y$. 
Our focus is on optimal labelings rather than classifiers.
\begin{definition} \label{def:max-margin-clustering}
    Consider any set $\FL$ of allowed linearly separable binary labelings $\y$ 
    for dataset $\{X_i\}_{i=1}^N$. Then, {\em max-margin clustering} for $\FL$ 
    is defined as $$\hat{\y} \;\;:=\;\;\arg\max_{\y\in\FL} |\y |  $$
    where $|\y|$ is the gap size for clustering $\y$, as in Definition~\ref{def:labeling_margin}.
\end{definition}
The achieved maximum gap value defining $\hat{\y}$ is finite. Indeed,
gap sizes for  $\y\in\FL$ are uniformly bounded by the diameter of the dataset $\{X_i\}_{i=1}^N$
$$|\y|\;\;\leq\;\; \max_{1\leq i<j\leq N} \|X_i-X_j\| \quad\quad\forall\y\in\FL$$ 
assuming that $\FL$ excludes trivial clusterings.

Our next theorem extends Theorem~\ref{th:logreg} to clustering based on $R_\infty$-decisiveness.
We define the total decisiveness as
\begin{equation} \label{eq:R_inf_decisiveness}
 R_\infty(\wc) \;:=\;   \overline{R_\infty (\softmax(\wc^\top X))}  \;\equiv\; -\overline{\ln \softmax^{y^\wc_X}(\wc^\top X)}   
\end{equation}
where the bar indicates averaging over all data points $\{X_i\}$.
The proof uses the equivalent expression on the right, which is a self-labeling version of NLL \eqref{eq:log_reg} 
based on \eqref{eq:self_labeling} and \eqref{eq:binary_Rinfty_sup}.

\begin{theorem}[\bf max-margin clustering for $R_\infty$] \label{th:Rinfty}
Consider any set $\FL$ of allowed or feasible linearly separable binary labelings $\y$ for dataset $\{X_i\}_{i=1}^N$, 
e.g. restricted to ``fair'' clusterings of the data. Assuming linear classifier $\wc(\gamma)$ minimizes
regularized decisiveness over classifiers $\wc\in{\bf V}^\FL$ consistent with $\FL$
$$\wc(\gamma) \;\;:=\;\; \arg\min_{\wc\in {\bf V}^\FL}  \gamma\|\wc\|^2 + R_\infty(\wc) $$ 
then $$\frac{\wc(\gamma)}{\| \wc(\gamma)\|} \;\;\xrightarrow{\gamma\rightarrow 0} \;\; \uc^{\hat{\y}}$$ 
for the maximum margin clustering $\hat{\y}$ in $\FL$, as in Definition~\ref{def:max-margin-clustering}.
\end{theorem}
\begin{proof}
The proof requires analysis of the loss $R_\infty(\softmax(\wc^\top X))$, but due to identity \eqref{eq:self_labeling},
it can follow the same steps as the analysis of the supervised loss \eqref{eq:margins_signed} from
Theorem 2.1 in \cite{rosset2003margin} (case ``a''). Indeed, they consider supervised classification with 
a given true linearly separable labeling $\y$, but it appears in their proof only in the margin expression $d^\wc_X(y)$ 
setting its right sign, see \eqref{eq:d_signed}.
Their analysis is restricted to classifiers consistent with $\y$, but we observe that
their supervised loss \eqref{eq:margins_signed} has an equivalent {\em constrained} formulation
\begin{equation} \label{eq:logloss_NLL}
R_\infty(\softmax(\wc^\top X)\,|\,y)\;\;\equiv\;\;r(d_X^\wc\|\wc\|) \quad\quad\quad \text{for}\;\wc\in{\bf V}^\y
\end{equation}
using a simpler unsigned expression for the margins 
\begin{equation} \label{eq:d_unsigned}
d_X^\wc\;:=\;\frac{ |\wc^\top X|}{\|\wc\|} 
\end{equation}
as \eqref{eq:d_signed} is guaranteed to be non-negative for all consistent classifiers, 
see Figure~\ref{fig:margins}(a). Our formulation of $R_\infty(\cdot\,|\,y)$ in \eqref{eq:logloss_NLL} 
does not depend on labeling $\y$, as as long as $\wc\in{\bf V}^\y$.

In the context of clustering, i.e. unsupervised classification without given (true) labeling $\y$,
we show that entropy values $R_\infty(\softmax(\wc^\top X))$ relate to 
logistic loss and distances to the clustering boundary, a.k.a. {\em margins}. 
Indeed, as implied by self-labeling identity \eqref{eq:self_labeling} and \eqref{eq:logloss_NLL}, we  get
\begin{equation} \label{eq:logloss_entropy}
R_\infty(\softmax(\wc^\top X))\;\;\equiv\;\;r(d_X^\wc\|\wc\|) \quad\quad\text{for any}\; \wc
\end{equation}
for any linear classifier $\wc$, which is always consistent with its ``hard-max'' labeling $\y^\wc$, by definition. 
Our equation \eqref{eq:logloss_entropy} uses the unsigned expression for margins \eqref{eq:d_unsigned}, 
see Figure~\ref{fig:margins}(b).

The formal max-margin analysis for classifiers $\wc\in{\bf V}^\y$ in \cite{rosset2003margin} uses only logistic loss values, 
which we expressed as \eqref{eq:logloss_NLL}. The same analysis directly applies to the identical decisiveness values 
\eqref{eq:logloss_entropy} for classifiers $\wc\in{\bf V}^\FL$, 
even when such classifiers may have different labelings $\y^\wc\in\FL$.
\end{proof}

\subsubsection{Generalization for $a>0$ } \label{sec:Ralpha}

Binary Renyi entropy $R_\alpha(\softmax) \equiv R_\alpha(\sigmoid)$, defined in (\ref{eq:binary_Ra_def}-\ref{eq:binary_H_def}) 
for distribution $\softmax = (\sigmoid,1-\sigmoid)$,
is symmetric $R_\alpha(\sigmoid)=R_\alpha(1-\sigmoid)$ around $\sigmoid=0.5$ for any order $\alpha\geq 0$, 
see Figure \ref{fig:renyi_a}.
We define a monotone non-increasing function, see Figure~\ref{fig:renyi_supervised}(b)
\begin{equation} \label{eq:f_alpha}
f_\alpha(\sigmoid)\;:=\begin{cases} R_\alpha(\sigmoid) & \text{if}\; \sigmoid\geq 0.5 \\ 
                          \ln 2 & \text{if} \; \sigmoid\leq 0.5
                \end{cases}
\end{equation}
that can be used for training softmax classifier $\softmax(\wc^\top X)$ 
based on the following supervision loss
\begin{equation} \label{eq:R_a_supervised}
R_\alpha(\softmax\,|\,y) \;\;:=\;\;  f_\alpha (\softmax^y).
\end{equation}
Function $f$ satisfies the following identities, see Figure~\ref{fig:renyi_supervised}(b),
$$R_\alpha(\softmax) \;\equiv\; \min\{f_\alpha(\sigmoid),f_\alpha(1-\sigmoid)\} 
\;\equiv\; f_\alpha(\max\{\sigmoid,1-\sigmoid\}) $$
analogous to the relation \eqref{eq:Rinfty_and_NLL} between $R_\infty(\softmax)$ and logistic regression,
see Figure~\ref{fig:renyi_supervised}(a). As in \eqref{eq:self_labeling},
the identities above imply the ``self-labeling'' identity
\begin{equation} \label{eq:self_labeling_alpha}
R_\alpha(\softmax)  \;=\;  R_\alpha(\softmax\,|\,y^\softmax)    
\end{equation}
between decisiveness $R_\alpha(\softmax)$ and the supervised loss \eqref{eq:R_a_supervised}.

Similarly to Section \ref{sec:Rinfty}, we will prove the margin maximization property for decisiveness 
$R_\alpha(\softmax)$ using the self-labeling relation \eqref{eq:self_labeling_alpha} and the
standard theories for supervised losses \cite{rosset2003margin}. 
Instead of logistic loss $r(x)$ in \eqref{eq:logloss}, we use
\begin{eqnarray} 
g_\alpha(x)  & := & f_\alpha(\sigmoid(x)) \nonumber \\[1ex] & \stackrel{\text{(\ref{eq:f_alpha},\ref{eq:binary_Ra_def})}}{=}  & 
\label{eq:g_alpha} 
\begin{cases} \frac{\ln\left( (1+e^{-x} )^{-\alpha} + (1+e^{x} )^{-\alpha}  \right)}{1-\alpha}
& \mbox{if $x\geq 0$} \\ \ln 2 & \mbox{if $x\leq 0$} \end{cases}  
\end{eqnarray}
which is a monotone non-increasing supervision loss enabling an expression for $R_\alpha(\softmax|y)$ 
\eqref{eq:R_a_supervised} analogous to \eqref{eq:margins_signed}
\begin{equation} \label{eq:g_margins_signed}
R_\alpha(\softmax(\wc^\top X)\,|\,y)\;\;\equiv\;\;g_\alpha(d_X^\wc(\y)\|\wc\|) 
\end{equation}
where $d_X^\wc(\y)$ are (signed) margins \eqref{eq:d_signed}.
Loss $g_\alpha(x)$ satisfies the ``fast-enough decay'' condition from Theorem 2.1 in \cite{rosset2003margin}.
\begin{property}[\bf exponential decay for $g_\alpha$]  \label{prop:decay_a}
For any positive value of parameter $\alpha>0$, loss function $g_\alpha $ in \eqref{eq:g_alpha} satisfies
\begin{equation} \nonumber
 \lim_{x\rightarrow\infty}\frac{g_\alpha(x\cdot(1-\epsilon))} {g_\alpha(x)}=\infty,\;\;\;\;\;\forall\epsilon 
\end{equation}
which is the sufficient condition for margin-maximizing classification according to  Theorem 2.1 from \cite{rosset2003margin}.
\end{property}
\begin{proof}
The technical condition above is easy to check. We leave this as an exercise to the reader.
Note that two special cases $\alpha\in\{1,\infty\}$ require alternative (asymptotically consistent)
expressions for Renyi entropy, e.g. $R_1$ can use the formulation in \eqref{eq:binary_H_def} 
instead of \eqref{eq:binary_Ra_def}. Then, the expressions for $g_1$ and $g_\infty$ are different from \eqref{eq:g_alpha}, 
but they also have the exponential decay property, as easy to check.
\end{proof}
Property \ref{prop:decay_a} allows us to generalize our max-margin clustering Theorem~\ref{th:Rinfty} 
to the total $R_\alpha$-decisiveness loss
\begin{eqnarray}  
R_\alpha(\wc)  & := &  \overline{R_\alpha(\softmax(\wc^\top X))}  \;\;\equiv\;\; 
                        \overline{R_\alpha(\softmax(\wc^\top X) \,|\, y^\wc_X )}   \nonumber  \\
               & \equiv &    \overline{f_\alpha(\softmax^{y^\wc_X}(\wc^\top X))} \label{eq:R_alpha_decisiveness}
\end{eqnarray}
where $f_\alpha$ replaces "$-\ln$" in the {\em self-labeling NLL} loss \eqref{eq:R_inf_decisiveness}.
\begin{theorem}[\bf  max-margin clustering for $R_\alpha$] \label{th:Ralpha}
Consider any set $\FL$ of allowed linearly separable binary clusterings/labelings $\y$ for dataset $\{X_i\}_{i=1}^N$, 
e.g. restricted to ``fair'' clusterings. Given any $\alpha>0$, assume that $\wc(\gamma)$ minimizes
regularized decisiveness over linear classifiers $\wc\in{\bf V}^\FL$ consistent with $\FL$
$$\wc(\gamma) \;\;:=\;\; \arg\min_{\wc\in {\bf V}^\FL}  \gamma\|\wc\|^2 + R_\alpha(\wc).$$ 
Then $$\frac{\wc(\gamma)}{\| \wc(\gamma)\|} \;\;\xrightarrow{\gamma\rightarrow 0} \;\; \uc^{\hat{\y}}$$ 
for the maximum margin clustering $\hat{\y}$ in $\FL$, as in Definition~\ref{def:max-margin-clustering}.
\end{theorem}
\begin{proof} 
We follow the same arguments as in the proof of Theorem~\ref{th:Rinfty}. 
Equation \eqref{eq:g_margins_signed} and self-labeling relation \eqref{eq:self_labeling_alpha} 
imply identical expressions for $R_\alpha$-values 
w.r.t. unsigned margins $d_X^\wc=\frac{|\wc^\top X|}{\|\wc\|}$ in two cases: 
(supervision) for classifiers $\wc\in{\bf V}^\y$ consistent with some given (true) labeling $\y$
\begin{equation} \nonumber 
R_\alpha(\softmax(\wc^\top X)\,|\,y)\;\;\equiv\;\;g_\alpha(d_X^\wc \|\wc\|)\quad\quad\quad \text{for}\;\wc\in{\bf V}^\y
\end{equation}
and (decisiveness) for any classifier 
\begin{equation} \nonumber 
R_\alpha(\softmax(\wc^\top X))\;\;\equiv\;\;g_\alpha(d_X^\wc \|\wc\|)   \quad\quad\quad\text{for any}\; \wc
\end{equation}
which are analogous to logistic loss \eqref{eq:logloss_NLL} and decisiveness \eqref{eq:logloss_entropy}.
The only difference between these supervised and unsupervised cases is the set of allowed
classifiers. Since classification loss $R_\alpha(\wc\,|\,\y)$ and decisiveness $R_\alpha(\wc)$ 
have identical dependence on unsigned margins $d_i^\wc=\frac{|\wc^\top X_i|}{\|\wc\|}$,
all arguments from Theorem 2.1 \cite{rosset2003margin} proving 
the margin maximizing property for the (exponential-decay) classification loss $R_\alpha(\wc\,|\,\y)$ 
directly apply to the clustering loss $R_\alpha(\wc)$.
\end{proof}

Note that there is no margin maximization for $\alpha=0$ since
$R_0(\softmax(\wc^\top X))= const$ for any finite data point $X$.

\subsubsection{Generalization for $K>2$} \label{sec:Ralpha_K}

We use the max-margin property for multi-label 
classification in Theorem 4.1 \cite{rosset2003margin} to prove the max-margin
property for clustering  with regularized $R_\alpha$-decisiveness when $K>2$. 
We need to introduce some additional terminology.

First, the regularization for max-margin $K$-clustering is based on the squared Frobenius 
norm of classifier matrix $\wc$ 
$$\|\wc\|^2_{\cal F}:=\sum_{k=1}^K \|\wc_k\|^2$$ 
where $\|\wc_k\|$ are $L_2$ norms of linear discriminators $\wc_k$ for each class $k$, which are the columns of matrix $\wc$.
As before, we exclude the bias parameters in the norms of $\wc_k$.

We use the unit-norm multi-class classifiers consistent with any given labeling $\y$
$${\bf U}^\y:=\{\wc\,|\,\y^\wc = \y,\;\|\wc\|_{\cal F}=1\}$$
as in the binary case in Section \ref{sec:Ralpha}. The next definition of margin size for multi-class labelings
is consistent with \cite{rosset2003margin}.
\begin{definition} \label{def:labeling_margin_K}
  Assume $\y=\{y_i\}_{i=1}^N$ is a linearly separable multi-class labeling for $\{X_i\}_{i=1}^N$ 
  such that $y_i\in\{1,\cdots,K\}$. 
  Then, the {\em gap} or {\em margin size} for labeling/clustering $\y$ is 
  \begin{eqnarray} \nonumber
  |\y| & := &  \max_{\uc\in {\bf U}^\y} \min_i \min_{k\neq y_i}\; |(\uc_{y_i}-\uc_k)^\top X_i |   \\
  & \equiv & \quad\quad\; \min_i \min_{k\neq y_i} \; | (\uc^\y_{y_i} - \uc^\y_k )^\top X_i |     \nonumber
  \end{eqnarray}
  where $\uc^\y$ is the max-margin unit-norm linear classifier for $\y$.
\end{definition}
The multi-class margin in Definition \ref{def:labeling_margin_K} is determined by the two closest classes/clusters.
The gaps between the other pairs of clusters are ignored as long as they are larger.
\begin{figure}[t]
\centering
\begin{tabular}{c @{\extracolsep{1ex}} c}
     \includegraphics[width=0.4\linewidth]{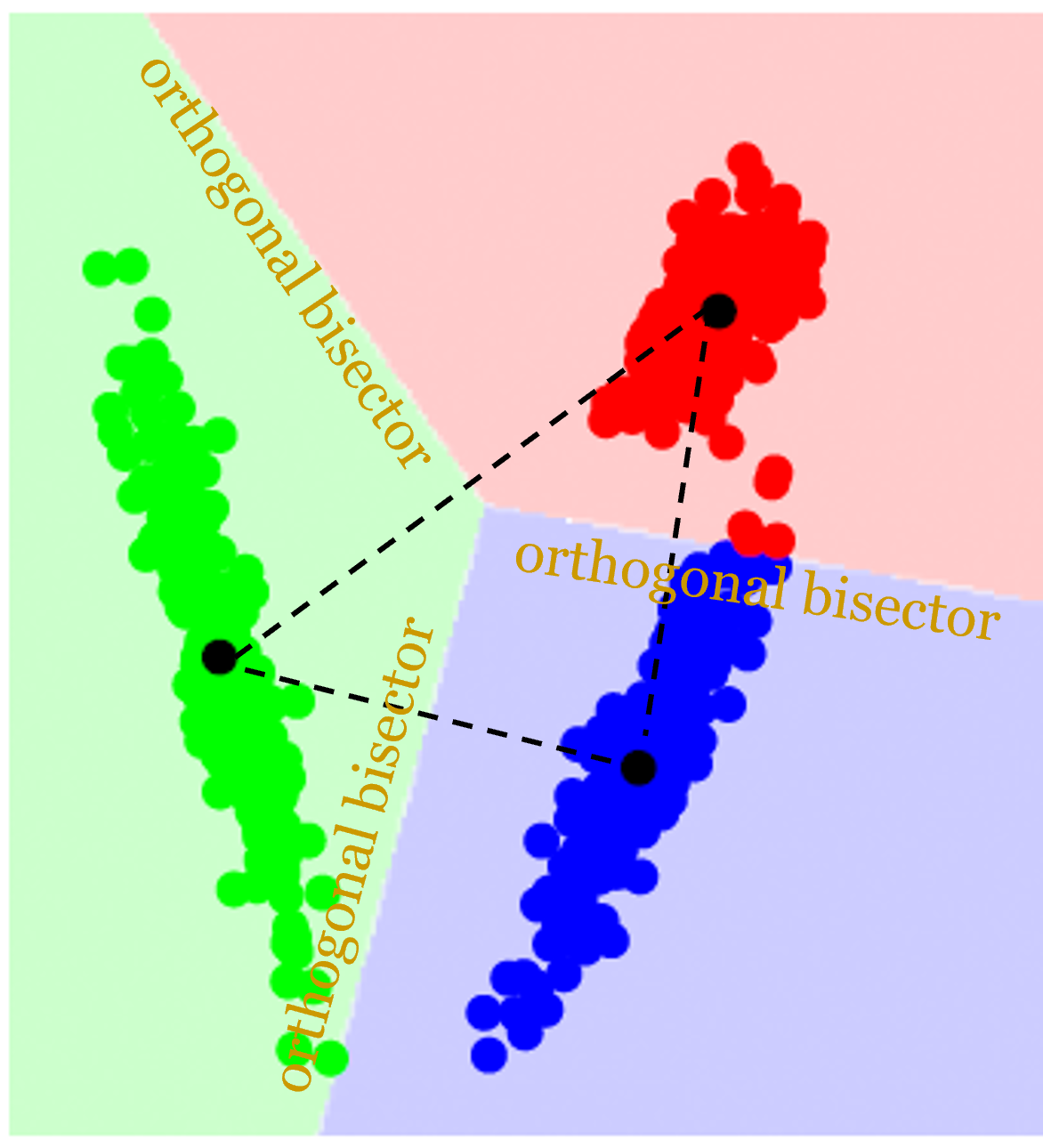} 
     & \includegraphics[width=0.4\linewidth]{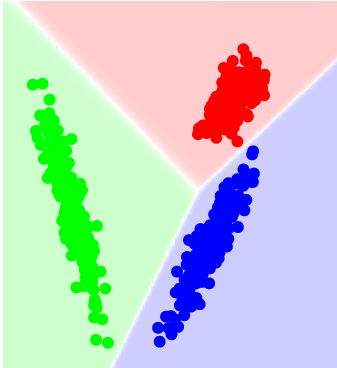}  \\
    (a) soft K-means \eqref{eq:sKM}  & (b) entropy clustering \eqref{eq:rec}
\end{tabular}
\caption{Multi-label clustering: K-means vs. entropy clustering. 
Consistently with Theorem \ref{th:Ralpha_K}, loss \eqref{eq:rec} produces max-margin clustering (b), as in Definitions 
\ref{def:labeling_margin_K}-\ref{def:max-margin-clustering_K}, satisfying the fairness. The margin is tight only between the red and blue clusters.}
\label{fig:multi-label}
\end{figure}

\begin{definition} \label{def:max-margin-clustering_K}
    Consider any set $\FL$ of allowed linearly separable multi-class labelings $\y$ 
    for dataset $\{X_i\}_{i=1}^N$ where $y_i\in\{1,\cdots,K\}$. Then, {\em max-margin clustering} for $\FL$ is 
    $$\hat{\y} \;\;:=\;\;\arg\max_{\y\in\FL} |\y |  $$
    where $|\y|$ is the gap size for clustering $\y$, as in Definition~\ref{def:labeling_margin_K}.
\end{definition}
Figure \ref{fig:multi-label}(b) is an example of multi-class clustering achieving the max-margin 
(Def. \ref{def:labeling_margin_K}) among fair solutions. 
The tightest spot between the red and blue clusters determines $|\hat{\y}|$.

\begin{figure}
\begin{tabular}{c@{\extracolsep{1ex}}c}
    \centering
    \includegraphics[width=4.0cm]{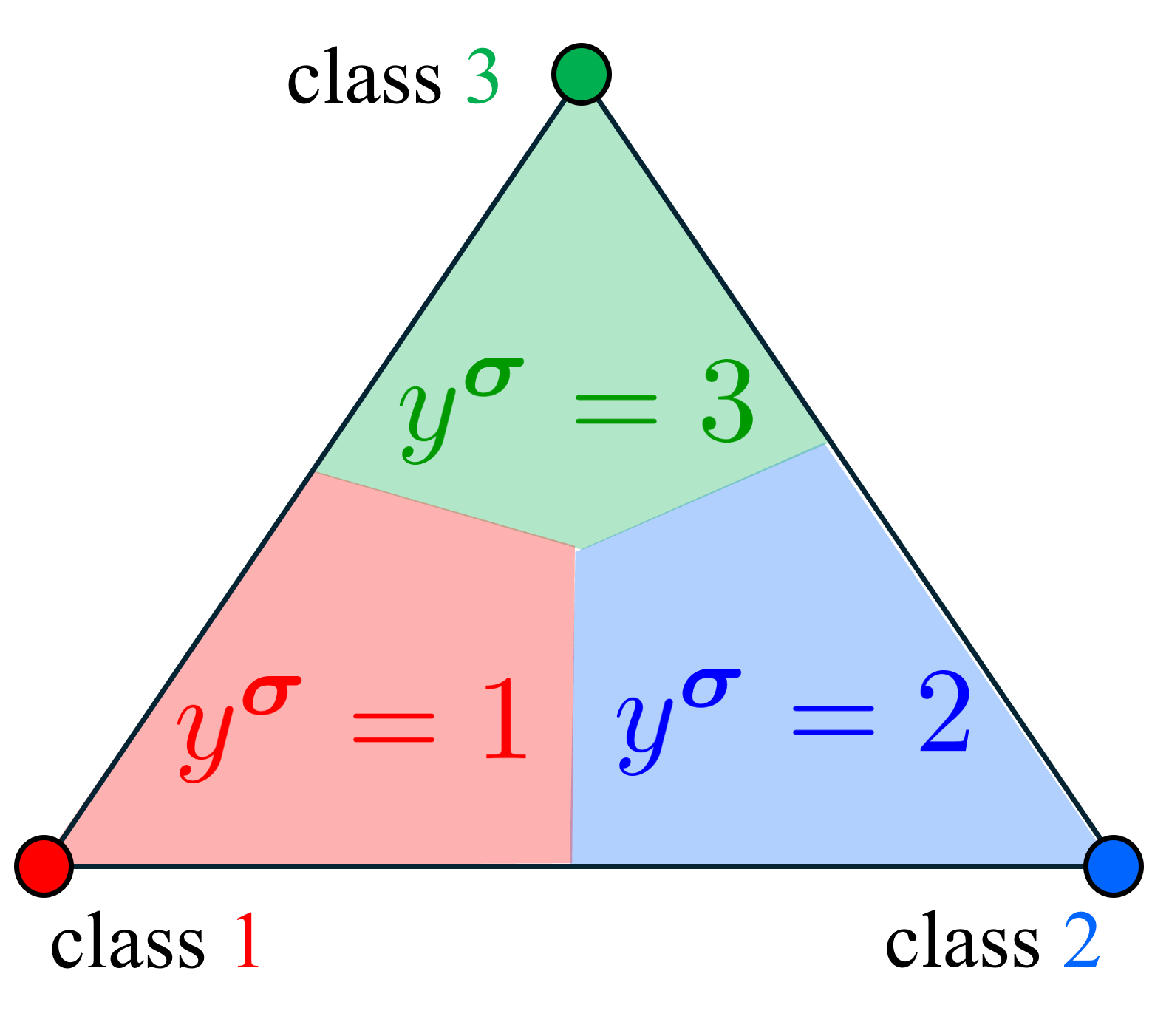}  &
     \includegraphics[width=4.0cm]{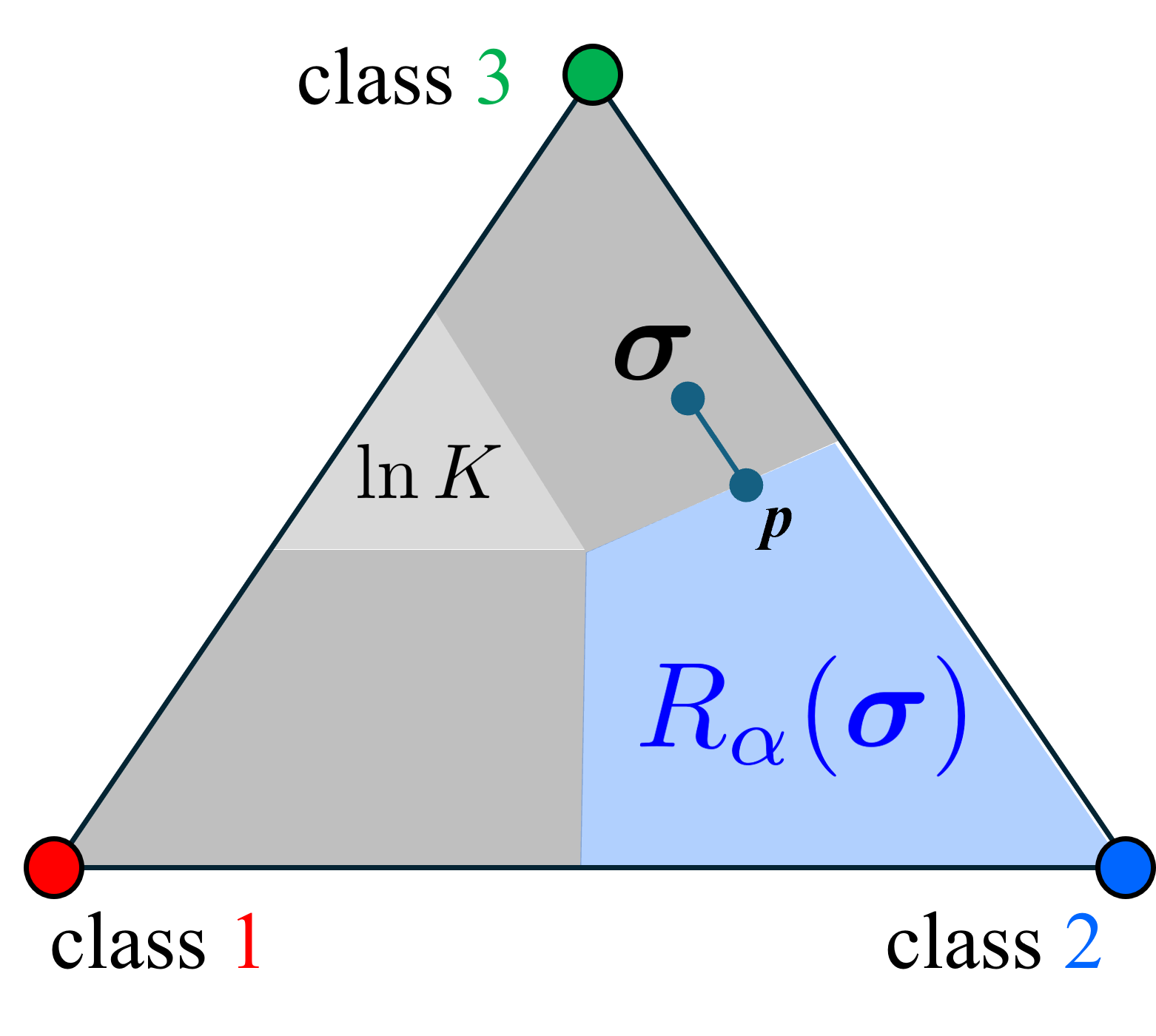}      \\   
    (a) Predictions $\softmax\in\Delta^3$  &
    (b) Supervised loss $R_\alpha(\softmax|y)$ \\ 
    where $y^\softmax = 1,2,3$ & for ground truth label $y=2$
\end{tabular}
\caption{Supervised loss $R_\alpha(\softmax\,|\,y)$: extending Figure~\ref{fig:renyi_supervised}(b) to $K=3$. 
Three colored areas of probability simplex $\Delta^3$ in (a) mark predictions $\softmax$
with different hard-max labels $\y^\softmax$. Assuming $y=2$, the blue region in (b) shows where supervised losses $R_\alpha(\softmax\,|\,y)$ in \eqref{eq:supervised_K_A} and \eqref{eq:supervised_K_B} 
are consistent with entropy $R_\alpha(\softmax)$. Loss \eqref{eq:supervised_K_A} is constant $\ln K$ 
in all gray regions. It is discontinuous at the blue/gray boundary, except in the center. 
The continuous variant \eqref{eq:supervised_K_B} is constant $\ln K$ only in the light gray region.
For $\softmax$ in the dark gray area, it is defined by the entropy $R_\alpha({\bf p})$ at the closest point blue $\bf p$.  
\label{fig:simplex_parts}}
\end{figure}

We state max-margin clustering theorem for multi-class decisiveness of order $\alpha$
consistent with the binary case \eqref{eq:R_alpha_decisiveness}
\begin{equation} \nonumber
R_\alpha(\wc)  \;\;:=\;\;  \overline{R_\alpha(\softmax(\wc^\top X))} 
\end{equation}
using the general definition of Renyi entropy \cite{Renyi1961} for $K\geq 2$
\begin{equation} \label{eq:multi_Ra_def}
R_\alpha (\softmax) \;:=\; \frac{\ln \sum_{k=1}^K (\softmax^k)^\alpha}{1-\alpha}
\end{equation}
extending \eqref{eq:binary_Ra_def} to multi-class predictions $\softmax=(\softmax^1,\cdots,\softmax^K)$.
\begin{theorem}[\bf  max-margin multi-class clustering for $R_\alpha$] \label{th:Ralpha_K}
Consider any set $\FL$ of allowed or feasible linearly separable multi-class labelings $\y$ for $\{X_i\}_{i=1}^N$, 
e.g. restricted to ``fair'' clusterings. Given any order $\alpha>0$, assume that $\wc(\gamma)$ minimizes
regularized decisiveness over linear classifiers $\wc\in{\bf V}^\FL$ consistent with $\FL$
$$\wc(\gamma) \;\;:=\;\; \arg\min_{\wc\in {\bf V}^\FL}  \gamma\|\wc\|^2_{\cal F} + R_\alpha(\wc).$$ 
Then $$\frac{\wc(\gamma)}{\| \wc(\gamma)\|} \;\;\xrightarrow{\gamma\rightarrow 0} \;\; \uc^{\hat{\y}}$$ 
for the maximum margin clustering $\hat{\y}$ in $\FL$, as in Definition~\ref{def:max-margin-clustering_K}.
\end{theorem}
\noindent {\bf Idea of proof:} Definition \ref{def:labeling_margin_K} above is consistent with
multi-class $L_2$-margin (10) in \cite{rosset2003margin}. Their multi-class max-margin
classification Theorem 4.1 extends to clustering with regularized
decisiveness $R_\alpha(\wc)$ due to self-labeling relation \eqref{eq:self_labeling_alpha}, 
which also works for $K>2$ with a properly constructed multi-class supervised loss $R_\alpha(\softmax\,|\,y)$. For example, 
generalizing (\ref{eq:f_alpha},\ref{eq:R_a_supervised}) to the case $K>2$, one can use supervised loss
\begin{equation} \label{eq:supervised_K_A}
R_\alpha(\softmax\,|\,y)\;:= \begin{cases} R_\alpha(\softmax) & \text{if}\; y^\softmax = y \\ 
                          \ln K & \text{o.w.}
                \end{cases}
\end{equation}
or its continuous variant, see Figure \ref{fig:simplex_parts},
\begin{equation} \label{eq:supervised_K_B}
R_\alpha(\softmax\,|\,y)\;:= R_\alpha\left(\arg\min_{{\bf p}\in\Delta^K:\,y^{\bf p}=y} \|{\bf p}-\softmax\|  \right)
\end{equation}
where the argument of entropy $R_\alpha(\bf p)$ is distribution ${\bf p}\in\Delta^K$ 
consistent with label $y$ that is the closest to prediction $\softmax$. 
It is necessary to check that, as a function of $(K-1)$-dimensional {\em margins}, 
supervised loss $R_\alpha(\cdot\,|\,y)$ satisfies 
the exponential decay condition (11) in Theorem 4.1 \cite{rosset2003margin}.
The remaining proof of Theorem \ref{th:Ralpha_K} follows the same arguments as in Theorem~\ref{th:Ralpha}.

\subsubsection{Discussion: entropy clustering vs SVM clustering} \label{sec:EC_vs_SVM}

After establishing the max-margin property for regularized decisiveness \eqref{eq:reg_decisiveness}
or \eqref{eq:reg_decisiveness_R}, it is interesting to review the implications for entropy clustering 
and juxtapose it with SVM clustering. In particular, the formulation in \cite{Schuurmans2004} is highly relevant.
They use the the regularized {\em hinge loss}, as in {\em soft SVM}, combined with a hard {\em fairness} constraint
\begin{eqnarray} \label{eq:mm-hinge}
L_{mm} & = & \gamma\|\wc\|^2 \;+\; \overline{\max\{0,1-t\,\wc^\top X\}} \\  \nonumber
    & & \text{subject to}\quad  -\epsilon\leq\bar{t}\leq \epsilon
\end{eqnarray}
where, unlike supervised SVM \cite{vapnik:1995}, binary targets $t\in\{\pm 1\}$ are optimization variables jointly estimated with the linear model parameters $\wc$. Following our terminology from Section \ref{sec:related_work}, objective \eqref{eq:mm-hinge} is a self-labeling loss with binary pseudo-labels $t$. 
It resembles \eqref{eq:vidaldi} where hard constraints represent fairness and estimated pseudo-labels 
are discrete/hard. 
One can also relate the hinge loss in \eqref{eq:mm-hinge} to a truncated variant of 
decisiveness\footnote{Minimizing out $t$ in the hinge loss \eqref{eq:mm-hinge} 
gives $\max\{0,1-|\wc^\top X|\}$.} encouraging linear discriminant $\wc$ 
such that $|\wc^\top X|\geq 1$ for all data points $X$.


However, one significant difference exists between \eqref{eq:mm-hinge} and \eqref{eq:vidaldi}. Minimization of the norm $\|\wc\|^2$ in supervised or unsupervised SVM, as in \eqref{eq:mm-hinge}, 
is known to be central to the problem. It is well understood that it directly corresponds to margin maximization. 
In contrast, entropy clustering methods \cite{MacKay1991,hu2017learning,ghasedi2017deep,ji2019invariant,YM.2020Self-labelling} typically skip $\|\wc\|^2$ in their losses, e.g. \eqref{eq:vidaldi}, though it may appear implicitly due to the omnipresence of the {\em weight decay}.
It is included in \eqref{eq:rec_ismail} \cite{ismail2021}, but for the wrong reason discussed 
in Section \ref{sec:Kmeans}.
It is also a part of \eqref{eq:mi+decay}, but it is
motivated only by a generic model simplicity argument in the context of complex models \cite{Perona2010}. It is also argued in \cite{bengio2004semi} that 
decisiveness $\overline{H(\softmax)}$ encourages 
``decision boundary away from the data points'', which may informally suggest margin maximization, 
but norm $\|\wc\|^2$ is not present in their argument. 
In fact, margin maximization is not guaranteed without norm regularization.

Similarly to SVM, our max-margin theories for regularized decisiveness are focused on a linear binary case 
where guarantees are the strongest, but the multi-label case is also discussed.
Non-linear extensions are possible due to high-dimensional embeddings $f_\wf(X)$ in \eqref{eq:postmodel_deep}, 
but the quality guarantees are weak, as in kernel SVM. One notable difference is that non-linear SVMs 
typically use {\em fixed} data kernels, i.e. implicit high-dimensional embeddings. 
In contrast, discriminative entropy clustering can learn $f_\wf(X)$.

Also, note that our max-margin theories apply to entropy clustering losses combining regularized decisiveness with 
fairness, e.g. \eqref{eq:rec}, only if fairness is not compromised in the solution. 
If decisiveness and fairness require some trade-off, the max-margin guarantee becomes less clear.
This is similar to soft SVM when the margin size is traded off with margin violations. 
Empirically, on balanced toy examples, the max-margin property is evident for \eqref{eq:rec} and 
its self-labeling surrogates, see Sec.\ref{sec:related_work} and Sec.\ref{sec:our_approach}. 
We conjecture that margin maximization could be formally extended to 
many self-labeling formulations of entropy clustering loss including the norm $\|\wc\|^2$, 
but leave the proof to the reader.


\section{Our self-labeling algorithm} \label{sec:our_approach}

The theories above improve the general understanding of discriminative entropy clustering and show certain conditions when
the linear classifier norm regularization leads to margin maximization. Note that the generic {\em weight decay} implicitly represents the norm regularization $\gamma\|\wc\|^2$, but it may not satisfy these conditions. For example, it often includes the bias parameter, and its strength $\gamma$ may not be set correctly.

This section addresses other limitations of prior entropy clustering algorithms in the context of Shannon's entropy $H=R_1$ 
as in \eqref{eq:rec}. 
We focus on self-labeling (Sec.\ref{sec:related_work}) and show that the standard cross-entropy formulation of decisiveness is sensitive to pseudo-label errors. Section~\ref{sec:our_loss} argues that the {\em reverse cross-entropy} is more robust to label noise. 
We also propose {\em strong fairness}. Section \ref{sec:EM} derives an efficient EM algorithm for minimizing our loss w.r.t. pseudo-labels, which is a critical step of our self-labeling algorithm.

\begin{figure*}
\centering
\begin{tabular}{cc}
    \includegraphics[height=5cm]{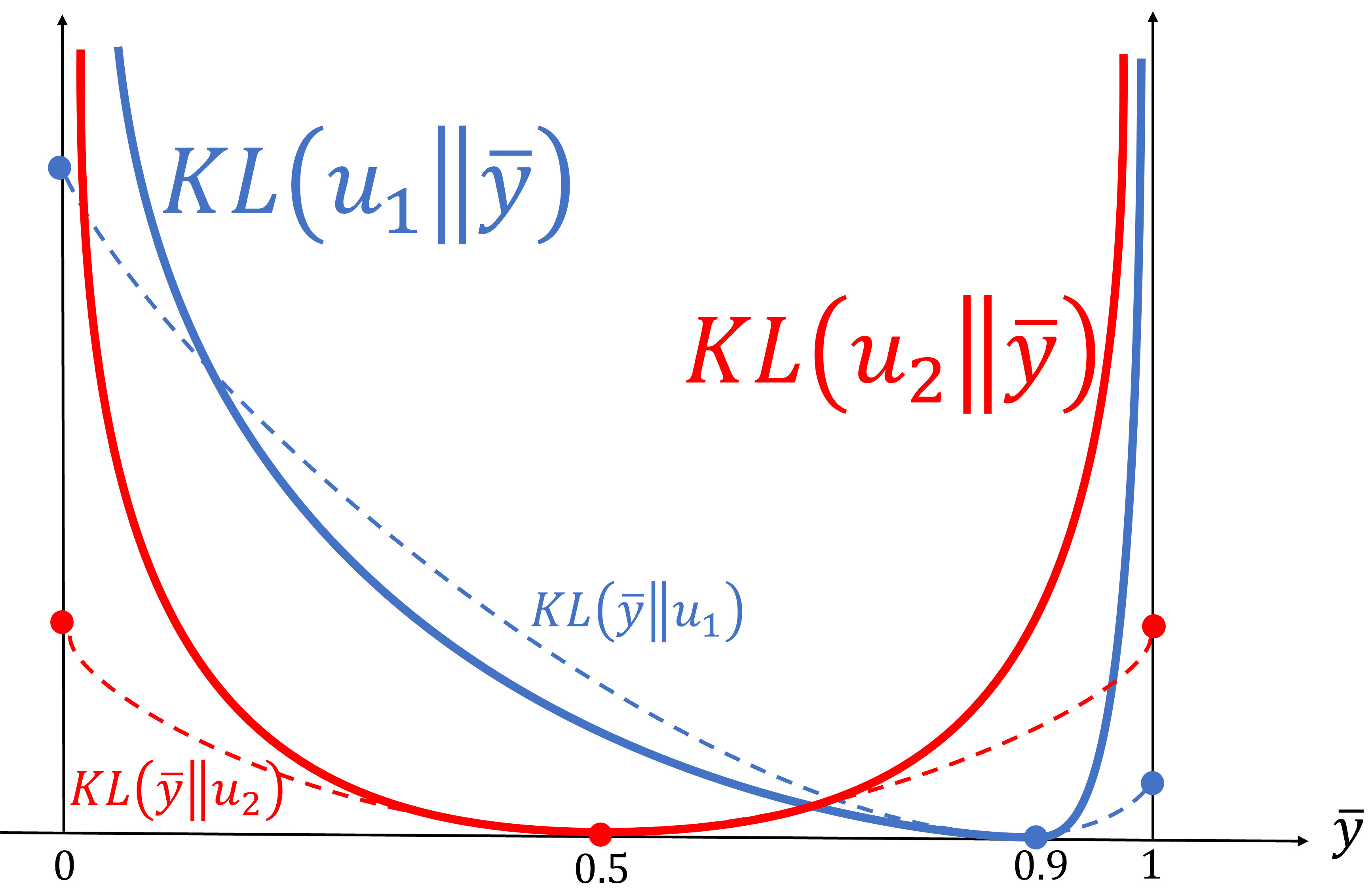} &
    \includegraphics[height=5cm]{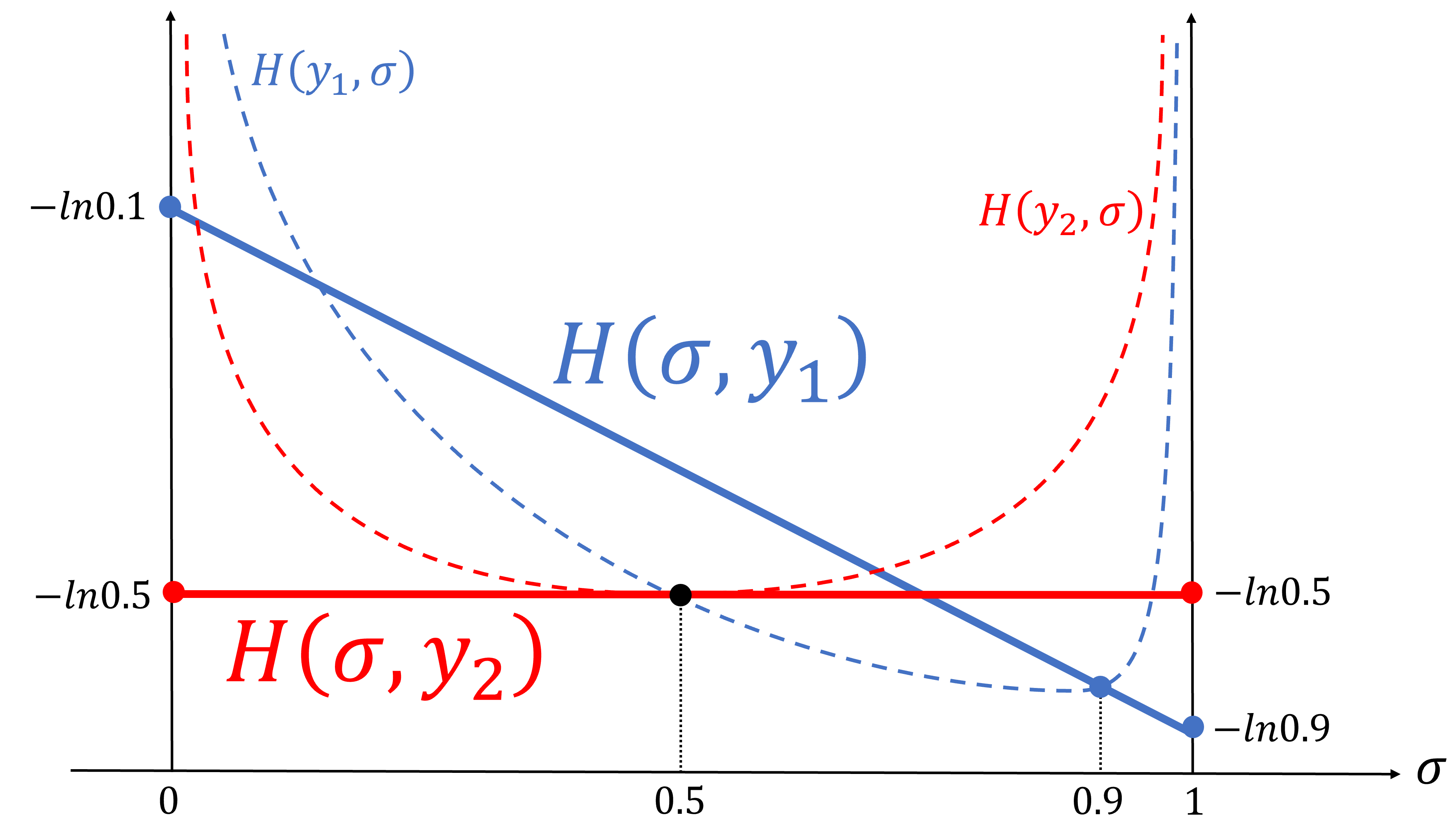} \\
    (a) {\em strong} fairness $KL(u\|\bar{y})$ & (b) {\em reverse} cross-entropy $H(\sigma,y)$
\end{tabular}
\caption{``Forward'' vs ``reverse'': (a) KL-divergence and (b) cross-entropy. Assuming binary classification $K=2$, probability distributions $\sigma$ or $\bar{y}$ are represented as points on [0,1]. The solid curves in (a) illustrate 
the {\em forward} KL-divergence $KL(u\|\bar{y})$ for average pseudo-labels $\bar{y}$ in \eqref{eq:semm_our}, though the same points could be made with average predictions $\bar{\sigma}$ instead of $\bar{y}$.
We show two examples of volumetric prior $u_1=(0.9,0.1)$ (blue) and $u_2=(0.5,0.5)$ (red). The 
reverse KL-divergence $KL(\bar{y}\|u)$ 
(dashed curves) represents the fairness in most prior work on entropy clustering, e.g. see \eqref{eq:mi+decay}, \eqref{eq:ismail}, and \eqref{eq:mi}. 
Unlike $KL(u\|\bar{y})$, 
it tolerates extremely unbalanced clusters, i.e. the trivial solutions at the endpoints of the interval [0,1].
The solid curves in (b) are the {\em reverse} cross-entropy $H(\sigma,y)$ in \eqref{eq:semm_our} as a function 
of predictions $\sigma$. The dashed curves 
are the standard forward cross-entropy $H(y,\sigma)$, as in \eqref{eq:vidaldi} or \eqref{eq:ismail}.
The plots in (b) show examples of two fixed pseudo-labels 
$y_1=(0.9,0.1)$ (blue) and $y_2=(0.5,0.5)$ (red). Our loss $H(\sigma,y)$ weakens 
the training (reduces gradients) on data points with higher label uncertainty (red vs blue lines). 
In contrast, the standard loss $H(y,\sigma)$ treats $y$ as targets and trains the network to mimic this uncertainty, see the optimum on the dashed curves. The training experiments 
in Figure \ref{fig:robustness_to_uncertainty} show that $H(\sigma,y)$ significantly improves robustness of the network to errors in labels $y$ compared to standard cross-entropy $H(y,\sigma)$.}
\label{fig:ce-forward-reverse}
\end{figure*}


\subsection{Self-labeling surrogate loss formulation} \label{sec:our_loss}

\begin{figure}[b!]
    \centering
    \includegraphics[height=4cm]{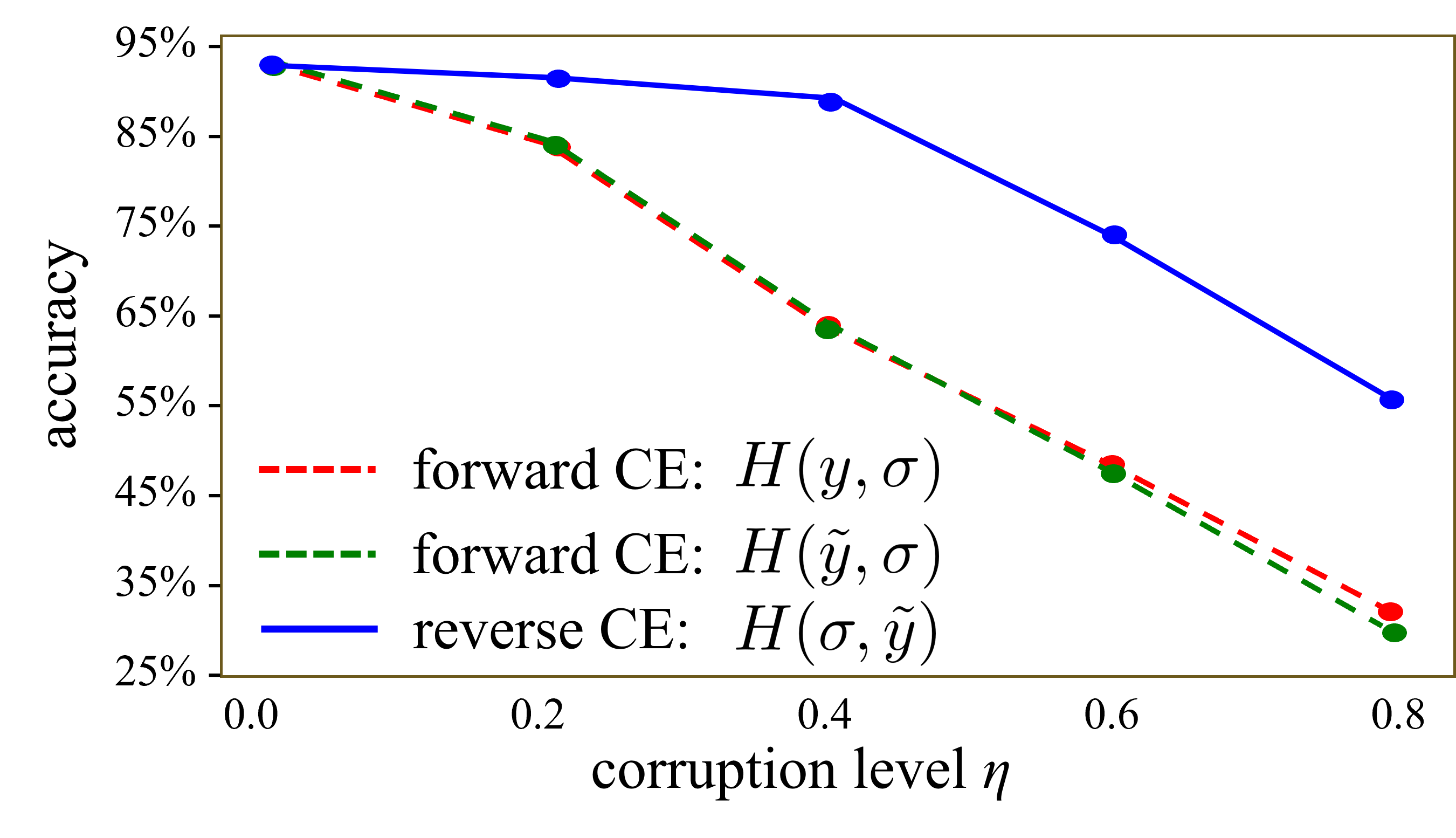}
    \caption{Robustness to noisy labels: reverse $H(\sigma, y)$ vs 
    standard cross-entropy $H(y, \sigma)$. We train ResNet-18 on fully-supervised {\em Natural Scene} dataset \cite{NSD} 
    where we corrupted some labels. The horizontal axis shows the corruption level, i.e. percentage $\eta$ of training images where 
    ground truth labels were randomly replaced. We use true posteriors as soft target distributions 
    $\tilde{y}=\eta*u+(1-\eta)*y$ that is a mixture of one-hot distribution $y$ for the observed corrupt label 
    and the uniform distribution $u$, as in \cite{muller2019does}.
    The test accuracy shows that the 
    reverse cross-entropy is more robust to labeling errors. 
    \label{fig:robustness_to_uncertainty}}
\end{figure}

We derive our self-labeling surrogate by
{\em splitting}\cite{boyd2004convex} the regularized entropy clustering loss \eqref{eq:rec} into two groups: regularized decisiveness and fairness. The latter, expressed by KL-divergence with distribution $u$ as in \eqref{eq:mi+decay}, uses auxiliary splitting variables $y\in\Delta^K$. Such soft pseudo-labels $y$ are tied to predictions $\sigma$ by an extra divergence term 
(the last one)
\begin{equation} \nonumber
    \gamma\, \|\wc\|^2 \;\;+\;
    \overline{H(\sigma)} \; + \; KL(\overline{y}\,\|\,u) \;+\; 
    \overline{KL(\sigma,y)}.
\end{equation}
Combining the entropy with the last KL term gives
\begin{equation} \label{eq:semm2}
    \gamma\, \|\wc\|^2 \;\;+\;  \overline{H(\sigma,y)} \; + \; KL(\overline{y}\,\|\,u)
\end{equation}
which has one notable difference from 
the standard self-labeling surrogate losses \eqref{eq:vidaldi} and \eqref{eq:ismail} from \cite{YM.2020Self-labelling,ismail2021}:
the cross-entropy term reverses the order of its arguments
$y$ and $\sigma$.
The standard order is motivated in the supervised
settings where $y$ is a true one-hot target and $\sigma$
is the estimated distribution. However, this motivation is questionable when both $\sigma$ and $y$ are {\em estimated} soft distributions. Moreover, Figure~\ref{fig:ce-forward-reverse}(b) argues that the reverse cross-entropy can improve the robustness of network predictions $\sigma$ to errors in estimated pseudo-labels $y$, as confirmed by our training experiments in Figure \ref{fig:robustness_to_uncertainty}. 
The reversal also works for estimating soft pseudo-labels $y$ as 
the second argument in the cross-entropy function is standard for an ``estimated'' distribution.

We also observe that the standard fairness term in (\ref{eq:semm2},\ref{eq:mi+decay},\ref{eq:ismail}) is the {\em reverse} KL divergence w.r.t. cluster volumes, i.e. the average predictions $\bar{\sigma}$ or pseudo-labels $\bar{y}$. 
It can tolerate highly unbalanced solutions where $\bar{y}_k=0$ for some cluster $k$, 
see the dashed curves in Figure~\ref{fig:ce-forward-reverse}(a). 
We propose the {\em forward}, a.k.a. {\em zero-avoiding}, KL divergence $KL(u\,\|\,\overline{y})$, see the solid curves Figure~\ref{fig:ce-forward-reverse}(a),
which assigns infinite penalties to highly unbalanced trivial clusterings.
We refer to this as {\em strong fairness}. 
Thus, our final self-labeling surrogate loss is 
\begin{equation}
  \label{eq:semm_our}
      L_{our}\;\; :=  \;\;\gamma\, \|\wc\|^2 \;\;+\;  \overline{H(\sigma,y)} \;\;+\; \lambda\,
      KL(u\,\|\,\overline{y}). 
\end{equation}

Due to the argument reversal in the cross-entropy and KL terms, our
loss \eqref{eq:semm_our} represents stronger constraints for $y$ and $\bar{y}$ when compared to \eqref{eq:ismail}. However, besides well-motivated numerical properties, it also matters that \eqref{eq:semm_our} has an efficient solver for pseudo-labels discussed in the next Section.

\subsection{Our EM algorithm for pseudo-labels} 
\label{sec:EM}

Efficient minimization of a self-supervised loss w.r.t pseudo-labels $y$ for given predictions $\sigma$ is a critical component of all iterative self-labeling techniques \cite{YM.2020Self-labelling,ismail2021}, see Sec.\ref{sec:related_work}. 
While our loss \eqref{eq:semm_our} is convex w.r.t. $y$, optimization is done 
over the probability simplex and a good practical solver is not a given. Note that $H(\sigma,y)$ works as a {\em log barrier} for the constraint $y\in\Delta^K$. This could be problematic for the first-order methods, but a basic Newton's method is a good match, e.g. \cite{kelley1995iterative}. The overall convergence rate of such second-order methods is fast, but computing the Hessian's inverse is costly. Below we derive an {\em expectation-maximization} (EM) algorithm that is more efficient, see Table~\ref{table: speed comparison}.
\begin{table}[h]
\centering
\noindent
\resizebox{0.4\textwidth}{!}{\begin{tabular}{p{3cm} p{1.2cm} p{1.2cm} p{1.2cm} p{1.2cm} p{1.2cm} p{1.2cm} }\toprule
\multicolumn{1}{c}{\textbf{}} & \multicolumn{3}{c}{\textbf{number of iterations}} & \multicolumn{3}{c}{\textbf{running time in sec.}} \\  
\multicolumn{1}{c}{\textbf{}} & \multicolumn{3}{c}{\textbf{(to convergence)}} & \multicolumn{3}{c}{\textbf{(to convergence)}}\\ 
\cmidrule(lr){2-4}
\cmidrule(lr){5-7}

\multicolumn{1}{c}{\textbf{K}} & $\bf 2$ & $\bf 20$ & $\bf 200$ & $\bf 2$ & $\bf 20$ & $\bf 200$ \\ 
\cmidrule(lr){1-1}
\cmidrule(lr){2-4}
\cmidrule(lr){5-7}

\multicolumn{1}{c}{Newton} & $3$ & $3$ & $4$ & $2.8e^{-2}$ & $3.3e^{-2}$ & $1.7e^{-1}$  \\\noalign{\smallskip} 
\cmidrule(lr){1-1}
\cmidrule(lr){2-4}
\cmidrule(lr){5-7}

\multicolumn{1}{c}{EM} & $2$ & $2$ & $2$ & $9.9e^{-4}$ & $2.0e^{-3}$ & $4.0e^{-3}$  \\ 
\bottomrule
\end{tabular}}
\caption{Our EM algorithm vs Newton's methods \cite{kelley1995iterative}.}
\label{table: speed comparison}
\end{table} 

Assume that model parameters and predictions in \eqref{eq:semm_our} 
are fixed, {\em i.e.} $\wc$ and $\sigma$.
Following {\em variational inference} \cite{bishop:2006}, we introduce $K$ auxiliary latent variables, 
distributions $S^k\in\Delta^N$ representing normalized support of each cluster $k$ over $N$ data points. 
In contrast, $N$ distributions $y_i\in\Delta^K$ show support for each class at every point $X_i$. 
We refer to each vector $S^k$ as a {\em normalized cluster} $k$. 
Note that here we focus on individual data points and explicitly index them by $i\in\{1,\dots,N\}$. 
Thus, we use $y_i \in\Delta^K$ and $\sigma_i\in\Delta^K$. Individual components of distribution $S^k\in\Delta^N$ 
corresponding to data point $X_i$ is denoted by scalar $S^k_i$.

First, we expand our loss \eqref{eq:semm_our} using our new latent variables $S^k\in\Delta^N$
\begin{align}
\label{eq:EM initial}
L_{our}\;\;&\eqc\;\;\overline{H(\sigma,y)}+\lambda\,H(u,\bar{y})+\gamma\,\|\wc\|^2 \\\nonumber
&=\;\;\overline{H(\sigma,y)}-\lambda\,\sum_{k}u^k\ln{\sum_{i}S_i^k\frac{y_i^k}{S_i^k N}}+\gamma\,\|\wc\|^2\\
&\leq\;\;\overline{H(\sigma,y)}-\lambda\,\sum_{k}\sum_{i}u^k S_i^k \ln{\frac{y_i^k}{S_i^k N}}+\gamma\,\|\wc\|^2 \label{eq:EM derivation}
\end{align}

Due to the convexity of negative $\log$, we apply Jensen's inequality to derive an upper bound, i.e.\ \eqref{eq:EM derivation}, to $L_{our}$. Such a bound becomes tight when:
\begin{equation}
\label{eq:E step}
\text{E-step}:\;\;\;\;\;\;\;\;\;\;\;\;\;\;\;\;\;\;S_i^k=\frac{y_i^k}{\sum_{j}y_j^k}\;\;\;\;\;\;\;\;\;\;\;\;\;\;\;\;\;\;\;\;\;\;\;\;\;\;\end{equation}
Then, we fix $S_i^k$ as \eqref{eq:E step} and solve the Lagrangian of \eqref{eq:EM derivation} with simplex constraint to update $y$ as:
\begin{equation}
\centering
\label{eq:M step}
\text{M-step}:\;\;\;\;\;\;\;\;\;\;\;\;y_i^k=\frac{\sigma_i^k+\lambda N u^k S_i^k}{1+\lambda N\sum_{c}u^c S_i^c}\;\;\;\;\;\;\;\;\;\;\;\;\;\;
\end{equation}
We run these two steps until convergence with respect to some predefined tolerance. Note that the minimum $y$ is guaranteed to be globally optimal since \eqref{eq:EM initial} is convex w.r.t. $y$.
 The empirical convergence rate is within 15 steps on MNIST. The comparison of computation speed on synthetic data is shown in Table~\ref{table: speed comparison}. While the number of iterations to convergence is roughly the same as Newton's methods, our EM algorithm is much faster in terms of running time and is extremely easy to implement using the highly optimized built-in functions from the standard PyTorch library that supports GPU.

\begin{figure}[t]
    \centering
    \includegraphics[width=\linewidth]{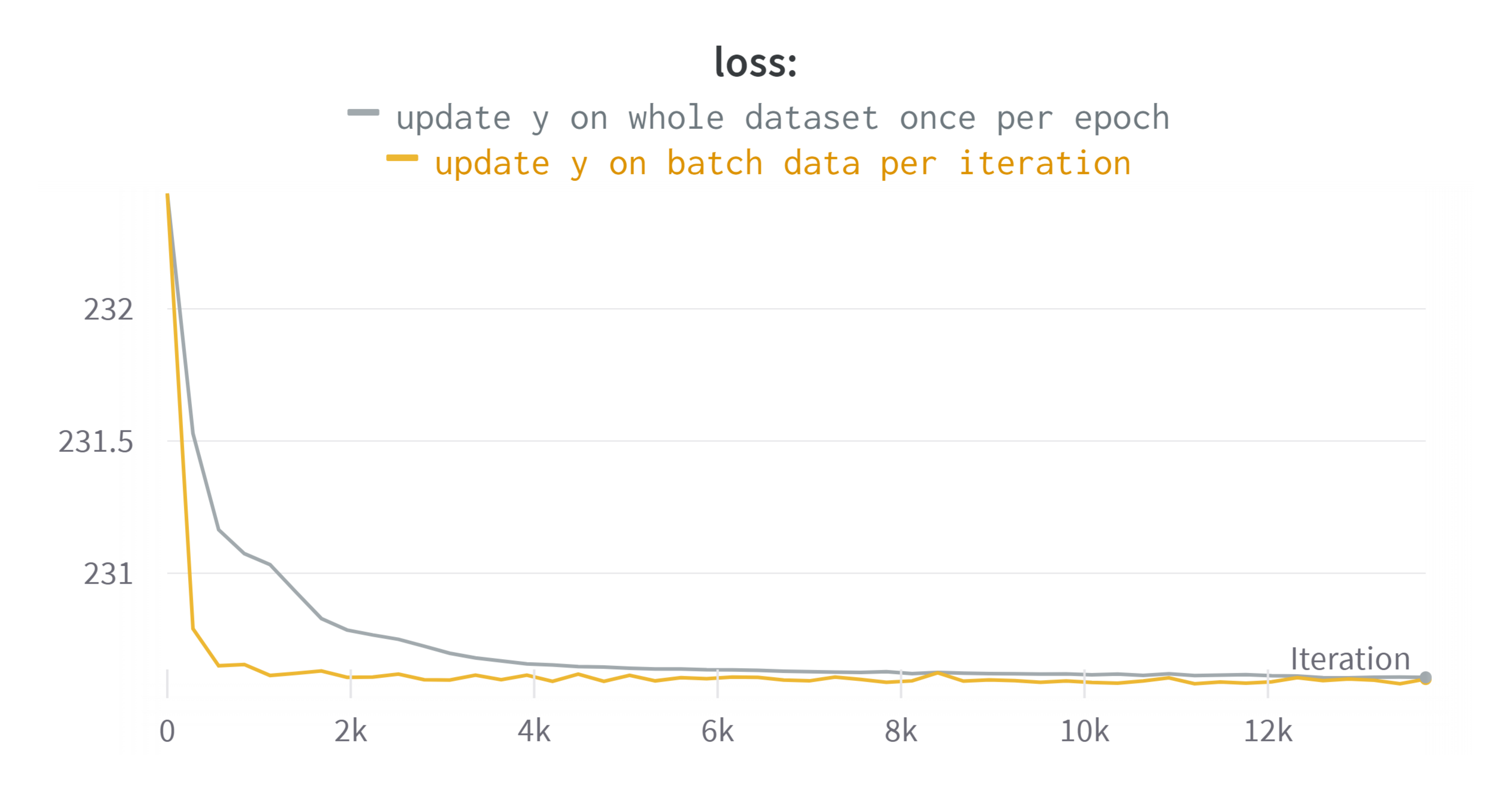}
    \caption{Loss \eqref{eq:semm_our} for two different update strategies for $y$. These plots are generated with a linear classifier on MNIST. We use the same initialization and run both strategies for 50 epochs. 
    Note that once-per-epoch updates (corresponding to the gray curve) achieve 52\% accuracy, while once-per-batch updates 
    (corresponding to the yellow curve) achieve 63\%.}
    \label{fig:full vs batch}
\end{figure}




Inspired by \cite{springenberg2015unsupervised,hu2017learning}, we also adapted our EM algorithm to allow for updating $y$ within each batch. In fact, the mini-batch approximation of \eqref{eq:EM initial} is an upper bound. Considering the first two terms of \eqref{eq:EM initial}, we can use Jensen's inequality
\begin{equation}
\label{eq:batch approx}
\overline{H(\sigma,y)}+\lambda\,H(u,\bar{y})\;\;\leq\;\;\mathbb{E}_B[\overline{H_B(\sigma,y)}+\lambda\,H(u,\bar{y}_B)]
\end{equation}
where $B$ is the batch randomly sampled from the whole dataset and the bar operator is the average over data points in the batch.
Now, we can apply our EM algorithm to update $y$ in each batch, which is even more efficient. Compared to other methods \cite{ghasedi2017deep,YM.2020Self-labelling,ismail2021} using auxiliary variables $y$, we can efficiently update $y$ on the fly while they only update once or just a few times per epoch due to the cost of computing $y$ for the whole dataset. Interestingly, we found that it is important to update $y$ on the fly, which makes convergence faster and improves the performance significantly, as shown in Figure~\ref{fig:full vs batch}. We use this ``batch version'' EM throughout all the experiments. Our full algorithm for the loss \eqref{eq:semm_our} is summarized as follows.

\SetKwInOut{KwInput}{Input}
\SetKwInOut{KwOutput}{Output}
\begin{algorithm}
\caption{Optimization for our loss}
\label{alg:our alg}
\KwInput{network parameters $[\wc,\wf]$ and dataset}
\KwOutput{network parameters $[\wc^*,\wf^*]$}
\For{each epoch}{
\For{each iteration}{
Initialize $y$ by the network output at current stage as a warm start\;\\
\While{not convergent}{
$S_i^k=\frac{y_i^k}{\sum_{j}y_j^k}$\;
$y_i^k=\frac{\sigma_i^k+\lambda N u^k S_i^k}{1+\lambda N\sum_{c}u^c S_i^c}$\;
}
Update $[\wf,\wc]$ using loss $\overline{H_B(\sigma,y)}+\gamma\,\|\wc\|^2$ via stochastic gradient descent
}}
\end{algorithm}

\subsection{Discussion}
\label{sec: algorithm discussion}
Here we briefly compare our optimization method for \eqref{eq:semm_our} and the algorithm for 
\eqref{eq:rec_ismail} in \cite{ismail2021}. Their loss is convex with respect to $\wc$. 
In contrast, the reverse cross-entropy in our loss \eqref{eq:semm_our} is non-convex as a natural consequence of 
our focus on robustness to pseudo-label errors. However, both losses are non-convex w.r.t. parameters $\wf$ 
of the interior network layers responsible for representation $f_\wf(X)$ in \eqref{eq:postmodel_deep}.
As for the optimization w.r.t. the pseudo-labels $y$, both losses are convex. Our EM algorithm 
converges to the global optima for $y$, while they use an approximate closed-form solution for $y$, but it does not have 
optimality guarantees. Both self-labeling algorithms iteratively optimize the corresponding surrogate losses 
over pseudo-labels $y$ and model parameters $\wc$, $\wf$. 

\section{Experimental results} \label{sec:experiments}
We test our approach to clustering on several standard datasets using different network architectures. We also evaluate different losses in semi-supervised settings. All the results for prior work are reproduced using either the corresponding public code or our implementation using the same experimental settings. 

\textbf{Datasets:} For the clustering problem, we use four standard benchmark datasets: MNIST \cite{MNIST}, CIFAR10/100 \cite{CIFAR} and STL10 \cite{STL}. We follow \cite{ji2019invariant} to use the whole dataset for training and testing unless otherwise specified.
As for the semi-supervised setting, we conduct experiments on CIFAR10 and STL10. We split the data into training and test sets as suggested by the official instructions of the datasets. 


\textbf{Evaluation:}
To evaluate the quality of clustering, we set the number of clusters to the number of ground-truth category labels. To calculate the accuracy, we use the 
standard Hungarian algorithm \cite{kuhn1955hungarian} to find the best one-to-one mapping between predicted clusters and ground-truth labels. We don't need this matching step if we use other metrics, i.e. NMI, and ARI. For the evaluation of semi-supervised classification, we directly calculate the accuracy of the test set without the mapping step.

\textbf{Implementation details:}
For clustering with fixed features, we set $\lambda$ in our loss to 100. The learning rate of the stochastic gradient descent is set to 0.1 and training takes 10 epochs. The batch size was set to 250. The coefficients for the $l_2$ norm of the weights are set to 0.001, 0.02, 0.009, and 0.02 for MNIST, CIFAR10, CIFAR100, and STL10 respectively. We also tuned the hyperparameters for all other methods.

 As for the training of VGG4 in Section \ref{sec: deep clustering}, we use Adam \cite{kingma2015adam} with learning rate $1e^{-4}$ for optimizing the network parameters. We set the batch size to 250 for CIFAR10, CIFAR100, and MNIST, and we used 160 for STL10.
 For each image, we generate two augmentations sampled from ``horizontal flip", ``rotation" and ``color distortion".
We set the $\lambda$ to 100 in our loss.
The weight decay coefficient is set to 0.01.
We report the mean accuracy and Std from 6 runs with different initializations while we use the same initialization for all methods in each run. We use 50 epochs for each run and all methods with tuned hyperparameters reach convergence within 50 epochs. 
As for the training of ResNet-18, we still use the Adam optimizer, and the learning rate is set to $1e^{-1}$ for the linear classifier and $1e^{-5}$ for the backbone. The weight decay coefficient is set to $1e^{-4}$. The batch size is 200 and the number of total epochs is 50. The $\lambda$ is still set to 100. We only use one augmentation per image, and the coefficient for the augmentation term is set to 0.5, 0.2 and 0.4 respectively for STL10, CIFAR10, and CIFAR100. Such coefficient is tuned separately for IMSAT and MIADM.


\subsection{Effect of Renyi entropy order $\alpha$}
Our results in Section \ref{sec:SVM} state that the regularized decisiveness \eqref{eq:reg_decisiveness_R} using Renyi entropy of any order $\alpha>0$ leads to margin maximization under certain conditions, e.g. linear model and convergence of the regularization constant $\gamma$ to zero. However, such conditions may not be satisfied in practice. For example, only fixed values of $\gamma$ can be evaluated. Also, unsurprisingly, better results are typically obtained by non-linear network models where a linear classifier head is trained jointly with fine-tuning the representation layers. The corresponding non-convex losses may also be harder or easier to optimize depending on $\alpha$.

It follows that different values of $\alpha$ may affect the practical results. However, our self-labeling algorithm in Section \ref{sec:our_approach} assumes $\alpha=1$ and uses the closed-form EM steps applicable only to the Shannon's decisiveness. To evaluate Renyi decisiveness for various $\alpha>0$, we apply the standard stochastic gradient descent to a combination of \eqref{eq:reg_decisiveness_R} with Shannon's fairness
\begin{equation} \label{eq:rec2}
    \gamma \|\wc\|^2 \;+\; \overline{R_\alpha(\sigma)} \;-\;  H(\overline{\sigma}). 
\end{equation}
We use the standard STL10 dataset following the experimental setting for clustering from \cite{van2020scan}. 
All hyperparameters are separately tuned in each column. 
\begin{table}[ht]
\begin{center}
\resizebox{0.48\textwidth}{!}{
\begin{tabular}{cccccc}
\hline\noalign{\smallskip}
 $\alpha$ & 0.1 & 0.5 & 1 & 5 & $\infty$ \\\hline\noalign{\smallskip}
 ACC & 62\% (3.5) & 63.45\% (3.2) & 70.23\% (2.0) & 71.05\% (1.6) & 73.34\% (1.6) \\\hline\noalign{\smallskip}
\end{tabular}}
\end{center}
\caption{Effect of order $\alpha$ in Renyi decisiveness \eqref{eq:reg_decisiveness_R}. The results use ResNet18 trained via standard stochastic gradient descent for loss \eqref{eq:rec2}.
ACC for $\alpha=1$ is the same as for IMSAT \cite{hu2017learning} in Tab.\ref{table: comparison to SOTA} (STL10).}
\label{table: numbers effects of alpha}
\end{table}

Table \ref{table: numbers effects of alpha} suggests that larger values of $\alpha$ may lead to better empirical results. Intuitively, this could be explained as follows. 
Smaller $\alpha$ induces larger gradients for the Renyi decisiveness, see Figure \ref{fig:renyi_a}, as the network prediction becomes more confident during training. 
This creates strong local minima for the network parameters that are hard to escape. 

The results in Table \ref{table: numbers effects of alpha} encourage the development of an efficient pseudo-labeling algorithm for $\alpha>1$, 
which is left for future work.
The experiments in the following Sections use our efficient EM-based algorithm from Section \ref{sec:our_approach}  assuming Shannon's entropy, i.e. $\alpha=1$. 
Note that our self-labeling optimization algorithm significantly outperforms
the standard stochastic gradient descent for $\alpha=1$, see "our" vs IMSAT in Table \ref{table: comparison to SOTA}.

\subsection{Clustering with fixed features}
\label{sec:low level clustering}

In this section, we test our loss as a proper clustering
loss and compare it to the widely used Kmeans (generative) and other closely related losses (entropy-based and discriminative). We use the pre-trained 
(ImageNet) Resnet-50 \cite{he2016deep} to extract the features.
For Kmeans, the model is parameterized by K cluster centers. Comparably, we use a one-layer linear classifier followed by softmax for all other losses including ours. Kmeans results were obtained using the scikit-learn package in Python. To optimize the model parameters for other losses, we use stochastic gradient descent. Here we report the average accuracy and standard deviation over 6 randomly initialized trials in Table~\ref{table: numbers fixed features}.

\begin{table}[ht]
\begin{center}
\resizebox{0.4\textwidth}{!}{
\begin{tabular}{ccccc}
\hline\noalign{\smallskip}
  & STL10 & CIFAR10 & CIFAR100 (20) & MNIST \\
\noalign{\smallskip}
\hline
\noalign{\smallskip}

$\text{K-means}$  & 85.20\%(5.9)  & 67.78\%(4.6) & 42.99\%(1.3) & 47.62\%(2.1) \\\noalign{\smallskip}
$\text{MIGD}$~\cite{MacKay1991,Perona2010} & 89.56\%(6.4) & 72.32\%(5.8) & 43.59\%(1.1) & 52.92\%(3.0) \\\noalign{\smallskip}
$\text{SeLa}$~\cite{YM.2020Self-labelling} & 90.33\%(4.8) & 63.31\%(3.7) & 40.74\%(1.1) & 52.38\%(5.2)\\\noalign{\smallskip}
$\text{MIADM}$~\cite{ismail2021} & 81.28\%(7.2) & 56.07\%(5.5) & 36.70\%(1.1) & 47.15\%(3.7)\\\noalign{\smallskip}
$\text{MIADM}\star$~\cite{ismail2021} & 88.64\%(7.1) & 60.57\%(3.3) & 41.2\%(1.4) & 50.61\%(1.3)\\\noalign{\smallskip}
\hline\noalign{\smallskip}
$\text{Our}$ & \bf 92.2\%(6.2) & \bf 73.48\%(6.2) & \bf 43.8\%(1.1)& \bf 58.2\%(3.1) \\

\hline
\end{tabular}}
\end{center}
\caption{Comparison of different methods using fixed features. The numbers are the average accuracy and the standard deviation over 6 trials. $\star$: our ``batch version" implementation of their method.}
\label{table: numbers fixed features}
\end{table}


\subsection{Joint clustering and feature learning}
\label{sec: deep clustering}
In this section, we train a deep network to jointly learn the features and cluster the data. We test our method on both a small architecture (VGG4) and a large one (ResNet-18). The only extra standard technique we add here is self-augmentation as discussed in Section \ref{sec:into_reg}.

To train the VGG4, we use random initialization for network parameters. 
From Table~\ref{table:numbers}, it can be seen that our approach consistently achieves the most competitive results in terms of accuracy (ACC). Most of the methods we compared in our work (including our method) are general concepts applicable to single-stage end-to-end training. To be fair, we tested all of them on the same simple architecture. But, these general methods can be easily integrated into other more complex systems with larger architecture such as ResNet-18.

\setlength{\tabcolsep}{6pt}
\begin{table}[ht]
\begin{center}\resizebox{0.4\textwidth}{!}{
\begin{tabular}{ccccc}
\hline\noalign{\smallskip}
  & STL10 & CIFAR10 & CIFAR100 (20) & MNIST \\
\noalign{\smallskip}
\hline
\noalign{\smallskip}

$\text{IMSAT}$~\cite{hu2017learning}  & 25.28\%(0.5)  & 21.4\%(0.5) & 14.39\%(0.7) &92.90\%(6.3) \\\noalign{\smallskip}
$\text{IIC}$~\cite{ji2019invariant} & 24.12\%(1.7) & 21.3\%(1.4) & 12.58\%(0.6) & 82.51\%(2.3) \\\noalign{\smallskip}
$\text{SeLa}$~\cite{YM.2020Self-labelling} & 23.99\%(0.9) & 24.16\%(1.5) & \bf 15.34\%(0.3) & 52.86\%(1.9)\\\noalign{\smallskip}
$\text{MIADM}$~\cite{ismail2021} & 17.37\%(0.9) & 17.27\%(0.6) & 11.02\%(0.5) & 17.75\%(1.3)\\\noalign{\smallskip}
$\text{MIADM}\star$~\cite{ismail2021} & 23.37\%(0.9) & 23.26\%(0.6) & 14.02\%(0.5) & 78.88\%(3.3)\\\noalign{\smallskip}\hline\noalign{\smallskip}
$\text{Our}$ & \bf 25.33\%(1.4) & \bf 24.16\%(0.8) & 15.09\%(0.5) & \bf 93.58\%(4.8)\\

\hline
\end{tabular}}
\end{center}
\caption{Quantitative results of accuracy for unsupervised clustering methods with VGG4. We only use the 20 coarse categories for CIFAR100 following \cite{ji2019invariant}.}
\label{table:numbers}
\end{table}

As for the training of ResNet-18, we found that random initialization does not work well. Following \cite{van2020scan}, we use the pre-trained weight from the self-supervised learning. For a fair comparison, we followed their experimental settings and compared ours to their second-stage results. Note that they split the data into training and testing. We also report two additional evaluation metrics, i.e. NMI and ARI.

In Table~\ref{table: comparison to SOTA}, we show the results using their pretext-trained network as initialization for our entropy clustering. We use only our clustering loss together with the self-augmentation (one augmentation per image this time) to reach higher numbers than SCAN, as shown in the table below. More importantly, we consistently improve over the most related method, MIADM, by a large margin, which clearly demonstrates the effectiveness of our proposed loss together with the optimization algorithm. 

\begin{table}[ht]
\centering
\noindent
\begin{center}\resizebox{0.45\textwidth}{!}{\begin{tabular}{p{3cm} p{1.2cm} p{1.2cm} p{1.2cm} p{1.2cm} p{1.2cm} p{1.2cm} p{1.2cm} p{1.2cm} p{1.2cm}}\toprule
\multicolumn{1}{c}{\textbf{}} & \multicolumn{3}{c}{CIFAR10} & \multicolumn{3}{c}{CIFAR100 (20)} & \multicolumn{3}{c}{STL10} \\  
\cmidrule(lr){2-4}
\cmidrule(lr){5-7}
\cmidrule(lr){8-10}

\multicolumn{1}{c}{\textbf{}} &  ACC &  NMI &  ARI &  ACC &  NMI &  ARI &  ACC &  NMI &  ARI\\ 
\cmidrule(lr){1-10}

\multicolumn{1}{c}{SCAN \cite{van2020scan}} & 81.8 (0.3) & 71.2 (0.4) & 66.5 (0.4) & 42.2 (3.0) & \bf 44.1 (1.0) & 26.7 (1.3) & 75.5 (2.0) & 65.4 (1.2) & 59.0 (1.6)\\\noalign{\smallskip} 

\multicolumn{1}{c}{IMSAT \cite{hu2017learning}} & 77.64\% (1.3) & 71.05\% (0.4) & 64.85\% (0.3) & 43.68\% (0.4) & 42.92\% (0.2) & 26.47\% (0.1) & 70.23\% (2.0) & 62.22\% (1.2) & 53.54\% (1.1)\\\noalign{\smallskip} 

\multicolumn{1}{c}{MIADM$\star$ \cite{ismail2021}} & 74.76\% (0.3) & 69.17\% (0.2) & 62.51\% (0.2) & 43.47\% (0.5) & 42.85\% (0.4) & 27.78\% (0.4) & 67.84\% (0.2) & 60.33\% (0.5) & 51.67\% (0.6)\\\noalign{\smallskip} 
\cmidrule(lr){1-10}

\multicolumn{1}{c}{Our} & \bf 83.09 (0.2) & \bf 71.65 (0.1) & \bf 68.05 (0.1) & \bf 46.79 (0.3) & 43.27 (0.1) & \bf 28.51 (0.1) & \bf 77.67 (0.1) & \bf 67.66 (0.3) & \bf 61.26 (0.4)  \\ 
\bottomrule
\end{tabular}}
\end{center}
\caption{Quantitative comparison using network ResNet-18.}
\label{table: comparison to SOTA}
\end{table}

\subsection{Semi-supervised classification} 
\label{sec: SS classification}
Although our paper is focused on self-labeled classification, we find it also interesting and natural to test our loss under a semi-supervised setting where partial data is provided with ground-truth labels. We use the standard cross-entropy loss for labeled data and directly add it to the self-labeled loss to train the network initialized by the pretext-trained network following \cite{van2020scan}. Note that the pseudo-labels for the labeled data are constrained to be the ground-truth labels.

\begin{table}[h]
\centering
\noindent
\resizebox{0.45\textwidth}{!}{
\begin{tabular}{P{3cm} P{1.5cm} P{1.5cm} P{1.5cm} P{1.5cm} P{1.5cm} P{1.5cm}}\toprule
\multicolumn{1}{c}{\textbf{}} & \multicolumn{2}{c}{\textbf{0.1}} & \multicolumn{2}{c}{\textbf{0.05}} & \multicolumn{2}{c}{\textbf{0.01}} \\  
\cmidrule(lr){2-3}
\cmidrule(lr){4-5}
\cmidrule(ll){6-7}
\multicolumn{1}{c}{\textbf{}} & \textbf{STL10} & \textbf{CIFAR10} & \textbf{STL10} & \textbf{CIFAR10} & \textbf{STL10} & \textbf{CIFAR10} \\ 
\cmidrule(lr){1-7}
\multicolumn{1}{c}{Only seeds} & 78.4\% & 81.2\% & 74.1\% & 76.8\% & 68.8\% & 71.8\% \\\noalign{\smallskip} 
\multicolumn{1}{c}{+ IMSAT~\cite{hu2017learning}} & 88.1\% & 91.5\% & 81.1\% & 85.2\% & 74.1\% & 80.2\% \\\noalign{\smallskip} 
\multicolumn{1}{c}{+ IIC~\cite{ji2019invariant}} & 85.2\% &  90.3\% & 78.2\% & 84.8\% & 72.5\% & 80.5\% \\ \noalign{\smallskip}
\multicolumn{1}{c}{+ SeLa~\cite{YM.2020Self-labelling}} & 86.2\% & 88.6\% & 79.5\% & 82.7\% & 69.9\% & 79.1\% \\ \noalign{\smallskip}
\multicolumn{1}{c}{+ MIADM~\cite{ismail2021}} & 84.9\% & 86.1\% & 77.9\% & 80.1\% & 69.6\% & 77.5\% \\ \noalign{\smallskip}
\cmidrule(lr){1-7}
\multicolumn{1}{c}{+ Our} & \bf 88.7\% & \bf 92.2\% & \bf 82.6\% & \bf 85.8\% & \bf 75.6\% & \bf 81.9\% \\ 
\bottomrule
\end{tabular}}
\caption{Quantitative results for semi-supervised classification on STL10 and CIFAR10 using ResNet18. The numbers 0.1, 0.05 and 0.01 correspond to different ratio of labels used for supervision. ``Only seeds'' means we only use standard cross-entropy loss on seeds for training.}
\label{table: numbers weakly supervised}
\end{table}

\section{Conclusions} \label{sec:conclusions}
Our paper clarified several important conceptual properties of the general discriminative entropy clustering methodology. We formally proved that linear entropy clustering has a maximum margin property, establishing a conceptual relation to
SVM clustering. 
Unlike prior work on discriminative entropy clustering, we show that classifier norm regularization is important 
for margin maximization.
We also provided counterexamples disproving a recent theoretical claim of the equivalence between {\em variance clustering} (soft K-means) and {\em linear entropy clustering} (linear softmax classifier optimizing decisiveness and fairness). We juxtapose the two methods as generative and discriminative approaches to MI-based clustering \cite{MacKay1991}. 

We also discussed several limitations of the existing self-labeling formulations of discriminative entropy clustering 
and proposed a new loss addressing such limitations. In particular, we replace the standard (forward) 
cross-entropy by the {\em reverse cross-entropy} that we show is significantly more robust to errors in estimated soft pseudo-labels. Our loss also uses a strong formulation of the fairness constraint motivated by a {\em zero-avoiding} version of KL divergence. We designed an efficient EM algorithm for minimizing our loss w.r.t. pseudo-labels; it is significantly faster than standard alternatives, e.g. Newton's method. 
Our empirical results improved the state-of-the-art on many standard benchmarks for end-to-end deep clustering methods.

\bibliography{PAMI}

\begin{thebibliography}{10}
\providecommand{\url}[1]{#1}
\csname url@samestyle\endcsname
\providecommand{\newblock}{\relax}
\providecommand{\bibinfo}[2]{#2}
\providecommand{\BIBentrySTDinterwordspacing}{\spaceskip=0pt\relax}
\providecommand{\BIBentryALTinterwordstretchfactor}{4}
\providecommand{\BIBentryALTinterwordspacing}{\spaceskip=\fontdimen2\font plus
\BIBentryALTinterwordstretchfactor\fontdimen3\font minus \fontdimen4\font\relax}
\providecommand{\BIBforeignlanguage}[2]{{%
\expandafter\ifx\csname l@#1\endcsname\relax
\typeout{** WARNING: IEEEtran.bst: No hyphenation pattern has been}%
\typeout{** loaded for the language `#1'. Using the pattern for}%
\typeout{** the default language instead.}%
\else
\language=\csname l@#1\endcsname
\fi
#2}}
\providecommand{\BIBdecl}{\relax}
\BIBdecl

\bibitem{MacKay1991}
J.~S. Bridle, A.~J.~R. Heading, and D.~J.~C. MacKay, ``Unsupervised classifiers, mutual information and 'phantom targets','' in \emph{NIPS}, 1991, pp. 1096--1101.

\bibitem{Perona2010}
A.~Krause, P.~Perona, and R.~Gomes, ``Discriminative clustering by regularized information maximization,'' \emph{Advances in neural information processing systems}, vol.~23, 2010.

\bibitem{hu2017learning}
W.~Hu, T.~Miyato, S.~Tokui, E.~Matsumoto, and M.~Sugiyama, ``Learning discrete representations via information maximizing self-augmented training,'' in \emph{International conference on machine learning}.\hskip 1em plus 0.5em minus 0.4em\relax PMLR, 2017, pp. 1558--1567.

\bibitem{chang2017deep}
J.~Chang, L.~Wang, G.~Meng, S.~Xiang, and C.~Pan, ``Deep adaptive image clustering,'' in \emph{Proceedings of the IEEE international conference on computer vision}, 2017, pp. 5879--5887.

\bibitem{ji2019invariant}
X.~Ji, J.~F. Henriques, and A.~Vedaldi, ``Invariant information clustering for unsupervised image classification and segmentation,'' in \emph{Proceedings of the IEEE/CVF International Conference on Computer Vision}, 2019, pp. 9865--9874.

\bibitem{YM.2020Self-labelling}
Y.~M. Asano, C.~Rupprecht, and A.~Vedaldi, ``Self-labelling via simultaneous clustering and representation learning,'' in \emph{International Conference on Learning Representations}, 2020.

\bibitem{ismail2021}
M.~Jabi, M.~Pedersoli, A.~Mitiche, and I.~B. Ayed, ``Deep clustering: On the link between discriminative models and k-means,'' \emph{IEEE Transactions on Pattern Analysis and Machine Intelligence}, vol.~43, no.~6, pp. 1887--1896, 2021.

\bibitem{jiang2017variational}
Z.~Jiang, Y.~Zheng, H.~Tan, B.~Tang, and H.~Zhou, ``Variational deep embedding: an unsupervised and generative approach to clustering,'' in \emph{Proceedings of the 26th International Joint Conference on Artificial Intelligence}, 2017, pp. 1965--1972.

\bibitem{yang2017towards}
B.~Yang, X.~Fu, N.~D. Sidiropoulos, and M.~Hong, ``Towards k-means-friendly spaces: Simultaneous deep learning and clustering,'' in \emph{international conference on machine learning}.\hskip 1em plus 0.5em minus 0.4em\relax PMLR, 2017, pp. 3861--3870.

\bibitem{caron2018deep}
M.~Caron, P.~Bojanowski, A.~Joulin, and M.~Douze, ``Deep clustering for unsupervised learning of visual features,'' in \emph{Proceedings of the European conference on computer vision (ECCV)}, 2018, pp. 132--149.

\bibitem{zhu1996region}
S.~C. Zhu and A.~Yuille, ``Region competition: Unifying snakes, region growing, and bayes/mdl for multiband image segmentation,'' \emph{IEEE transactions on pattern analysis and machine intelligence}, vol.~18, no.~9, pp. 884--900, 1996.

\bibitem{chan2001active}
T.~F. Chan and L.~A. Vese, ``Active contours without edges,'' \emph{IEEE Transactions on image processing}, vol.~10, no.~2, pp. 266--277, 2001.

\bibitem{rother2004grabcut}
C.~Rother, V.~Kolmogorov, and A.~Blake, ``" grabcut" interactive foreground extraction using iterated graph cuts,'' \emph{ACM transactions on graphics (TOG)}, vol.~23, no.~3, pp. 309--314, 2004.

\bibitem{Hinton1986learning}
D.~E. Rumelhart, G.~E. Hinton, and R.~J. Williams, ``Learning representations by back-propagating errors,'' \emph{Nature}, vol. 323, no. 6088, pp. 533--536, 1986.

\bibitem{bengio2004semi}
Y.~Grandvalet and Y.~Bengio, ``Semi-supervised learning by entropy minimization,'' \emph{Advances in neural information processing systems}, vol.~17, 2004.

\bibitem{Schuurmans2004}
L.~Xu, J.~Neufeld, B.~Larson, and D.~Schuurmans, ``Maximum margin clustering,'' in \emph{Advances in Neural Information Processing Systems}, L.~Saul, Y.~Weiss, and L.~Bottou, Eds., vol.~17.\hskip 1em plus 0.5em minus 0.4em\relax MIT Press, 2004.

\bibitem{boudiaf2020unifying}
M.~Boudiaf, J.~Rony, I.~M. Ziko, E.~Granger, M.~Pedersoli, P.~Piantanida, and I.~B. Ayed, ``A unifying mutual information view of metric learning: cross-entropy vs. pairwise losses,'' in \emph{European conference on computer vision}.\hskip 1em plus 0.5em minus 0.4em\relax Springer, 2020, pp. 548--564.

\bibitem{hjelm2018learning}
R.~D. Hjelm, A.~Fedorov, S.~Lavoie-Marchildon, K.~Grewal, P.~Bachman, A.~Trischler, and Y.~Bengio, ``Learning deep representations by mutual information estimation and maximization,'' in \emph{International Conference on Learning Representations}, 2018.

\bibitem{oord2018representation}
A.~v.~d. Oord, Y.~Li, and O.~Vinyals, ``Representation learning with contrastive predictive coding,'' \emph{arXiv preprint arXiv:1807.03748}, 2018.

\bibitem{tschannen2019mutual}
M.~Tschannen, J.~Djolonga, P.~K. Rubenstein, S.~Gelly, and M.~Lucic, ``On mutual information maximization for representation learning,'' in \emph{International Conference on Learning Representations}, 2019.

\bibitem{mackay1992practical}
D.~J. MacKay, ``A practical bayesian framework for backpropagation networks,'' \emph{Neural computation}, vol.~4, no.~3, pp. 448--472, 1992.

\bibitem{mackay1992bayesian}
D.~J. MacKay, ``Bayesian interpolation,'' \emph{Neural computation}, vol.~4, no.~3, pp. 415--447, 1992.

\bibitem{chen2020simple}
T.~Chen, S.~Kornblith, M.~Norouzi, and G.~Hinton, ``A simple framework for contrastive learning of visual representations,'' in \emph{International conference on machine learning}.\hskip 1em plus 0.5em minus 0.4em\relax PMLR, 2020, pp. 1597--1607.

\bibitem{he2020momentum}
K.~He, H.~Fan, Y.~Wu, S.~Xie, and R.~Girshick, ``Momentum contrast for unsupervised visual representation learning,'' in \emph{Proceedings of the IEEE/CVF conference on computer vision and pattern recognition}, 2020, pp. 9729--9738.

\bibitem{chen2021exploring}
X.~Chen and K.~He, ``Exploring simple siamese representation learning,'' in \emph{Proceedings of the IEEE/CVF conference on computer vision and pattern recognition}, 2021, pp. 15\,750--15\,758.

\bibitem{chopra2005learning}
S.~Chopra, R.~Hadsell, and Y.~LeCun, ``Learning a similarity metric discriminatively, with application to face verification,'' in \emph{2005 IEEE computer society conference on computer vision and pattern recognition (CVPR'05)}, vol.~1.\hskip 1em plus 0.5em minus 0.4em\relax IEEE, 2005, pp. 539--546.

\bibitem{schroff2015facenet}
F.~Schroff, D.~Kalenichenko, and J.~Philbin, ``Facenet: A unified embedding for face recognition and clustering,'' in \emph{Proceedings of the IEEE conference on computer vision and pattern recognition}, 2015, pp. 815--823.

\bibitem{sohn2016improved}
K.~Sohn, ``Improved deep metric learning with multi-class n-pair loss objective,'' \emph{Advances in neural information processing systems}, vol.~29, 2016.

\bibitem{ghasedi2017deep}
K.~Ghasedi~Dizaji, A.~Herandi, C.~Deng, W.~Cai, and H.~Huang, ``Deep clustering via joint convolutional autoencoder embedding and relative entropy minimization,'' in \emph{Proceedings of the IEEE international conference on computer vision}, 2017, pp. 5736--5745.

\bibitem{bishop:2006}
C.~M. Bishop, \emph{Pattern Recognition and Machine Learning}.\hskip 1em plus 0.5em minus 0.4em\relax Springer, 2006.

\bibitem{cuturi2013sinkhorn}
M.~Cuturi, ``Sinkhorn distances: Lightspeed computation of optimal transport,'' \emph{Advances in neural information processing systems}, vol.~26, 2013.

\bibitem{tanaka2018joint}
D.~Tanaka, D.~Ikami, T.~Yamasaki, and K.~Aizawa, ``Joint optimization framework for learning with noisy labels,'' in \emph{Proceedings of the IEEE conference on computer vision and pattern recognition}, 2018, pp. 5552--5560.

\bibitem{song2022learning}
H.~Song, M.~Kim, D.~Park, Y.~Shin, and J.-G. Lee, ``Learning from noisy labels with deep neural networks: A survey,'' \emph{IEEE Transactions on Neural Networks and Learning Systems}, 2022.

\bibitem{guo2017calibration}
C.~Guo, G.~Pleiss, Y.~Sun, and K.~Q. Weinberger, ``On calibration of modern neural networks,'' in \emph{International conference on machine learning}.\hskip 1em plus 0.5em minus 0.4em\relax PMLR, 2017, pp. 1321--1330.

\bibitem{muller2019does}
R.~M{\"u}ller, S.~Kornblith, and G.~E. Hinton, ``When does label smoothing help?'' \emph{Advances in neural information processing systems}, vol.~32, 2019.

\bibitem{pereyra2017regularizing}
G.~Pereyra, G.~Tucker, J.~Chorowski, {\L}.~Kaiser, and G.~Hinton, ``Regularizing neural networks by penalizing confident output distributions,'' \emph{ICLR workshop, arXiv:1701.06548}, 2017.

\bibitem{vapnik:2001}
A.~Ben-Hur, D.~Horn, H.~Siegelman, and V.~Vapnik, ``Support vector clustering,'' \emph{Journal of Machine Learning Research}, vol.~2, pp. 125 -- 137, 2001.

\bibitem{cristianini2000introduction}
N.~Cristianini and J.~Shawe-Taylor, \emph{An introduction to support vector machines and other kernel-based learning methods}.\hskip 1em plus 0.5em minus 0.4em\relax Cambridge university press, 2000.

\bibitem{mahajan2012planar}
M.~Mahajan, P.~Nimbhorkar, and K.~Varadarajan, ``The planar {K}-means problem is {NP}-hard,'' \emph{Theoretical Computer Science}, vol. 442, pp. 13--21, 2012.

\bibitem{rosset2003margin}
S.~Rosset, J.~Zhu, and T.~Hastie, ``Margin maximizing loss functions,'' \emph{Advances in neural information processing systems}, vol.~16, 2003.

\bibitem{Renyi1961}
A.~{R}\'{e}nyi, ``On measures of entropy and information,'' \emph{Fourth Berkeley Symp. Math. Stat. Probab.}, vol.~1, pp. 547--561, 1961.

\bibitem{vapnik:1995}
V.~Vapnik, \emph{The Nature of Statistical Learning Theory}.\hskip 1em plus 0.5em minus 0.4em\relax Springer, 1995.

\bibitem{boyd2004convex}
S.~Boyd and L.~Vandenberghe, \emph{Convex optimization}.\hskip 1em plus 0.5em minus 0.4em\relax Cambridge university press, 2004.

\bibitem{NSD}
``Natural {S}cenes {D}ataset [{NSD}],'' \url{https://www.kaggle.com/datasets/nitishabharathi/scene-classification}, 2020.

\bibitem{kelley1995iterative}
C.~T. Kelley, \emph{Iterative methods for linear and nonlinear equations}.\hskip 1em plus 0.5em minus 0.4em\relax SIAM, 1995.

\bibitem{springenberg2015unsupervised}
J.~T. Springenberg, ``Unsupervised and semi-supervised learning with categorical generative adversarial networks,'' in \emph{International Conference on Learning Representations}, 2015.

\bibitem{MNIST}
Y.~Lecun, L.~Bottou, Y.~Bengio, and P.~Haffner, ``Gradient-based learning applied to document recognition,'' \emph{Proceedings of the IEEE}, vol.~86, no.~11, pp. 2278--2324, 1998.

\bibitem{CIFAR}
A.~Torralba, R.~Fergus, and W.~T. Freeman, ``80 million tiny images: A large data set for nonparametric object and scene recognition,'' \emph{IEEE transactions on pattern analysis and machine intelligence}, vol.~30, no.~11, pp. 1958--1970, 2008.

\bibitem{STL}
A.~Coates, A.~Ng, and H.~Lee, ``An analysis of single-layer networks in unsupervised feature learning,'' in \emph{Proceedings of the fourteenth international conference on artificial intelligence and statistics}.\hskip 1em plus 0.5em minus 0.4em\relax JMLR Workshop and Conference Proceedings, 2011, pp. 215--223.

\bibitem{kuhn1955hungarian}
H.~W. Kuhn, ``The hungarian method for the assignment problem,'' \emph{Naval research logistics quarterly}, vol.~2, no. 1-2, pp. 83--97, 1955.

\bibitem{kingma2015adam}
D.~P. Kingma and J.~Ba, ``Adam: A method for stochastic optimization,'' in \emph{ICLR (Poster)}, 2015.

\bibitem{van2020scan}
W.~Van~Gansbeke, S.~Vandenhende, S.~Georgoulis, M.~Proesmans, and L.~Van~Gool, ``Scan: Learning to classify images without labels,'' in \emph{Computer Vision--ECCV 2020: 16th European Conference, Glasgow, UK, August 23--28, 2020, Proceedings, Part X}.\hskip 1em plus 0.5em minus 0.4em\relax Springer, 2020, pp. 268--285.

\bibitem{he2016deep}
K.~He, X.~Zhang, S.~Ren, and J.~Sun, ``Deep residual learning for image recognition,'' in \emph{Proceedings of the IEEE conference on computer vision and pattern recognition}, 2016, pp. 770--778.

\end{thebibliography}
\bibliographystyle{IEEEtran}

\newpage

\section{Biography Section}
 
\vspace{11pt}

\begin{IEEEbiography}[{\includegraphics[width=1in,height=1.25in,clip,keepaspectratio]{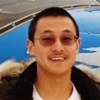}}]{Zhongwen (Rex) Zhang}
 is currently a Machine Learning Scientist in Synlico Inc. He completed his PhD degree in computer science at the University of Waterloo, Canada, supervised by Prof. Yuri Boykov. His PhD thesis is titled ``Unsupervised Losses for Clustering and Segmentation of Images: Theories \& Optimization Algorithms". He obtained M.Sc. in computer science in 2019 from the Western University, Canada, for his thesis titled ``Vessel Tree Reconstruction with Divergence Prior". Previously in 2016 he received B.Eng. in Biomedical Engineering from the Zhejiang University, China. His research interests included deep clustering, weakly-supervised image segmentation, and unsupervised vascular tree reconstruction. He is now working on T-cell engineering for solid tumor cancer therapy with the help of machine learning methodologies.
\end{IEEEbiography}

\begin{IEEEbiography}[{\includegraphics[width=1in,height=1.25in,clip,keepaspectratio]{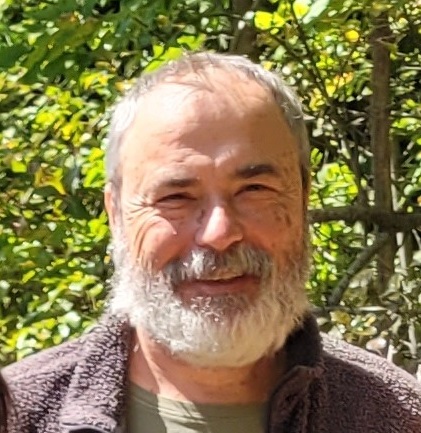}}]{Yuri Boykov}
is currently a Professor at the Cheriton School of Computer Science at the University of Waterloo, Canada. Previously, he was a Professor at the University of Western Ontario and a research scientist at Siemens Corporate Research institute in Princeton. His research is concentrated in the area of computer vision and medical image analysis with a focus on modeling and optimization for weakly-supervised segmentation, restoration, stereo, model fitting, recognition, photo-video editing, and other visual data analysis problems. His work includes one of the ten most influential papers in IEEE Transactions of Pattern Analysis and Machine Intelligence (TPAMI Top Picks for 30 years). In 2017 Google Scholar listed his work on segmentation as a "classic paper in computer vision and pattern recognition" (from 2006). In 2011 he received the Helmholtz Prize from IEEE and the Test of Time Award by the International Conference on Computer Vision. The Faculty of Science at the Western University recognized his work by awarding the Distinguished Research Professorship in 2014 and the Florence Bucke Prize in 2008. He received a "Diploma of Higher Education with Honors" at the Moscow Institute of Physics and Technology in 1992 and defended his Ph.D. at the Department of Operations Research at Cornell University, Ithaca, NY, in 1996.
\end{IEEEbiography}

\vfill

\end{document}